\PassOptionsToPackage{table}{xcolor}
\documentclass[12pt]{article}
\usepackage[top=1in, bottom=1in, left=1in, right=1in]{geometry}
\usepackage{indentfirst}
\usepackage{graphicx}
\usepackage{amssymb}
\usepackage{enumerate}
\usepackage{amsmath}
\usepackage{diagbox}
\usepackage{comment}
\usepackage{tikz}
\usepackage{multirow}
\usepackage{booktabs}

\begin{document}

\title{An Extensive Evaluation of Machine Learning and Deep Learning Approaches for Sentiment Analysis: Insights from Amazon Customer Reviews}

\author{ 
{Muhammed Ihsan Corak $^{1}$, Cuiyuan Wang $^{2}$ and Xiaowen Zhang $^{3}$*}\\
$^{1}$ Computer Science Dept., CUNY College of Staten Island \\ m.ihsan.corak@gmail.com\\
$^{2}$ Computer Science Dept., CUNY LaGuardia Community College \\ cuiyuanwang18@gmail.com\\
$^{3}$ Computer Science Dept., CUNY College of Staten Island \\ xiaowen.zhang@csi.cuny.edu}

\date{}  
\maketitle

\begin{abstract}

The rapid growth of user-generated text on online platforms creates opportunities for understanding customer opinion and supporting data-driven decisions. This study compares five traditional machine learning classifiers (Random Forest, Logistic Regression, Bernoulli Naive Bayes, Linear Support Vector Classifier, and Multinomial Naive Bayes) and six deep learning architectures (LSTM, Bi-LSTM, GRU, RNN, Attention, and FastText) for three-class sentiment classification of 568,454 Amazon Fine Food Reviews. Machine learning models used Bag-of-Words (BoW) and TF-IDF features limited to 5,000 terms; deep learning models used learned embeddings over a 20,000-word vocabulary. Models were evaluated on the original imbalanced data and on a random-oversampled balanced set of 1,070,469 reviews, using accuracy, macro and per-class precision, recall, and F1-score, and training time. On the balanced set, Random Forest with BoW achieved 99\% accuracy and the Attention model achieved 98.7\%. Linear SVC reached 89\% accuracy in 1 min 15 s, and Logistic Regression reached 90\% in 3 min 37 s. On the imbalanced set, the Attention model achieved 97.8\% accuracy, and FastText achieved 95.3\% in 3 min 26 s, while the RNN reached 91.9\% with a neutral-class F1-score of 0.08. Deep learning models were more stable across class distributions, whereas the best classical model depended on balancing. We also report a user-facing application that applies the trained Random Forest and Attention models to new review text. Because the labels are derived from star ratings, the results measure agreement with rating-derived sentiment classes rather than annotated sentiment; the dataset is a single source, and the findings should be read with that limitation in mind.

\end{abstract}



\section{Introduction}

In today’s data-driven world, online reviews have become an essential factor influencing consumer purchasing decisions and guiding business strategies. With the increasing dependence on digital platforms, customers frequently share their experiences and opinions online, providing organizations with valuable insights into customer satisfaction, product quality, and emerging market trends. Among these platforms, Amazon stands out as one of the largest e-commerce ecosystems, hosting millions of user-generated reviews that form a rich and diverse source of unstructured textual data. Analyzing such large-scale customer feedback systematically can enable businesses to improve service quality, optimize product offerings, and strengthen customer engagement.

Sentiment analysis, a subfield of Natural Language Processing (NLP), focuses on identifying and classifying opinions expressed in text as positive, negative, or neutral. It has become a powerful analytical tool for extracting emotional tone from textual data and converting subjective information into actionable business intelligence. Traditional machine learning models including Random Forest Classifier (RFC), Logistic Regression Classifier (LRC), Bernoulli Naive Bayes (BNB), Multinomial Naive Bayes (MNB) and Linear Support Vector Classifier (L-SVC) have long been used to perform sentiment classification. These approaches rely on manually crafted features and text representation methods like Bag of Words (BoW) and Term Frequency–Inverse Document Frequency (TF-IDF). While these models are efficient and interpretable, their ability to capture deep semantic meaning and contextual relationships in language remains limited.

In recent years, deep learning models have demonstrated superior performance in sentiment analysis due to their capability to learn hierarchical representations directly from text data. Architectures such as Long Short-Term Memory (LSTM), Bidirectional Long Short-Term Memory (Bi-LSTM), Gated Recurrent Units (GRU), Recurrent Neural Networks (RNN), and Attention-based mechanisms can effectively capture long-range dependencies and contextual nuances. Word embeddings such as FastText further enhance model understanding by encoding semantic relationships between words. These developments have significantly advanced the field, making deep learning models particularly effective for complex text classification tasks. However, they often come with higher computational costs and longer training times, which may affect scalability and real-time applicability.

This study presents a comprehensive analysis of sentiment classification on Amazon customer reviews [1] by integrating both traditional machine learning and deep learning approaches. Using a large dataset containing over half a million customer reviews, the research systematically evaluates multiple models trained with BoW and TF-IDF features as well as deep learning models leveraging word embeddings. A key component of this work is the examination of model performance across both balanced and imbalanced datasets to assess the effect of data distribution on classification accuracy and fairness. To enhance computational efficiency, experiments were conducted in a GPU T4 x2 environment, utilizing two NVIDIA T4 GPUs for parallel training and accelerated model optimization.

Beyond comparative performance analysis, a practical sentiment analysis software system was developed to demonstrate real-time classification capabilities and user interaction. This integration of research and application underscores the importance of combining methodological rigor with practical usability. The findings from this study offer a holistic perspective on the relative strengths, limitations, and trade-offs between machine learning and deep learning models. By providing insights into model selection, data balancing strategies, and computational efficiency, this research contributes a unified framework for large-scale sentiment analysis and offers guidance for both academic research and industry applications.

The remainder of this paper is organized as follows. Section 2 reviews related work on sentiment analysis of Amazon reviews and identifies the gap this study addresses. Section 3 describes the dataset, preprocessing, and data balancing procedures. Section 4 presents the machine learning models and their results. Section 5 presents the deep learning models and their results. Section 6 compares the two families under balanced and imbalanced conditions. Section 7 reports the best-performing models and the application prototype, and Section 8 discusses the findings and limitations. Section 9 concludes.


\section{Literature Review}
The studies reviewed above share three recurring characteristics. First, most evaluate a small number of models, often one classical and one neural architecture, so that the relative standing of the wider model families is unclear. Second, class imbalance is noted as important but is rarely tested systematically: the balancing strategy, its effect on each model family, and its validity (for example, whether oversampling leaks information into evaluation) are seldom examined. Third, the labels used are typically star ratings, and the mismatch between rating and expressed sentiment, particularly for three-star reviews, is acknowledged but not quantified in terms of its effect on model comparison.

This study addresses these gaps in three ways. It evaluates eleven models across two feature representations on the same large corpus. It reports results on both imbalanced data and a balanced set constructed by random oversampling, and it reports how each family responds to that change. It also documents the labeling scheme and the class composition explicitly, so that the results can be interpreted in terms of what the labels measure.

Sentiment analysis has evolved from basic text classification to advanced machine learning and deep learning techniques enabling more accurate and efficient interpretation of user-generated textual data. With the rise of digitalization, large-scale datasets such as Amazon reviews offer valuable opportunities to benchmark and compare different sentiment classification methodologies, helping address challenges like dataset bias through balancing techniques. Studies across platforms-from e-commerce to social media-highlight innovations such as ensemble methods and multi-polarity classification, all contributing to the development of more effective models and reinforcing sentiment analysis as a crucial tool for extracting meaningful insights and supporting informed business decision-making.

The study [2] compares machine learning algorithms for sentiment classification on Amazon reviews, using a Kaggle dataset with 4 million reviews. LSTM networks outperformed other models, achieving 90\% accuracy and 0.96  Area Under Curve (AUC), with even higher accuracy (92\%) for furniture reviews. Unlike the reviewed study, which excludes neutral sentiments and uses TF-IDF for feature extraction, this project includes neutral sentiments and employs both TF-IDF and BoW methods. The findings highlight LSTM's effectiveness across product categories and its potential for expansion to include neutral sentiment classification.

Another study [3] investigates sentiment analysis on Amazon mobile phone reviews, highlighting the challenge of ambiguous 3-star ratings. Using models like LRC, Naive Bayesian (NB), RFC, Bi-LSTM with Global Vectors (GloVe) embeddings, and Bidirectional Encoder Representations from Transformers (BERT), the research identifies BERT as the most effective model, achieving 94\% accuracy for multiclass and 98\% for binary classification. Although sharing similarities with this project in sentiment classification, it differs by focusing on mobile phone reviews, emphasizing 3-star ratings, and employing GloVe embeddings and BERT. This contrasts with this project's analysis of food reviews and its use of uses VADER, TF-IDF, and BoW for feature extraction.

The reviewed research [4] analyzes Amazon product reviews using GRU, RNNs, and Support Vector Machine (SVM) on 28,000 reviews from 60 products, achieving 81.82\% accuracy by integrating review embeddings with product data. It highlights the importance of machine learning in understanding customer feedback and dataset balancing, with suggestions for future work in expanding product categories. While both projects focus on sentiment classification using Amazon reviews, this project uses over 500,000 reviews with TF-IDF and BoW, whereas the reviewed study VADER relies on star ratings meanwhile, this project applies sentiment scoring for a more nuanced analysis.

Previous studies have taken diverse approaches to sentiment analysis. One significant contribution was made by a study that proposed a Pearson Correlation Coefficient-based Harris Hawks Optimization (PCCHHO) algorithm combined with RNN and LSTM models [5]. This approach emphasized feature selection using the Pearson correlation coefficient for initial dimensionality reduction, followed by Harris Hawks Optimization for non-redundant feature selection. The use of RNN-LSTM aimed to classify sentiments with high precision, achieving 95.8\% accuracy on Amazon product reviews. The study highlighted the effectiveness of feature extraction and optimization techniques in improving sentiment classification but did not focus on user-centric interface.

Another study employed a CNN model for classifying text reviews into positive or negative categories [6]. The research compared CNN with several word embedding models, such as Word2Vec (Word to Vector), GloVe, and FastText, finding that the inclusion of stop words improved accuracy by 2\%. The study emphasized the benefits of training word embeddings to learn better features in less computation time. Although the CNN-based approach demonstrated strong performance, it doesn't includes a user interface and did not explore multi-class sentiment classification.

The study utilized supervised machine learning techniques to polarize Amazon reviews into categories using bag-of-words and TF-IDF feature extraction methods combined with chi-square analysis [7]. This research focused on active learning strategies to label large datasets and achieved satisfactory results through manual and automatic learning processes. Despite its emphasis on feature extraction and supervised learning, the study did not delve into advanced deep learning architectures or the benefits of contextual embeddings for sentiment analysis.

The work by [8] provides a comprehensive survey of sentiment analysis techniques, covering both traditional machine learning and deep learning approaches. Traditional machine learning techniques, such as NB, SVM, and LRC, have been widely used for sentiment classification but require extensive feature engineering and struggle with capturing contextual dependencies. Deep learning models, particularly RNNs, LSTM networks, and CNNs, have demonstrated superior performance in handling sequential text data and learning complex linguistic features. Additionally, the study discusses the impact of pre-trained language models such as BERT and GPT, which have significantly improved sentiment analysis by leveraging transfer learning. The limitations of deep learning models, including computational cost and data dependency, are also highlighted. This survey provides an important foundation for understanding the advancements in sentiment analysis and emphasizes the need for comparative studies across different model architectures.

A related study by [9] presents a comparative analysis of sentiment classification models applied to Amazon customer reviews, similar to the focus of this research. The authors evaluate traditional machine learning models, including LRC, RFC, and NB, against DL architectures such as CNNs and RNNs. Their findings suggest that deep learning models tend to outperform traditional approaches, particularly in capturing nuanced sentiment variations. However, the study also notes that machine learning models remain competitive, offering lower computational costs and better interpretability. Additionally, the authors emphasize the importance of dataset balancing techniques, as class imbalance in sentiment datasets can significantly impact model performance. This study aligns with our research objectives by exploring the trade-offs between machine learning and deep learning models in sentiment classification.

Further insights are provided by [10], which reviews recent advancements in machine learning and deep learning for sentiment analysis. The study categorizes sentiment analysis models based on their structural complexity and evaluates their effectiveness in various applications, including social media monitoring and customer feedback analysis. The authors discuss emerging challenges in the field, such as the need for better context understanding, handling sarcasm, and improving model interpretability. Their findings underscore the potential of attention mechanisms and transformer-based architectures in enhancing sentiment classification accuracy. Additionally, they highlight the growing trend of hybrid models that combine ML and DL techniques to leverage the strengths of both approaches.

Overall, the existing body of research demonstrates a clear progression from traditional machine learning techniques toward more advanced deep learning and transformer-based architectures, driven by the increasing complexity and scale of sentiment-rich datasets. While earlier studies highlight the effectiveness of models such as NB, SVM, LRC, and even CNNs for binary or domain-specific tasks, more recent work consistently shows that architectures like LSTM, Bi-LSTM, GRU, and BERT excel at capturing contextual and sequential nuances in text. At the same time, challenges such as class imbalance, ambiguous ratings, and the need for efficient feature extraction continue to shape methodological choices across studies. Despite differing datasets, feature engineering strategies, and model architectures, the reviewed literature collectively emphasizes the importance of comparative evaluations to understand trade-offs between accuracy, efficiency, and interpretability. These insights underscore the need for comprehensive, large-scale analyses-such as the one undertaken in this study-to systematically assess model performance across machine learning and deep learning methods while incorporating balanced datasets and multi-class sentiment perspectives.


\section{Materials and Methods}

\subsection{Data Ingestion}\label{AA}

The dataset utilized in this study consists of Amazon Fine Food Reviews collected over a thirteen-year period, spanning from October 1999 to October 2012. This dataset serves as a comprehensive resource for sentiment analysis research, as it encompasses both textual review content and associated metadata, including product identifiers, user information, and numerical ratings. In total, the dataset contains 568,454 individual reviews, reflecting diverse customer opinions and experiences across a broad range of products.

Each review record includes attributes such as the product ID, user ID, profile name, helpfulness score, review score, review summary, and detailed text review. The distribution of reviews across star ratings reveals a notable imbalance: 363,122 reviews correspond to 5-star ratings, 80,655 to 4-star, 42,640 to 3-star, 29,769 to 2-star, and 52,268 to 1-star ratings, as illustrated in Figure 1. Figure 2 presents the oversampled balanced dataset. The data balancing techniques will be discussed in the subsequent sections. This skew toward higher ratings is typical of e-commerce feedback datasets, highlighting the need for balanced analysis techniques.

Additionally, the dataset comprises 256,059 unique users and 74,258 distinct products, indicating a high degree of user and product diversity. While the primary focus is on fine food products, the dataset also includes reviews from other Amazon categories, contributing to a more generalized representation of customer sentiment. The inclusion of heterogeneous product categories enhances the dataset’s applicability for large-scale sentiment modeling and domain-independent text analysis.

\begin{figure}[h]
    \centering
    \begin{minipage}{0.45\linewidth}
        \centering
        \includegraphics[width=\linewidth]{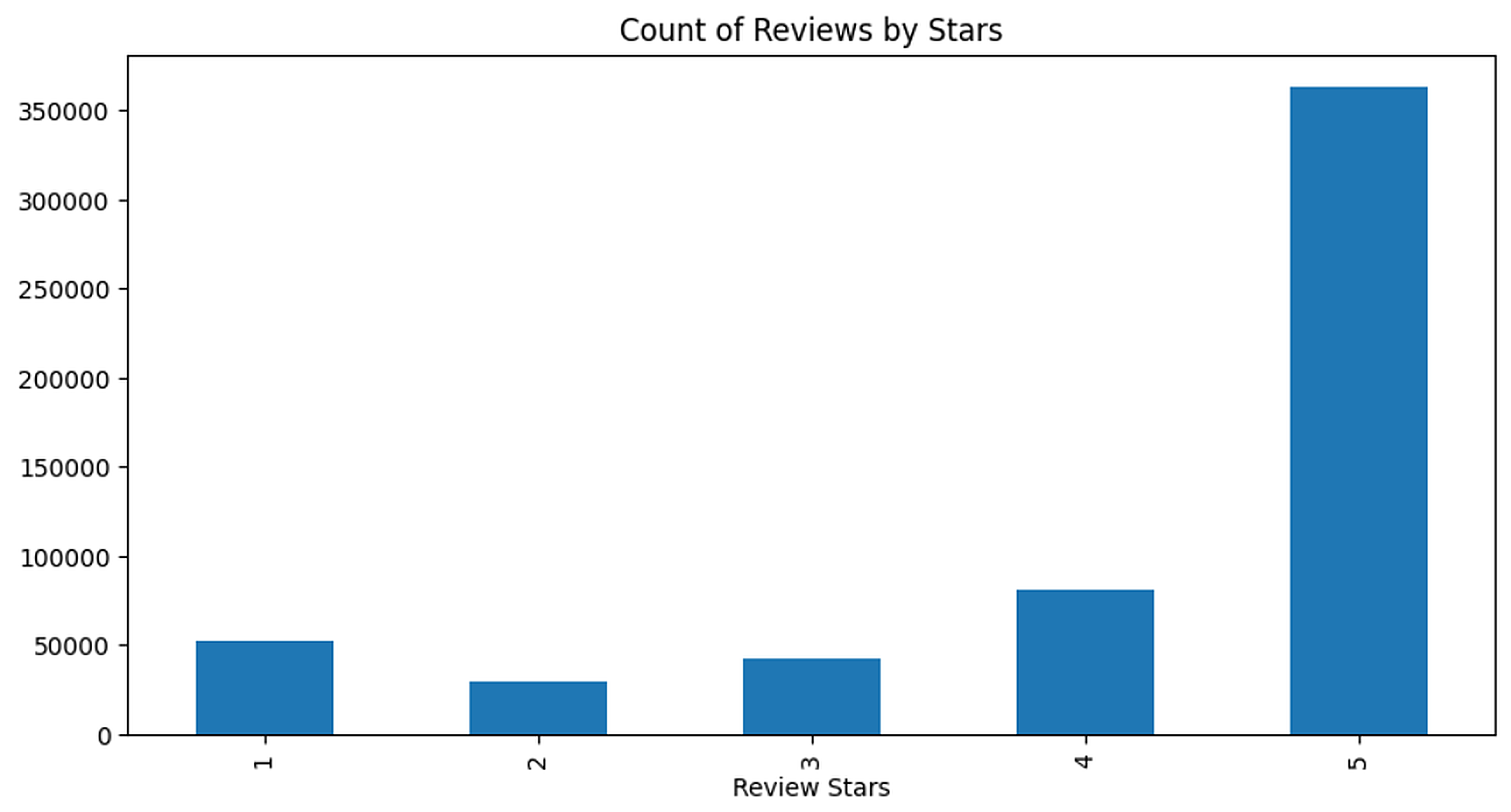}
        \caption{Count Reviews by Stars}
        \label{fig:count-reviews}
    \end{minipage}
    \hfill
    \begin{minipage}{0.5\linewidth}
        \centering
        \includegraphics[width=\linewidth]{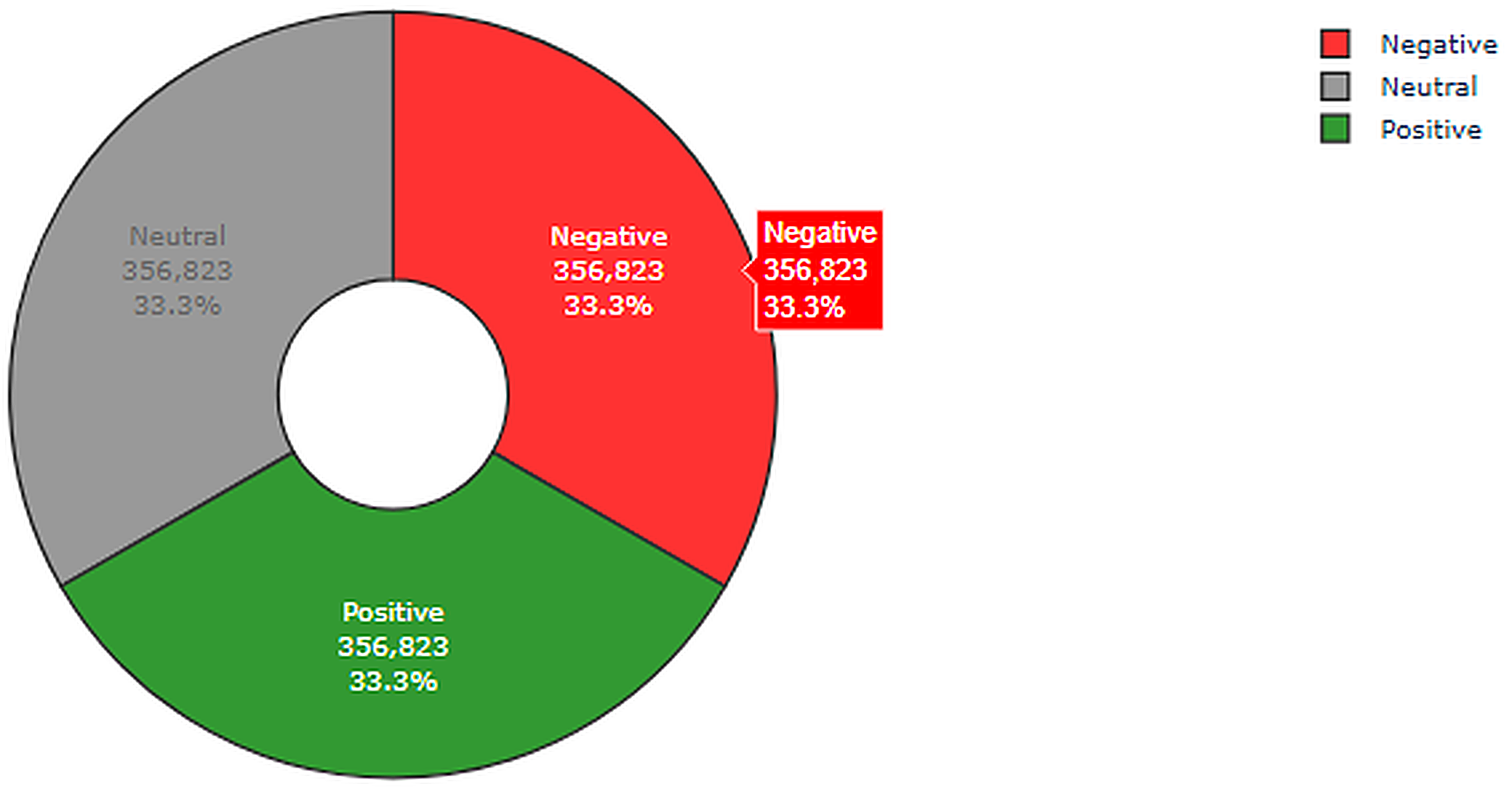}
        \caption{Oversampled Dataset}
        \label{fig:oversampled}
    \end{minipage}
\end{figure}

\subsection{Sentiment and Review Count Over Time}\label{AA}

\begin{figure}[h]
    \centering
    \includegraphics[width=0.8\linewidth]{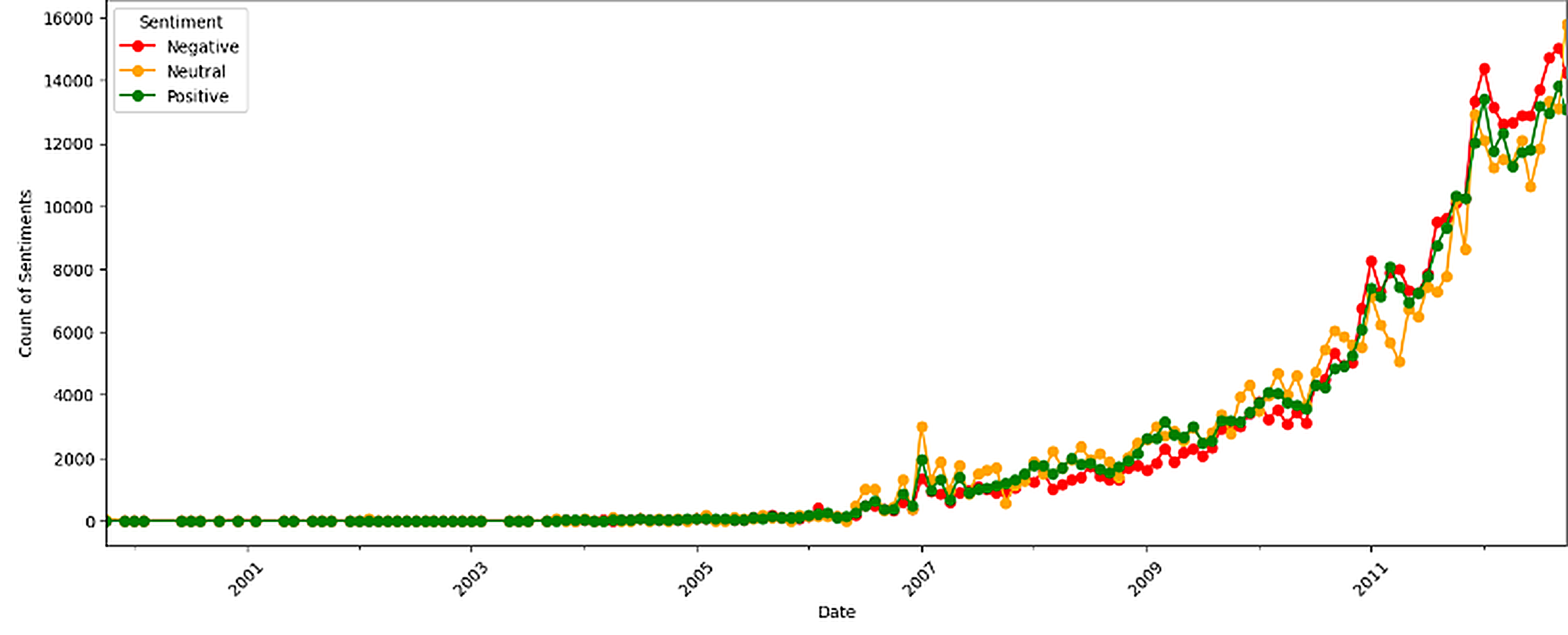}
    \caption{Review Count Over Time}
    \label{fig:enter-label}
\end{figure}

Figure 3 shows, the dataset spans over a decade, capturing reviews from October 1999 to October 2012. Analyzing the review counts over this period reveals relatively similar levels from the early 1999 until 2007. However, a noticeable upward trend begins between 2007 and 2009, followed by a significant increase from 2009 to 2012. This shift may reflect changes in customer behavior during this time, potentially correlated with the rapid growth of e-commerce platforms and online food delivery services. These developments likely influenced consumer engagement and review activity, as evidenced by the significant increase in sentiment score counts in the dataset. The findings highlight the interaction between technological developments and customer feedback trends and offer insights into changing consumer behavior in this transformative period.
\subsection{Top Used Words For Each Sentiment Class}\label{AA}

\begin{table} 
\renewcommand{\arraystretch}{1.3} 
\begin{center}
\resizebox{0.75\textwidth}{!}{%
\begin{tabular}{|c|c|c|c|c|c|} 
\hline
\multicolumn{2}{|c|}{\textbf{Positive}} & \multicolumn{2}{c|}{\textbf{Negative}} & \multicolumn{2}{c|}{\textbf{Neutral}} \\ 
\hline
\textbf{Word} & \textbf{Count} & \textbf{Word} & \textbf{Count} & \textbf{Word} & \textbf{Count} \\ 
\hline
good               & 200,638          & never           & 39,275          & fine              & 16,110           \\ 
\hline
great              & 167,175          & bad             & 32,186          & grocery store     & 12,679           \\ 
\hline
love               & 127,520          & disappointed    & 16,624          & first time        & 7,795            \\ 
\hline
even               & 75,000           & taste like      & 11,651          & normal            & 7,265            \\ 
\hline
well               & 73,408           & awful           & 5,109           & even though       & 6,755            \\ 
\hline
delicious          & 41,471           & worse           & 3,654           & okay              & 5,880            \\ 
\hline
perfect            & 32,959           & disgusting      & 2,388           & average           & 3,609            \\ 
\hline
worth              & 22,831           & nothing like    & 2,193           & expiration date   & 3,263            \\ 
\hline
great product      & 9,269            & never buy       & 1,626           & not bad           & 3,256            \\ 
\hline
highly recommended & 5,027            & bad taste       & 465             & ive tried         & 199              \\ 
\hline
\end{tabular}%
}
\end{center}
\caption{Sentiment Word Counts for Each Class} 
\label{tab:sentiment_words}
\end{table}

Defining top-used words, including bigrams (pairs of consecutive words), for each sentiment in this project is essential for understanding sentiment-specific patterns and linguistic trends. Bigrams capture contextual nuances, such as “highly recommended'' for positive reviews or “poor quality'' for negative ones, offering deeper insights into customer sentiment. This helps identify positive themes like satisfaction (great, love) and negative pain points (bad, disappointed), providing actionable insights into preferences and complaints. While “good'' is the most used word on positive reviews, it was “never'' for negative reviews and “fine'' for neutral reviews as showed in Table 1.

This step improves data preprocessing, enabling noise reduction and better feature engineering, while enhancing model performance by focusing on meaningful sentiment indicators. Moreover, it supports clear communication through visualizations like word clouds and reinforces findings with transparency, aiding reproducibility, and cross-dataset comparisons. By uncovering key patterns, it bridges raw data and actionable insights, enriching both academic rigor and business applicability.

\subsubsection{Weights of Words}

\begin{figure}
    \centering
    \includegraphics[width=0.5\linewidth]{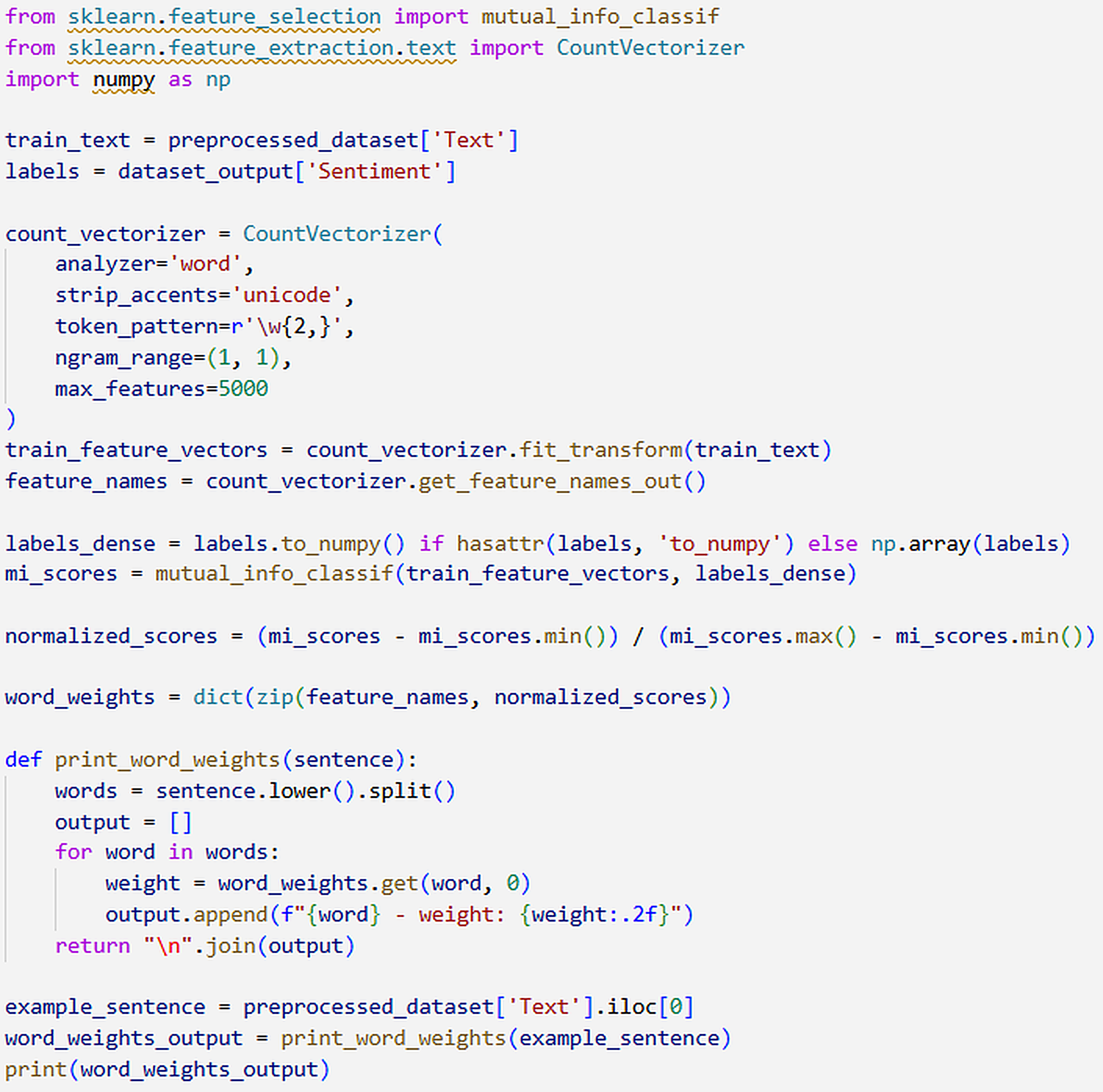}
    \caption{Word Weights Coding}
    \label{fig:placeholder}
\end{figure}

Word weighting is important because it enables the quantification of sentiment strength in textual data, providing a more nuanced representation of the emotions expressed in the text. In sentiment analysis, the significance of individual words can vary greatly depending on their context, polarity, and intensity. By assigning appropriate weights to words based on their sentiment value, we can better capture the underlying emotional tone of the content. In this study, we utilized VADER (Valence Aware Dictionary and sEntiment Reasoner), a lexicon and rule-based sentiment analysis tool, to assign sentiment weights to words (Figure 4). VADER evaluates the sentiment of words by considering their inherent valence (positive or negative orientation) as well as contextual factors such as punctuation and capitalization, which contribute to the intensity of sentiment. By applying VADER's word weights, we ensure a more accurate representation of sentiment, where positive words receive higher positive weights, negative words are given negative weights, and neutral words are assigned a value of zero. This method enhances the precision of sentiment classification models, ensuring that the analysis captures both the strength and direction of sentiment, which is particularly valuable in applications such as customer review analysis and social media monitoring.

Neutral sentiment words like “not bad'' (0.431), “okay'' (0.2263), and “fine'' (0.2023) are assigned non-zero weights because they carry mild positive connotations despite being categorized as neutral, as can be seen in Table 2. These words suggest a slightly favorable or acceptable tone, with “not bad'' implying mild positivity, “okay'' indicating satisfactory acceptance, and “fine'' reflecting adequacy. The assigned weights capture these subtle nuances, helping the model recognize the slight sentiment bias they contribute.

\begin{table}

\renewcommand{\arraystretch}{1.3} 
\begin{center}
\resizebox{0.75\textwidth}{!}{%
\begin{tabular}{|c|c|c|c|c|c|} 
\hline
\multicolumn{2}{|c|}{\textbf{Positive}} & \multicolumn{2}{c|}{\textbf{Negative}} & \multicolumn{2}{c|}{\textbf{Neutral}} \\ 
\hline
\textbf{Word} & \textbf{Weight} & \textbf{Word} & \textbf{Weight} & \textbf{Word} & \textbf{Weight} \\ 
\hline
highly recommended & 0.8534            & never buy       & -0.93             & not bad           & 0.431             \\ 
\hline
love               & 0.6369            & bad taste       & -0.5423           & okay              & 0.2263            \\ 
\hline
great              & 0.6249            & bad             & -0.5423           & fine              & 0.2023            \\ 
\hline
great product      & 0.6249            & disgusting      & -0.5267           & average           & 0.0               \\ 
\hline
delicious          & 0.5719            & never           & -0.5234           & normal            & 0.0               \\ 
\hline
perfect            & 0.5719            & worse           & -0.4767           & grocery store     & 0.0               \\ 
\hline
good               & 0.4404            & disappointed    & -0.47             & expiration date   & 0.0               \\ 
\hline
not bad           & 0.431             & awful           & -0.4588           & first time        & 0.0               \\ 
\hline
well               & 0.2732            & taste like      & -0.2852           & ive tried         & 0.0               \\ 
\hline
worth              & 0.2263            & nothing like    & -0.2755           & even though       & 0.0               \\ 
\hline
\end{tabular}%
}
\end{center}
\caption{Sentiment Word Weights for Each Class}
\label{tab:sentiment_words_weights}
\end{table}

\subsubsection{Preprocessing}\label{AA}
The preprocessing stage represents a fundamental step in NLP workflows, serving as the foundation for accurate and reliable sentiment analysis. Its primary objective is to transform raw and unstructured textual data into a clean, standardized, and machine-readable format suitable for model training and evaluation. Since raw customer reviews often contain noise such as spelling variations, redundant characters, and non-textual artifacts, preprocessing is essential to enhance data quality and analytical consistency.

In this study, a comprehensive preprocessing pipeline was implemented (Figure 5) to ensure the dataset’s uniformity and analytical readiness. The process began with tokenization, which segmented text into individual words or tokens, forming the basic input units for feature extraction and model training. Subsequently, duplicate and irrelevant entries were removed to eliminate redundancy and enhance dataset integrity. Punctuation marks, URLs, hashtags, emojis, and stop words were systematically removed or replaced using carefully designed cleaning functions. All text data were then converted to lowercase to ensure case-insensitive consistency across the corpus.

\begin{figure}[h]
    \centering
    \includegraphics[width=0.5\linewidth]{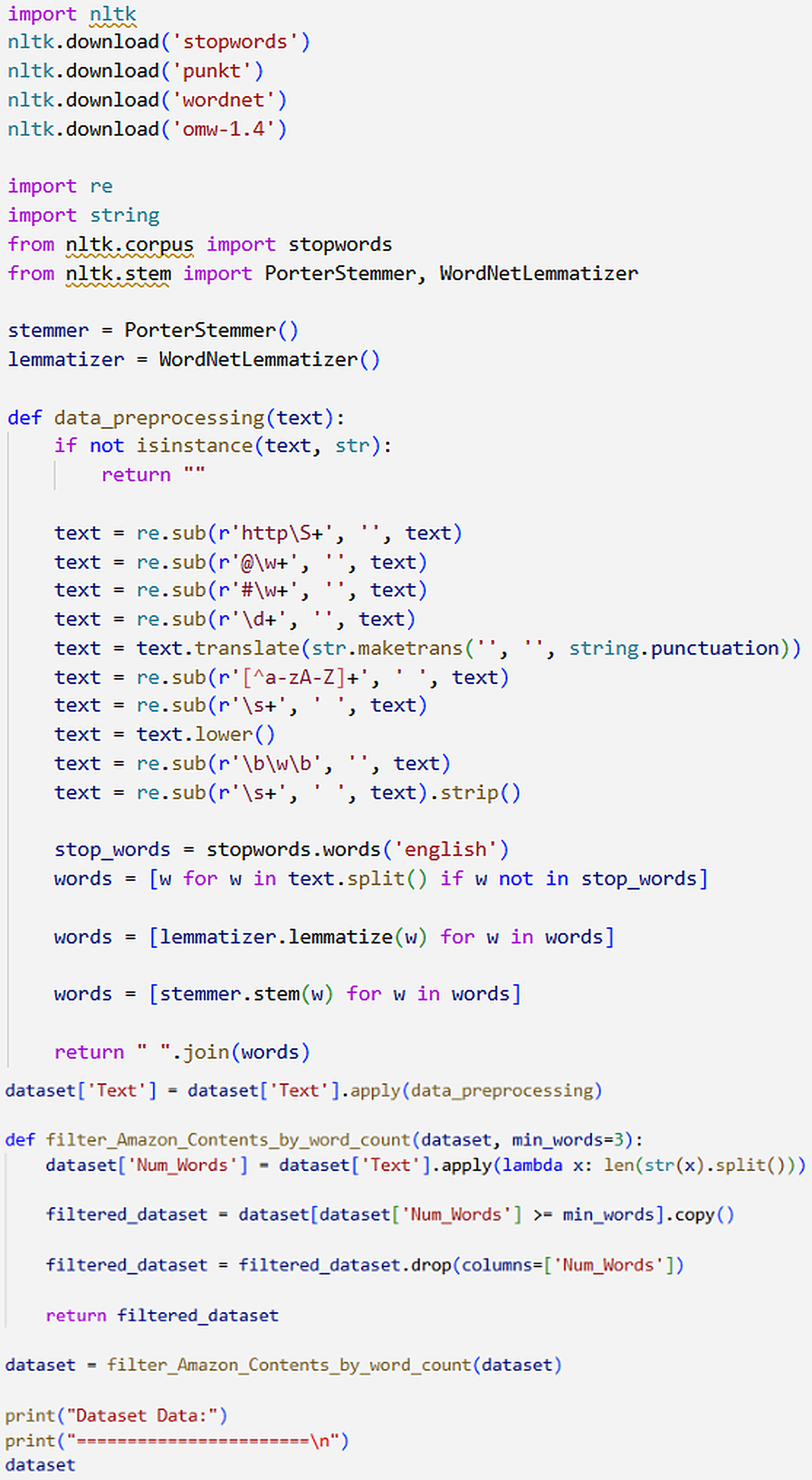}
    \caption{Preprocessing Coding}
    \label{fig:placeholder}
\end{figure}

\begin{figure}[h]
    \centering
    \includegraphics[width=0.5\linewidth]{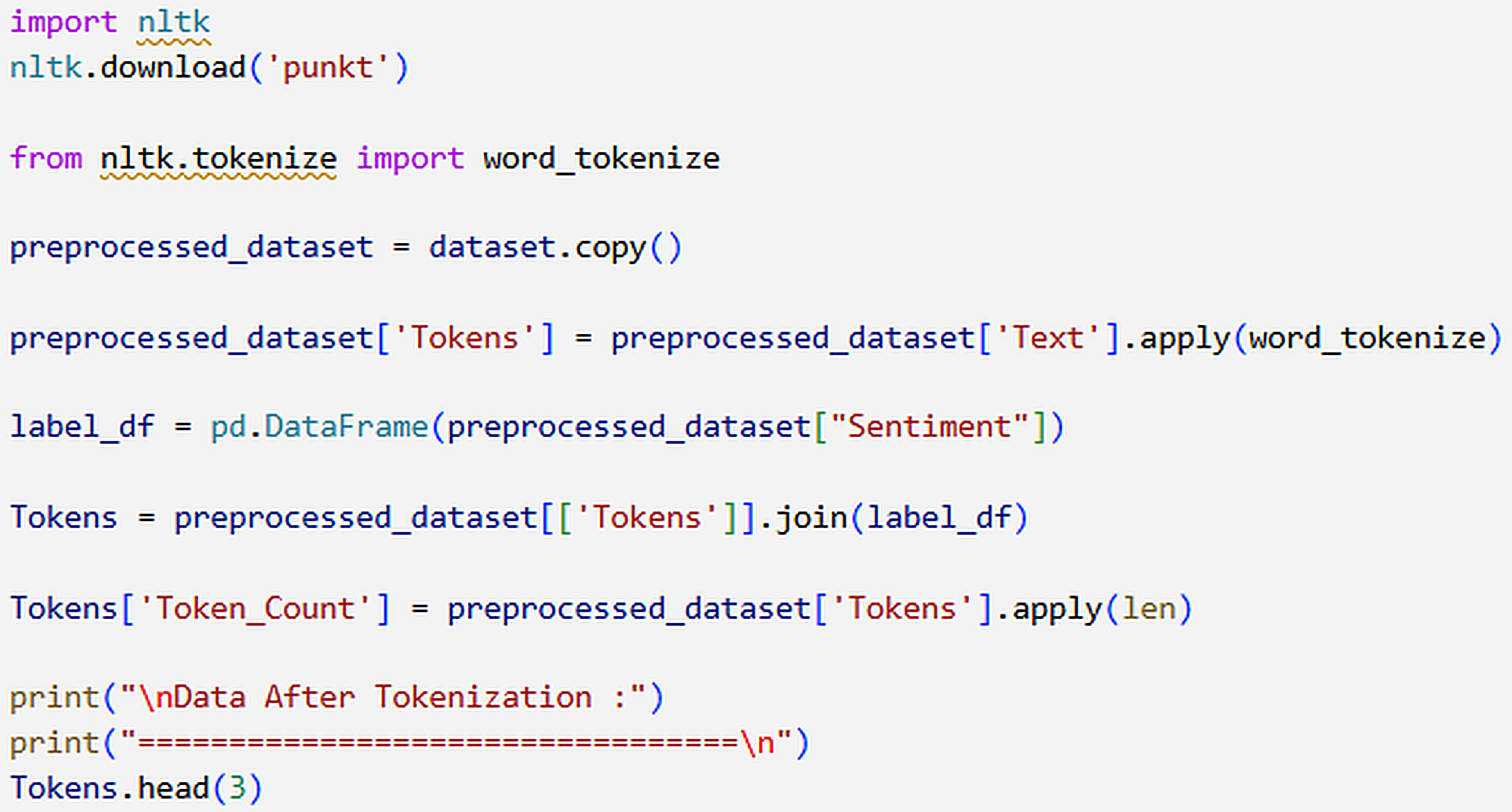}
    \caption{Tokenization Coding}
    \label{fig:placeholder}
\end{figure}

Further linguistic normalization was achieved through the application of stemming and lemmatization, both of which reduced words to their base or root forms, minimizing vocabulary size and improving generalization across linguistic variants. Regular expressions (RegEx) were extensively employed throughout the process to identify and filter specific text patterns, such as special characters and formatting inconsistencies.

As a result of these preprocessing operations-including noise removal, text normalization, and data quality refinement-the dataset was reduced from its original 568,454 reviews to 395,417 cleaned and usable entries, consisting of 356,823 positive, 32,312 negative, and 6,012 neutral reviews and also balancing techniques applied. (Figure 6 to 9) This substantial reduction underscores the importance of rigorous text cleaning in eliminating noisy and unreadable content. The final preprocessed dataset achieved a higher level of clarity, relevance, and interpretability, thus providing a solid foundation for the subsequent feature extraction, model training, and evaluation stages.

\begin{table} 
\caption{Text preprocessing in the machine learning and deep learning pipelines.}
\label{tab:preproc}
\centering
\footnotesize
\begin{tabular}{p{3.6cm}p{5.2cm}p{5.2cm}}
\toprule
\textbf{Step} & \textbf{ML pipeline} & \textbf{DL pipeline} \\
\midrule
Lowercasing & Yes & Yes \\
Removal of URLs, mentions, hashtags, digits, punctuation & Yes & HTML tags and non-alphabetic characters removed \\
Stopword removal & Yes (NLTK English list) & No \\
Stemming / lemmatization & No & No \\
Vectorization & BoW or TF-IDF, 5000 features & Integer sequences, padded to 100 \\
\bottomrule
\end{tabular}
\end{table}

\subsubsection{Data Balancing Methods}
The dataset exhibited class imbalance, with the majority class significantly outnumbering the minority classes. To address this, random oversampling was performed, increasing the minority classes (neu. and neg.) to create a balanced dataset of 1,070,469 reviews, as illustrated in Figure 2. To validate the effectiveness of this technique and mitigate bias, models were tested on undersampled subsets (25\%, 50\%, and 75\%) of the dataset, ensuring that oversampling did not artificially inflate performance. 

\begin{figure}
    \centering
    \includegraphics[width=0.5\linewidth]{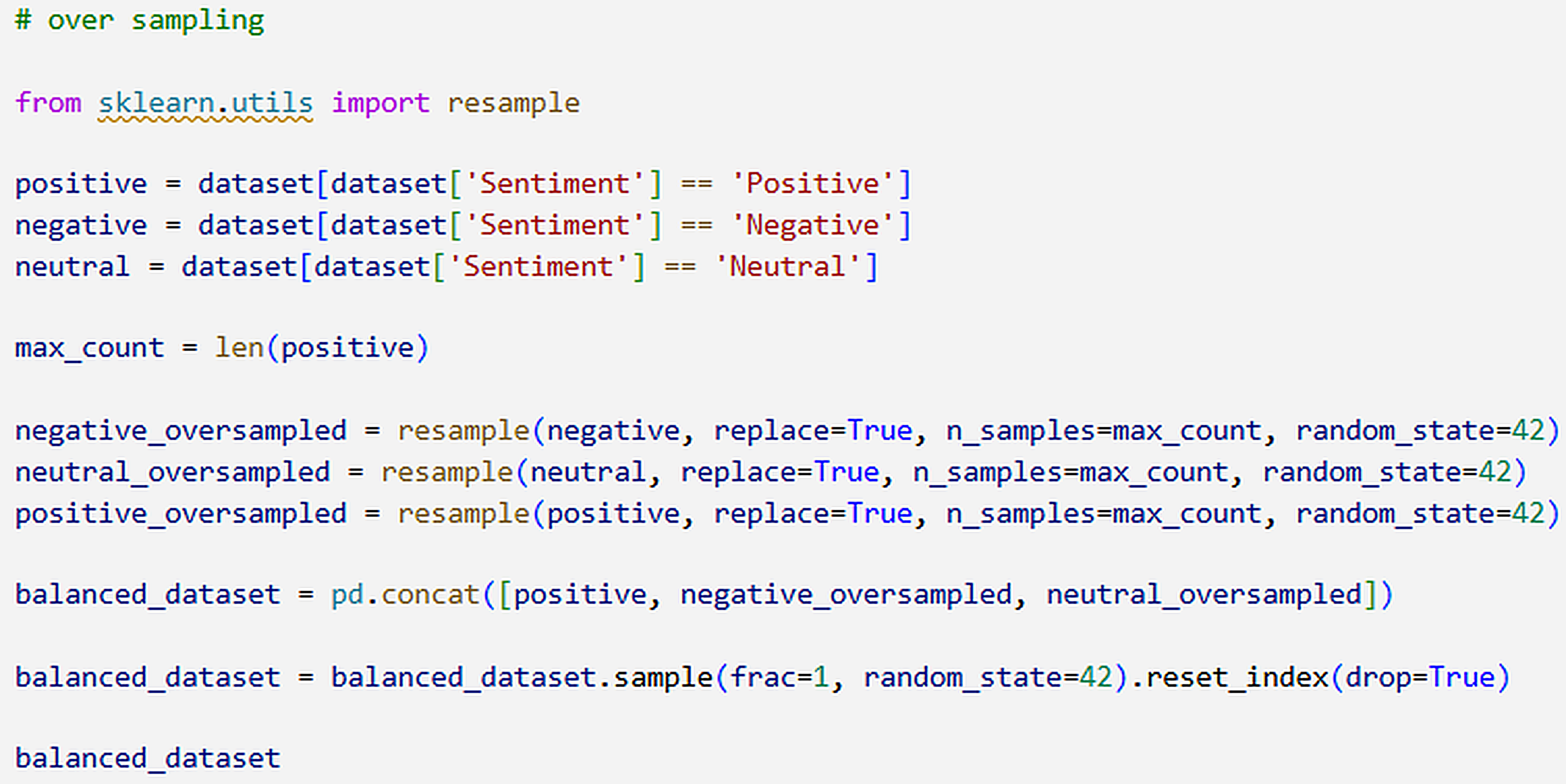}
    \caption{Over Sampling}
    \label{fig:placeholder}
\end{figure}

\begin{figure}
    \centering
    \includegraphics[width=0.5\linewidth]{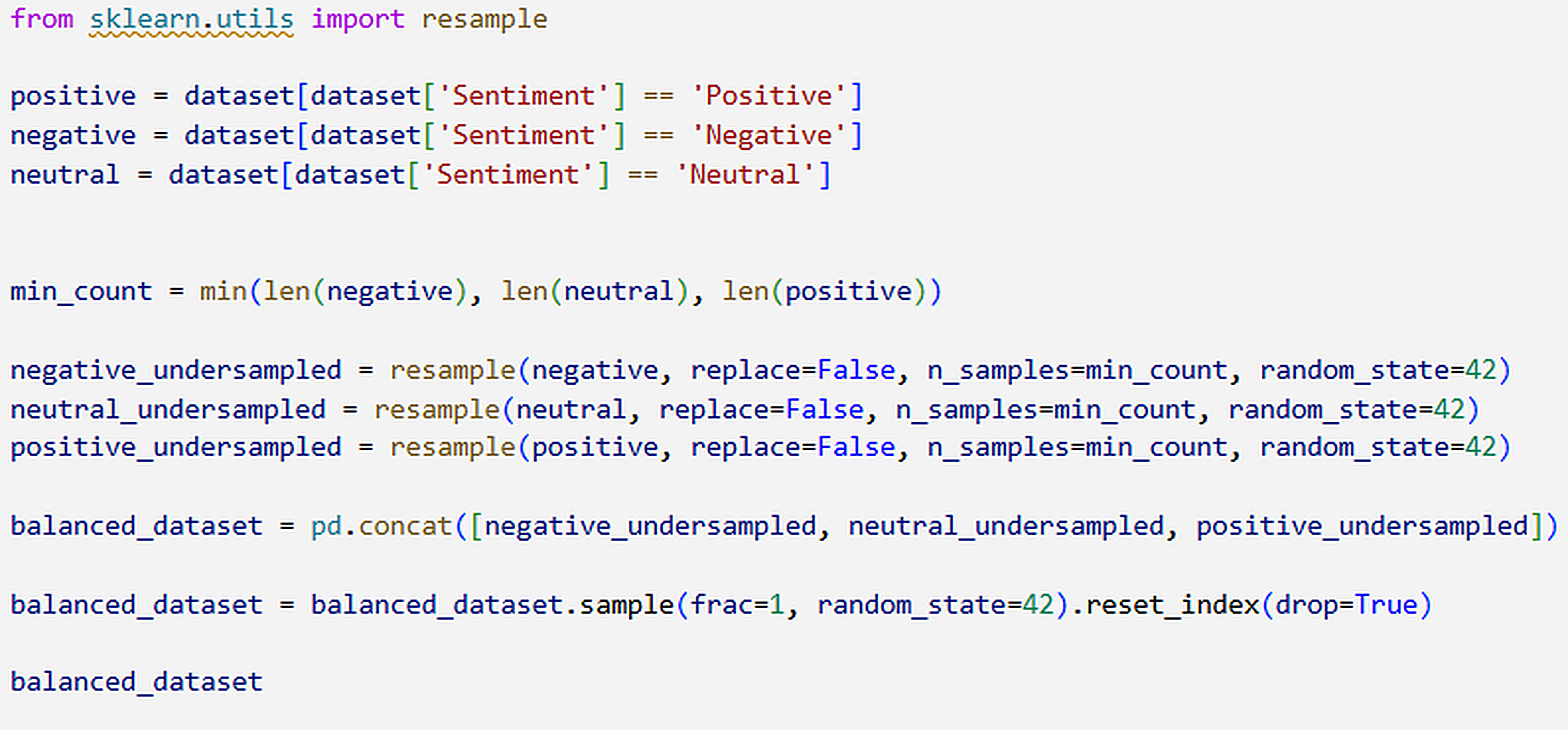}
    \caption{Under Sampling}
    \label{fig:placeholder}
\end{figure}

\begin{figure}
    \centering
    \includegraphics[width=0.5\linewidth]{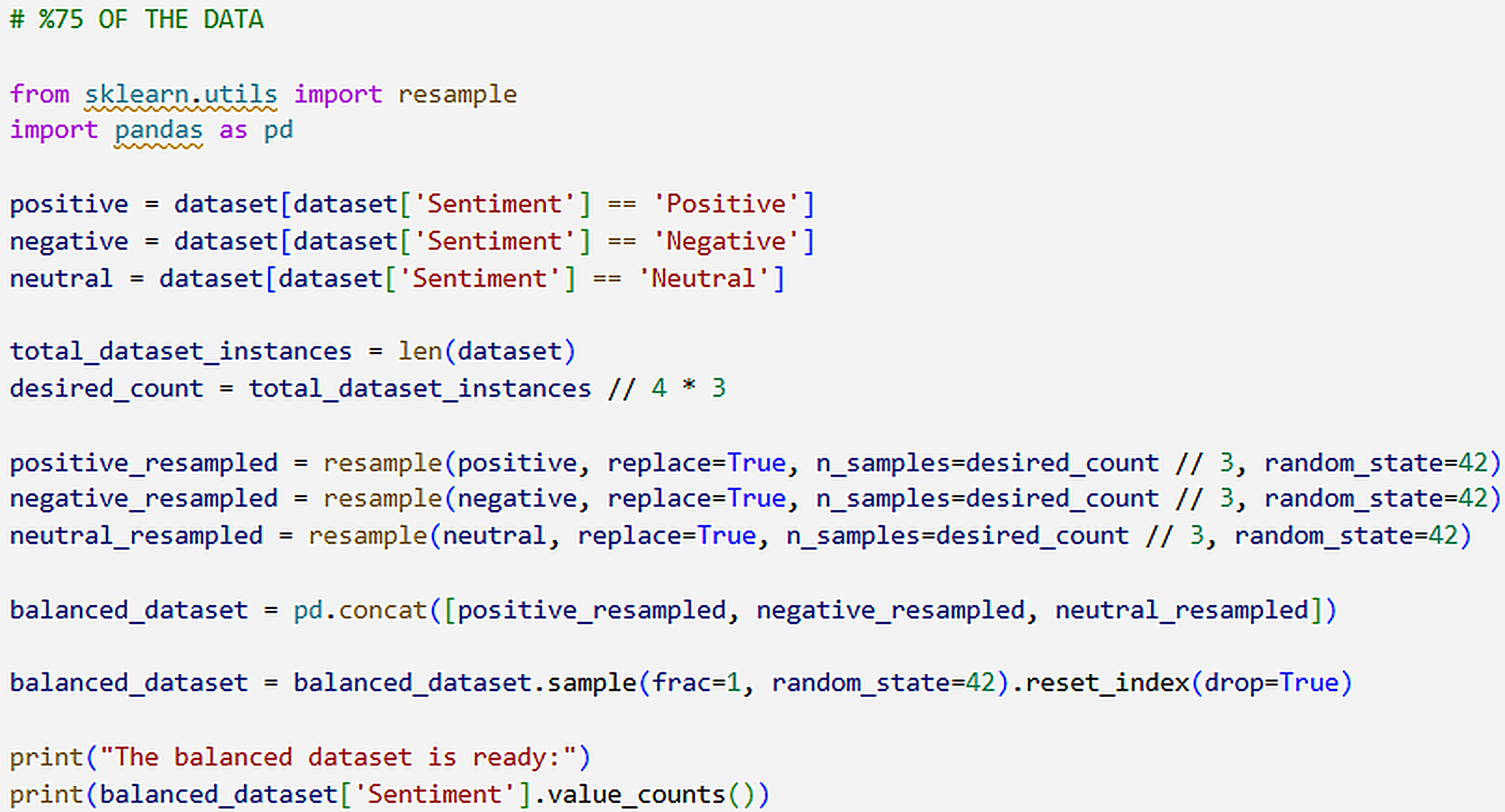}
    \caption{Sampling to \%75 of the Data}
    \label{fig:placeholder}
\end{figure}

\subsubsection{Comparing Various Amount of the Dataset for Accuracy}\label{AA}

Oversampling was applied to balance the unbalanced dataset. In case the oversampling is not done correctly, various problems may arise, such as overfitting, noise amplification, biased and skewed models, unrealistic data, computational problems, or misleading model evaluation, which negatively affect the performance of the model and the reliability of the results. Therefore, testing the oversampled data is necessary to make sure the data has been trained correctly. To achieve this, the model was tested on undersampled subsets of the dataset, including 25\%, 50\%, and 75\% of the data. This approach allows for a more thorough evaluation of whether the oversampling process preserved the integrity of the data and did not artificially inflate performance.

\begin{table} 
\renewcommand{\arraystretch}{1.3} 
\begin{center} 
\resizebox{0.75\linewidth}{!}{ 
\begin{tabular}{|c|c|c|c|c|}
\hline
\diagbox{Dataset Sizes}{Sentiments} & \textbf{\textit{Positive}} & \textbf{\textit{Neutral}} & \textbf{\textit{Negative}} & \textbf{\textit{Total}} \\ 
\hline
Imbalanced Dataset & 356,823 & 32,312 & 6,012 & 395,147 \\ 
\hline
\%25 of the Dataset & 32,928 & 32,928 & 32,928 & 98,784 \\ 
\hline
\%50 of the Dataset & 65,857 & 65,857 & 65,857 & 197,571 \\ 
\hline
\%75 of the Dataset & 98,786 & 98,786 & 98,786 & 296,358 \\ 
\hline
\%100 of the Dataset & 131,715 & 131,715 & 131,715 & 395,145 \\ 
\hline
Oversampled Dataset & 356,823 & 356,823 & 356,823 & 1,070,469 \\ 
\hline
\end{tabular}
}
\label{tab1}
\end{center}
\vspace{-0.25cm}
\caption{Various Percentages of the Dataset}
\end{table}

In summary, consistent model performance across undersampled datasets validates that the oversampling process was done correctly and did not distort the dataset's integrity or artificially enhance accuracy. These tests suggest the oversampling process preserved the integrity of the data without introducing the possible issues that might have occurred in the oversampled dataset.

 \section{Machine Learning Models}
In this study, five machine learning models -RFC, LRC, BNB, L-SVC, MNB- were applied to evaluate the performance of sentiment analysis on the Amazon customer review dataset. Two feature extraction techniques, BoW and TF-IDF, were utilized to transform the textual data into numerical representations suitable for model training and evaluation. The analysis focuses on a comparative assessment of the top three models for each feature extraction method, as well as a comprehensive comparison of the five models and the two feature extraction approaches. The evaluation metrics - accuracy, precision, recall and F1 score - were calculated to measure the overall performance of each model. Furthermore, the metrics were further analyzed at a class level (positive, negative, and neutral) for every model and feature extraction method, providing a detailed insight into their effectiveness across different sentiment categories. This multidimensional evaluation facilitates a strong understanding of the interaction between model architectures and feature extraction techniques in sentiment classification tasks. The overall operational workflow is illustrated in Figure 10,

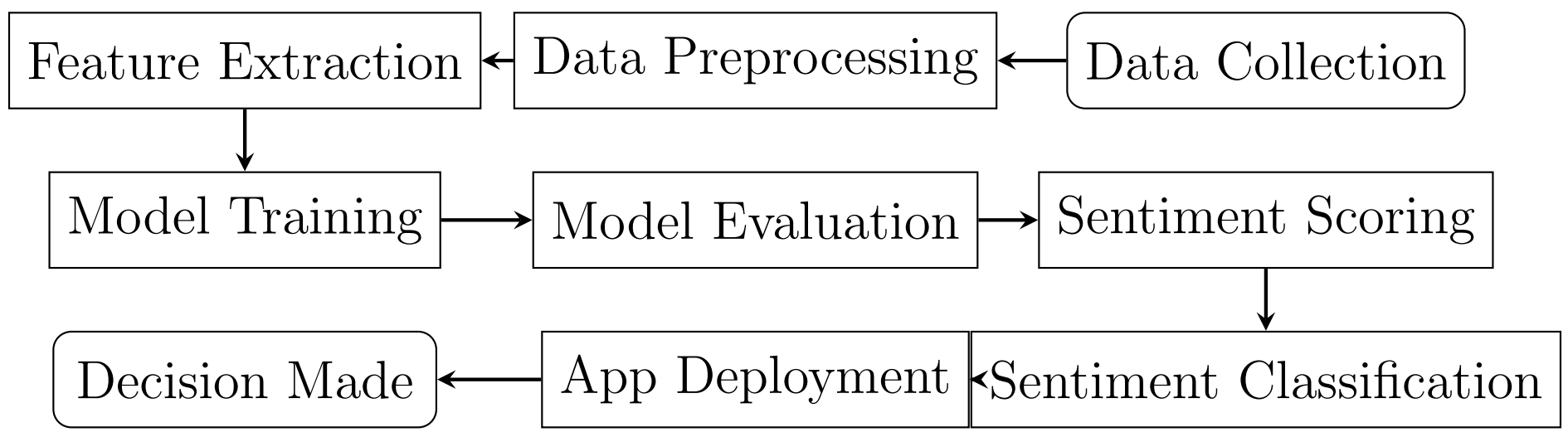
\begin{figure}[h]
    \centering
    \resizebox{0.75\textwidth}{!}{%
        \begin{tikzpicture}[node distance=2cm]

            \tikzstyle{startstop} = [rectangle, rounded corners, minimum width=3cm, minimum height=0.75cm, text centered, draw=black]
            \tikzstyle{process} = [rectangle, minimum width=3cm, minimum height=0.75cm, text centered, draw=black]
            \tikzstyle{arrow} = [thick,->,>=stealth]

            \node (collection) [startstop] {Data Collection};
            \node (preprocessing) [process, left of=collection, xshift=-2cm] {Data Preprocessing};
            \node (feature) [process, left of=preprocessing, xshift=-2cm] {Feature Extraction};
            \node (training) [process, below of=feature, yshift=0.75cm] {Model Training};
            \node (evaluation) [process, right of=training, xshift=2cm] {Model Evaluation};
            \node (score) [process, right of=evaluation, xshift=2cm] {Sentiment Scoring};
            \node (classification) [process, below of=score, yshift=0.75cm] {Sentiment Classification};
            \node (deployment) [process, left of=classification, xshift=-2cm] {App Deployment};
            \node (decision) [startstop, left of=deployment, xshift=-2cm] {Decision Made};

            \draw [arrow] (collection.west) -- (preprocessing.east);
            \draw [arrow] (preprocessing.west) -- (feature.east);
            \draw [arrow] (feature) -- (training);
            \draw [arrow] (training) -- (evaluation);
            \draw [arrow] (evaluation) -- (score);
            \draw [arrow] (score) -- (classification);
            \draw [arrow] (classification) -- (deployment.east);
            \draw [arrow] (deployment.west) -- (decision.east);

        \end{tikzpicture}%
    }
    \caption{Operational Flow Overview for Machine Learning}
    
    \label{fig:operational_flow}
\end{figure}

\subsection{General Aspects of Machine Learning}\label{AA}

\subsubsection{Feature Extraction}\label{AA}
The evaluation considered both overall performance and class-level metrics, allowing for a comprehensive comparison of the top three performing models for each feature extraction method, as well as a broader assessment of all five models across both techniques. Metrics including accuracy, precision, recall, and F1 score were calculated to quantify model effectiveness, providing a multidimensional understanding of how model architectures interact with feature representations in sentiment analysis tasks.

\begin{figure}
    \centering
    \includegraphics[width=0.5\linewidth]{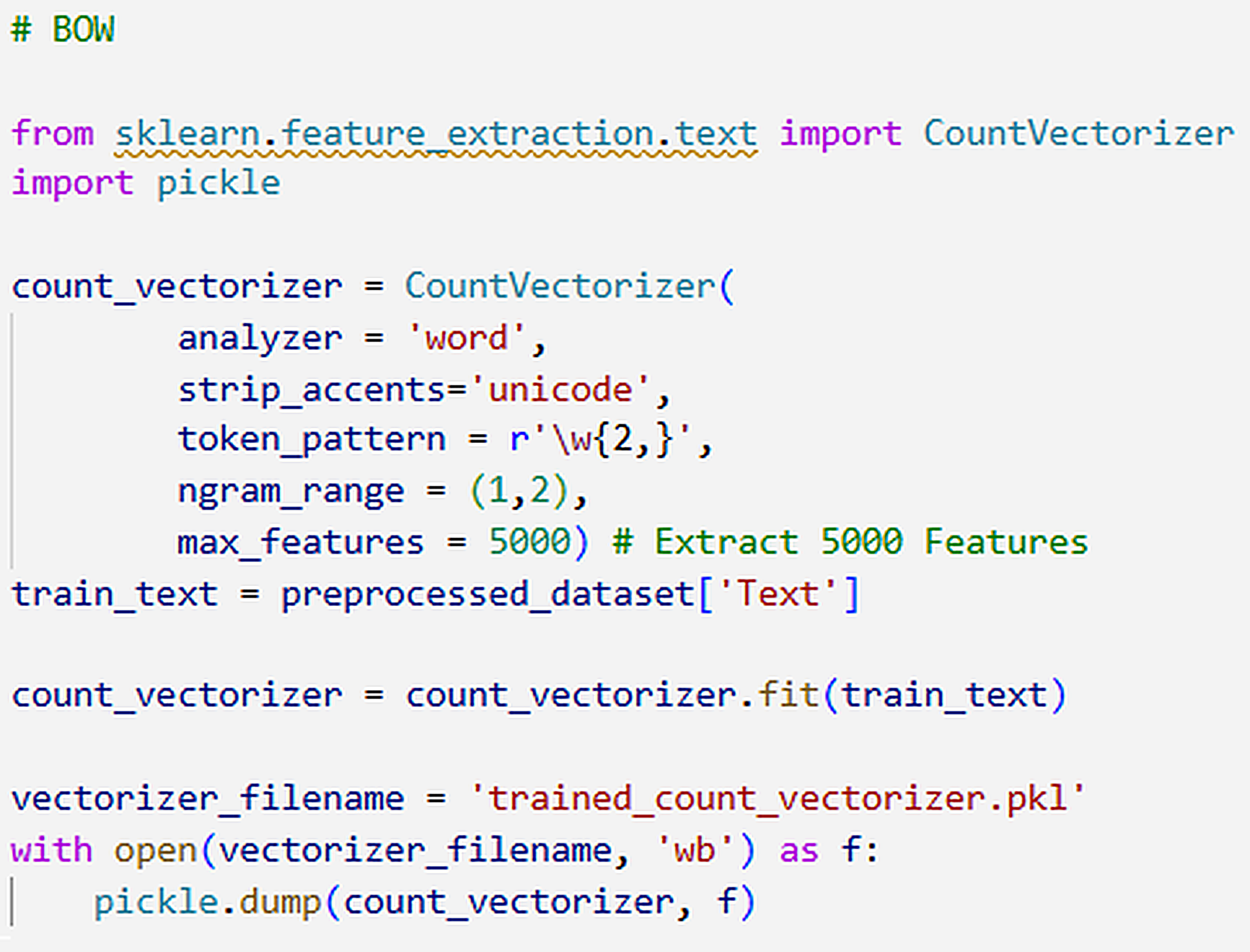}
    \caption{BoW Feature Extraction Coding}
    \label{fig:placeholder}
\end{figure}

For BoW, scikit-learn’s CountVectorizer was employed to convert textual reviews into numerical vectors. The vectorizer preprocesses the text by tokenizing into words, removing single-character tokens, normalizing accented characters, and extracting both unigrams and bigrams (n-gram range = 1–2). To maintain computational efficiency and reduce sparsity, the feature set was limited to the 5000 most frequent tokens. The selection of 5000 features strikes a balance between capturing sufficient vocabulary for meaningful sentiment analysis and ensuring manageable computational complexity. The use of unigrams and bigrams allows the model to capture both individual word contributions and common two-word phrases, which improves the representation of context in reviews. The resulting BoW representations served as input to the machine learning models. To ensure reproducibility, the trained BoW vectorizer was saved using Python’s pickle module, enabling consistent feature extraction for future experiments and applications.(Figure 11)

\begin{figure}
    \centering
    \includegraphics[width=0.5\linewidth]{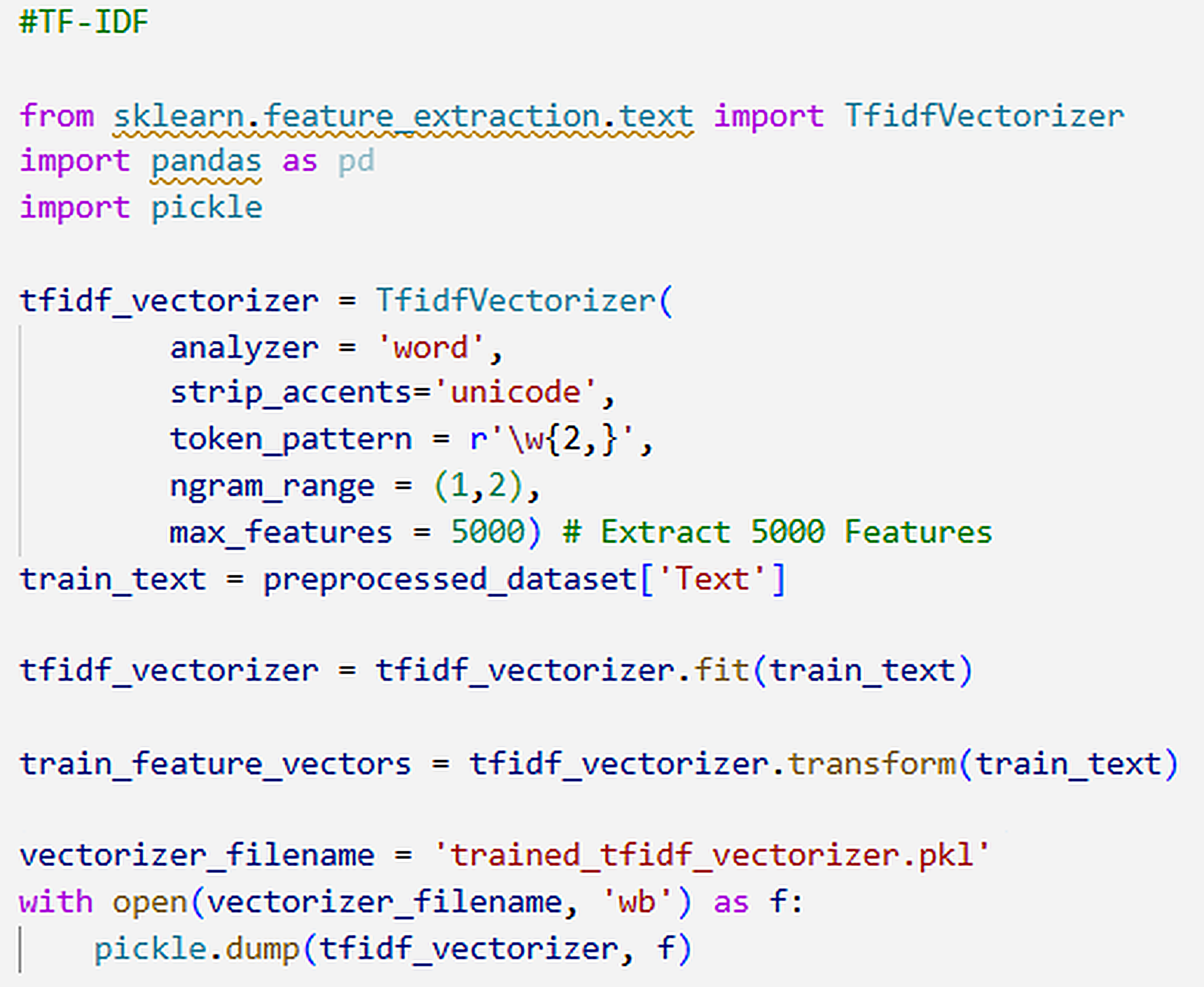}
    \caption{TF-IDF Feature Extraction Coding}
    \label{fig:placeholder}
\end{figure}

For TF-IDF, scikit-learn’s TfidfVectorizer was applied to account for the relative importance of terms across the corpus. Similar preprocessing steps were applied, including tokenization, removal of single-character tokens, accent normalization, and extraction of unigrams and bigrams. The 5000 most frequent terms were selected to form the feature set, and term weights were computed based on their TF-IDF values, capturing both term frequency and inverse document frequency information. This approach ensures that common but less informative words are weighted lower, while rare but sentiment-relevant words receive higher importance. The resulting TF-IDF weighted vectors were then used as input for the same set of machine learning models. The trained TF-IDF vectorizer was also saved for reproducibility, ensuring that future studies or real-time applications can employ the exact same feature mapping.(Figure 12)

This dual-feature extraction strategy enabled a thorough investigation of the comparative strengths of BoW and TF-IDF representations, providing insights into their impact on model performance across different sentiment categories and guiding the selection of the most effective combination of model and feature representation for large-scale sentiment analysis.

\subsubsection{Word Cloud}\label{AA}

The Word Cloud method is a visual representation of textual data where the frequency or importance of words within a dataset is depicted using varying font sizes. Larger words in the word cloud indicate higher frequency or relevance, while smaller words represent less frequent terms. The coding structure is illustrated in Figure 13, and the resulting word clouds are shown in Figures 14–15-16. In the context of a large-scale sentiment analysis project on the Amazon customer review dataset, this method plays a crucial role in providing an intuitive and immediate understanding of the most frequently used terms across millions of reviews. It helps identify prominent themes, recurring terms, and potential sentiment-indicative keywords, such as “excellent,'' “disappointed,'' “fast,'' or “poor,'' which may otherwise be overlooked in raw data. This is particularly useful in studies dealing with massive datasets, as word clouds offer a quick way to visualize the general sentiment and focus of the reviews. Moreover, they assist in uncovering key patterns and variations in customer sentiment, providing insights that can guide further preprocessing steps, feature engineering, and model development. By revealing the linguistic trends and customer priorities embedded in the data, word clouds serve as a valuable exploratory tool to complement deeper analytical methodologies.

In this study, we used the Word Cloud method for every class separately as positive, negative, and neutral, as shown in Figures 14, 15, and 16. Furthermore, we have used such methods as top-used words and weights for each word to describe the aspects of the project, enhance the analysis for the dataset, and explore new features to serve the sentiment decisions. Methods that are mentioned will be shown in the next sections.

\begin{figure}
    \centering
    \includegraphics[width=0.5\linewidth]{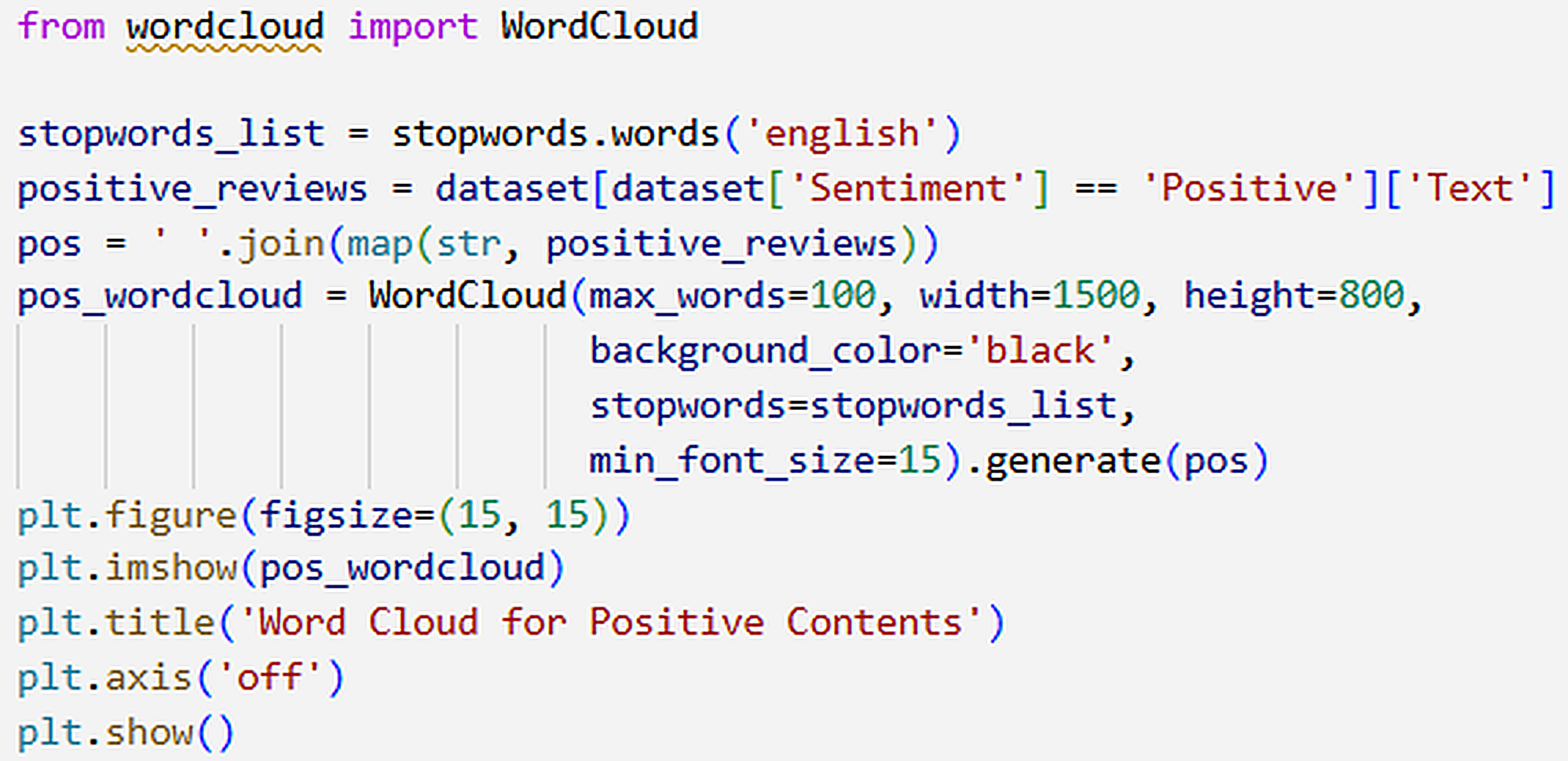}
    \caption{Word Cloud Coding}
    \label{fig:placeholder}
\end{figure}

\begin{figure}[h]
    \centering
    \begin{minipage}{0.45\linewidth}
        \centering
        \includegraphics[width=\linewidth]{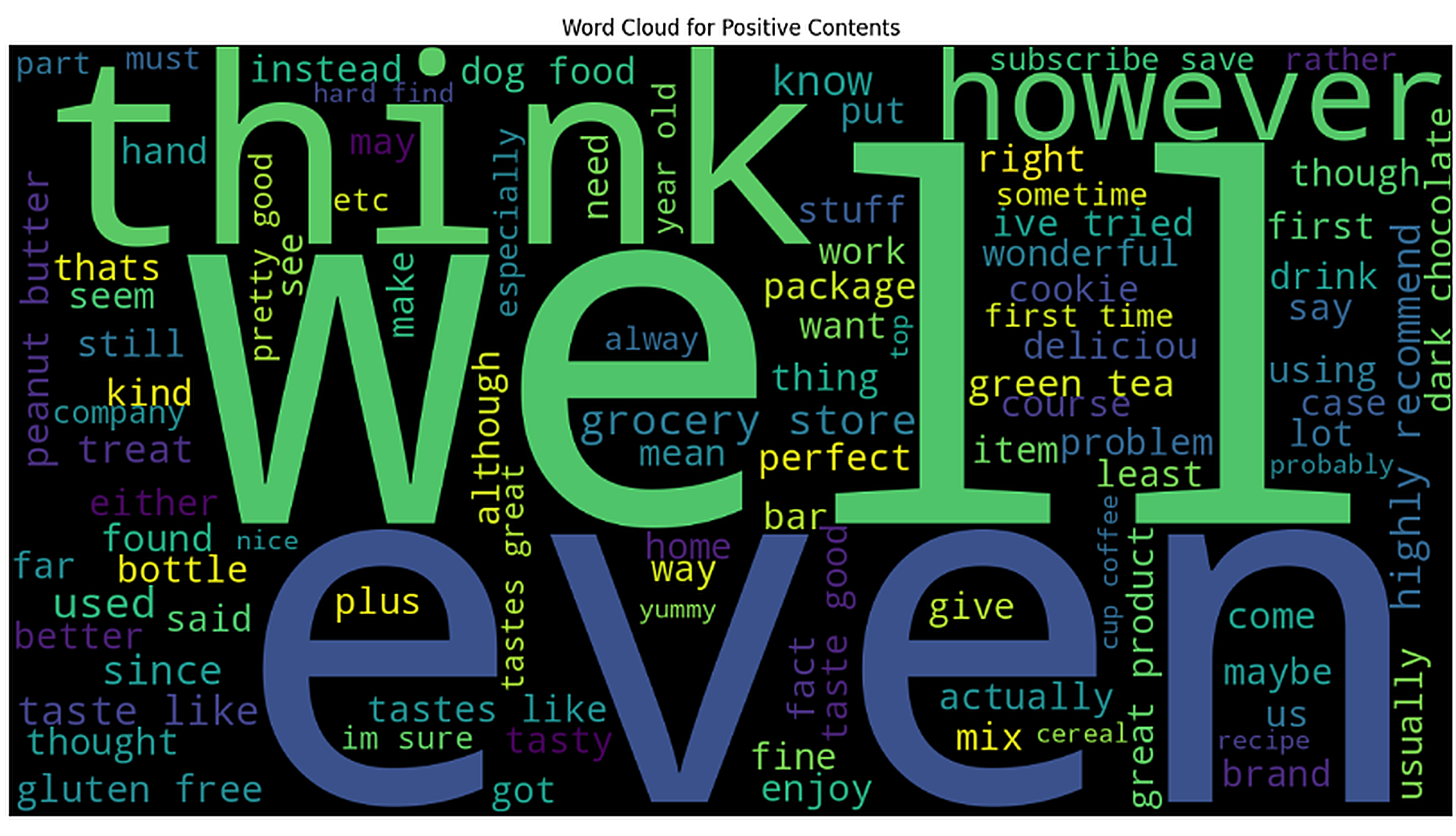}
        \caption{Positive Sentiment Word Cloud}
        \label{fig:positive_wordcloud}
    \end{minipage}
    \hfill
    \begin{minipage}{0.45\linewidth}
        \centering
        \includegraphics[width=\linewidth]{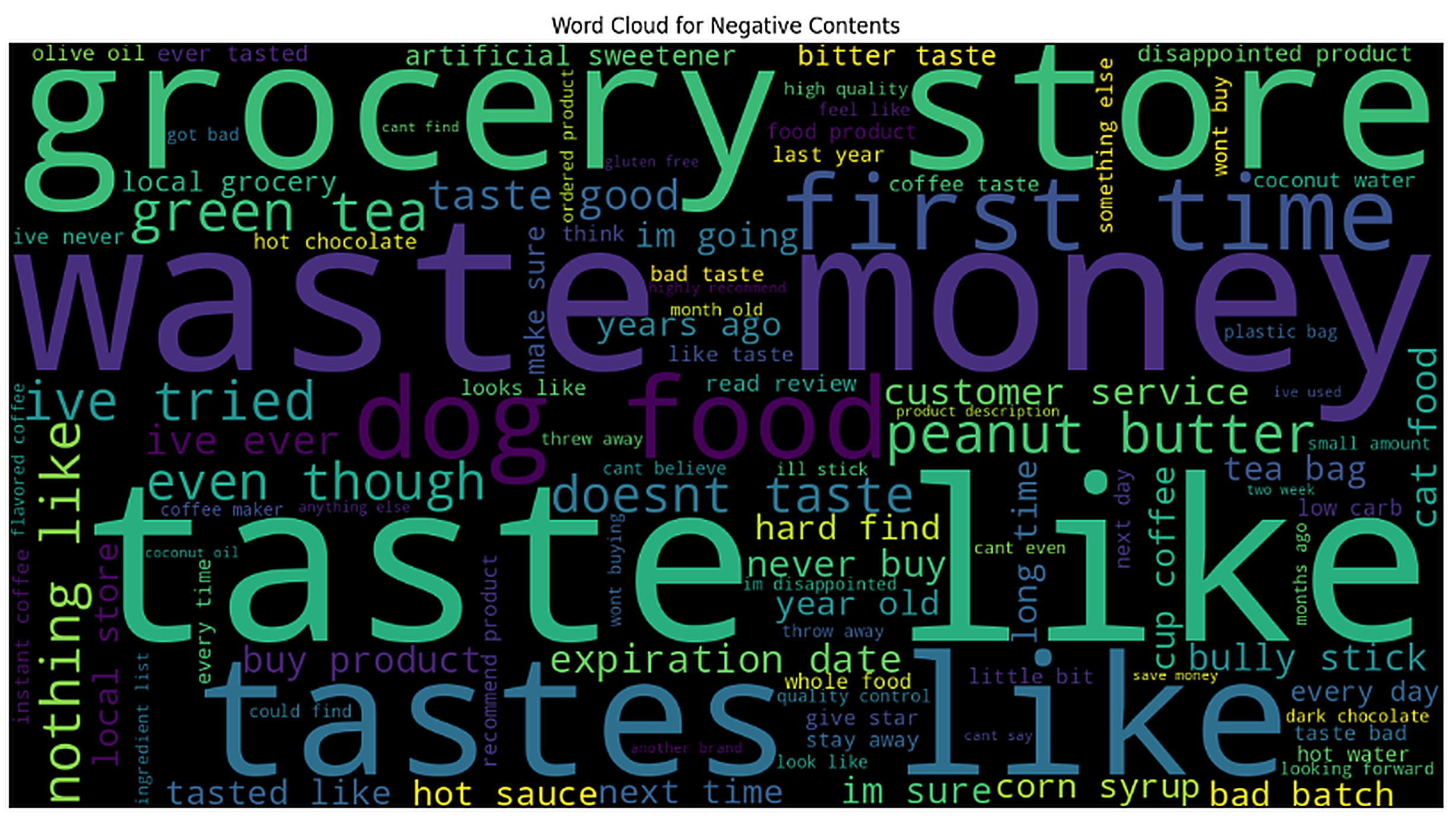}
        \caption{Negative Sentiment Word Cloud}
        \label{fig:oversampled}
    \end{minipage}
\end{figure}

\begin{figure}[h] 
\centerline{\includegraphics[width=0.45\linewidth]{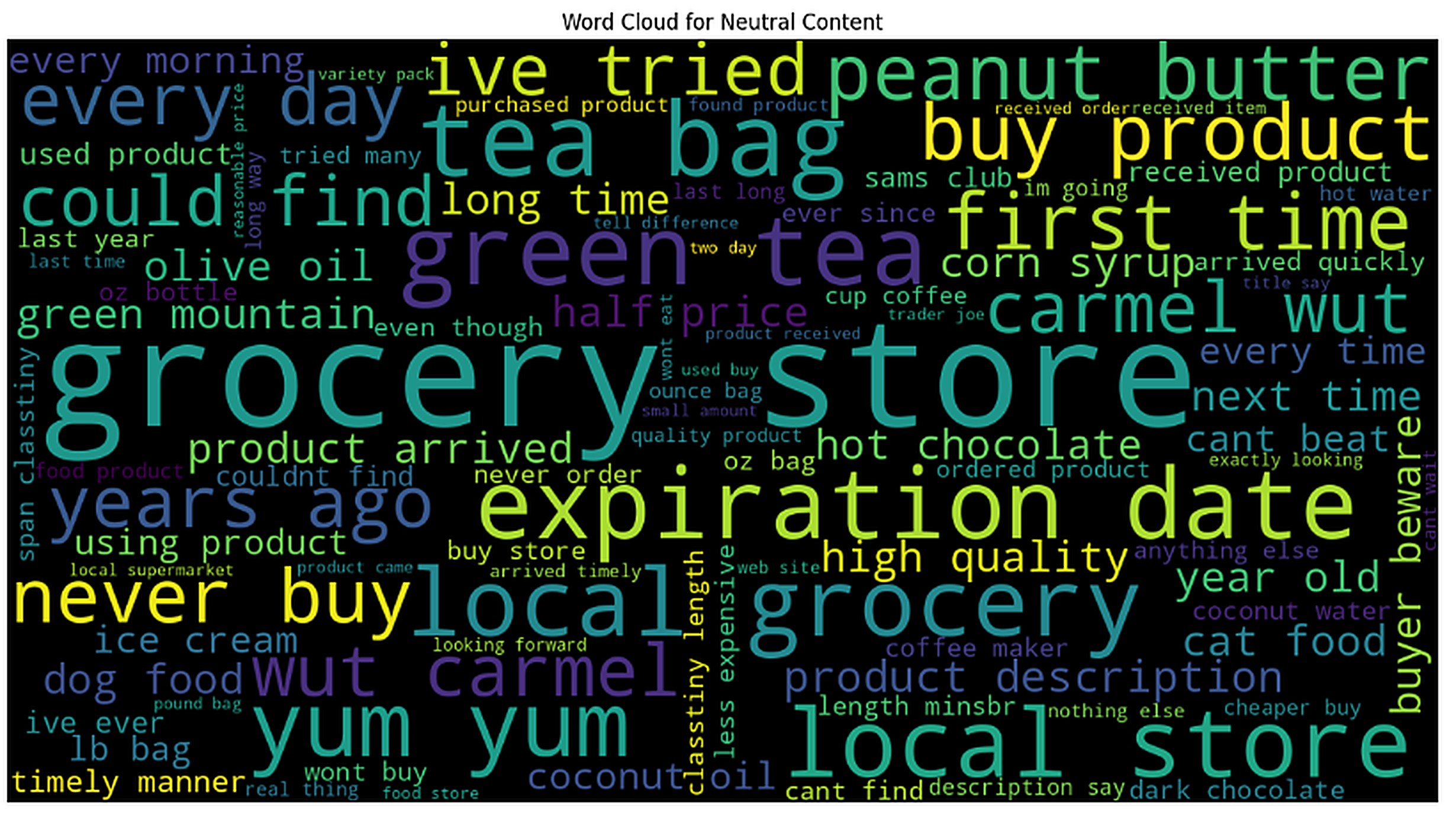}}
\caption{Neutral Sentiment Word Cloud}
\label{fig:neutral_wordcloud}
\end{figure}

\subsection{Confusion Matrix}\label{AA}

The Confusion Matrix, is a crucial evaluation tool for classification tasks, providing a detailed breakdown of correct and incorrect predictions across all classes. For a 3-class sentiment analysis (positive, neutral, negative), it organizes true labels as rows and predicted labels as columns in a 3x3 grid. Diagonal cells represent correct classifications, while off-diagonal cells indicate misclassifications, such as labeling a positive review as neutral.

This method is essential for identifying model weaknesses and calculating metrics like precision, recall, and F1-score for each class. It offers deeper insights into performance, helping refine the model and address challenges like overlapping features or class imbalance. Figure 17 displays a sample confusion matrix and Figure 18 contains the coding, further details will be provided in the subsequent sections.

\begin{figure}[h]
    \centering
    \includegraphics[width=0.45\linewidth]{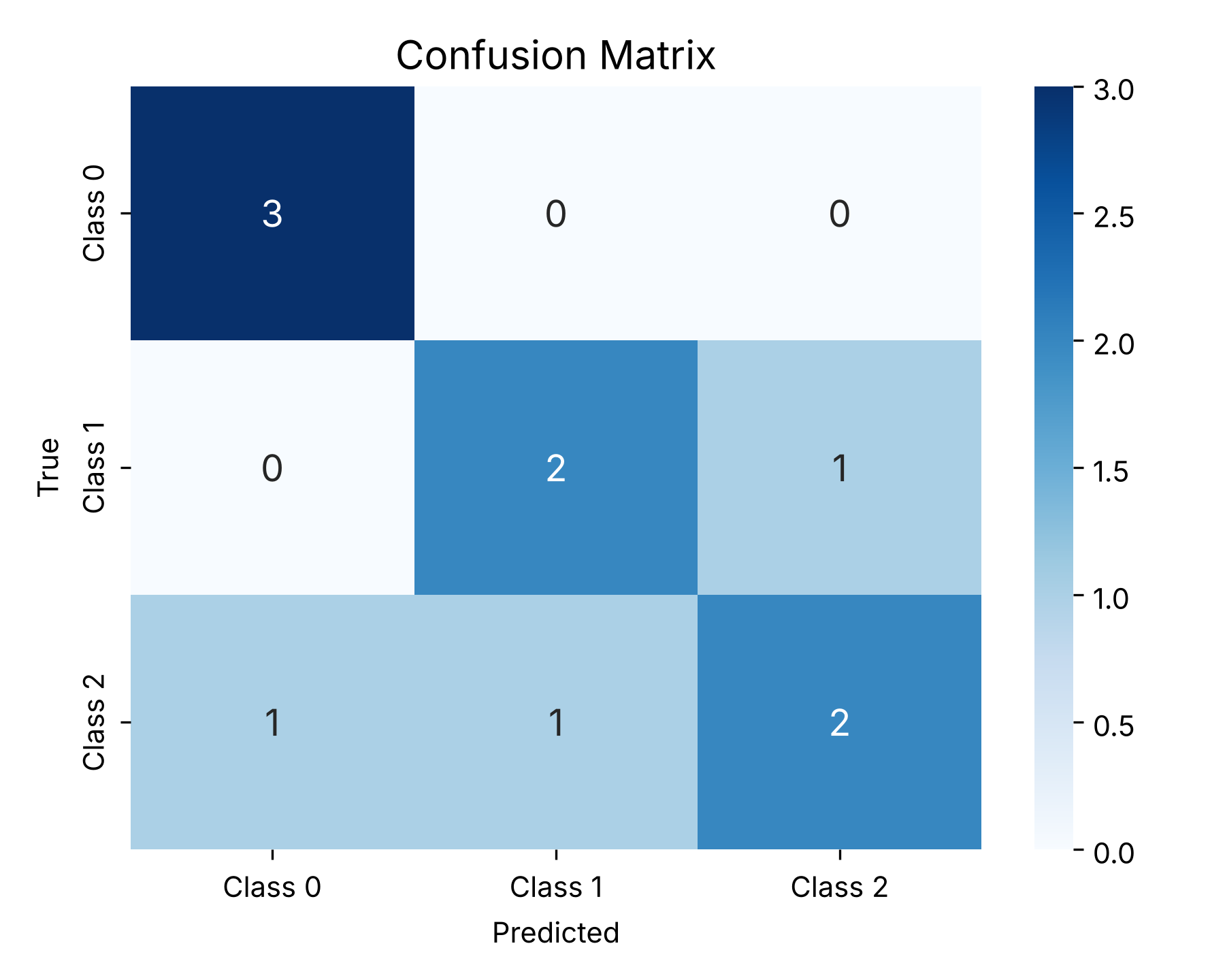}
    \caption{Sample Confusion Matrix}
    \label{fig:enter-label}
\end{figure}

\begin{figure}[h]
    \centering
    \includegraphics[width=0.6\linewidth]{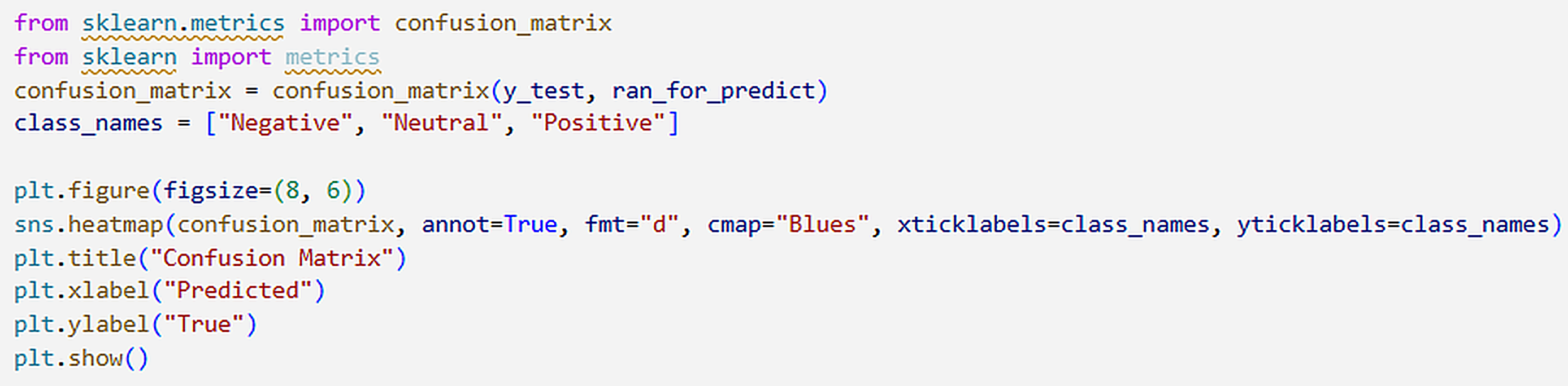}
    \caption{Confusion Matrix Coding}
    \label{fig:placeholder}
\end{figure}

\begin{table}[h]
\caption{Parameters of the machine learning models. Unless stated, values are scikit-learn defaults.}
\label{tab:ml_params}
\centering
\footnotesize
\begin{tabular}{llllll}
\toprule
\textbf{Parameter} & \textbf{RFC} & \textbf{LRC} & \textbf{L-SVC} & \textbf{BNB} & \textbf{MNB} \\
\midrule
Implementation & RandomForestClassifier & LogisticRegression & LinearSVC & BernoulliNB & MultinomialNB \\
Number of trees & \texttt{n\_estimators}=5 & -- & -- & -- & -- \\
Max iterations & -- & \texttt{max\_iter}=100 & \texttt{max\_iter}=1000 & -- & -- \\
Regularization & -- & L2, $C$=1.0 & L2, $C$=1.0 & $\alpha$=1.0 & $\alpha$=1.0 \\
Loss / solver & Gini impurity & lbfgs & squared hinge & -- & -- \\
Class prior & -- & -- & -- & training-split proportions & training-split proportions \\
Binarization & -- & -- & -- & \texttt{binarize}=0.0 & -- \\
Fit prior & -- & -- & -- & True & True \\
Class weights (oversampled set) & none & none & none & none & none \\
Class weights (imbalanced set) & balanced & balanced & balanced & none & none \\
Random seed & \texttt{random\_state}=0 & -- & -- & -- & -- \\
Feature selection per split & $\sqrt{M}$ & -- & -- & -- & -- \\
Max depth & None & -- & -- & -- & -- \\
\bottomrule
\end{tabular}

\vspace{0.5em}
\begin{minipage}{\linewidth}
\footnotesize
\textit{Feature extraction.} BoW: \texttt{CountVectorizer}, unigrams and bigrams (\texttt{ngram\_range}=(1,2)), \texttt{token\_pattern}=\texttt{\textbackslash w\{2,\}}, \texttt{strip\_accents}=unicode, \texttt{max\_features}=5000. TF-IDF: \texttt{TfidfVectorizer} with the same settings; \texttt{max\_features}=5000 in the oversampled run and 1000 in the imbalanced run.\\
\textit{Data split.} 80/20 train/test split, \texttt{random\_state}=2020, not stratified.
\end{minipage}
\end{table}

\subsection{Presentation of Machine Learning Models}
Five machine learning models-RFC, LRC, BNB, L-SVC, and MNB-were implemented to classify sentiment within the dataset. To prepare the textual data for these models, two feature extraction methods, TF-IDF and BoW, were employed. Analysis of the results showed that the BoW method provided superior performance compared to TF-IDF. As a result, the primary focus is on comparing the first three models trained using the BoW method. For improved visualization, the model metrics are presented in Figure 19.

Performance evaluation is conducted using confusion matrices and detailed metrics, including precision, recall, F1 score, and accuracy, for each sentiment class. This approach ensures a comprehensive comparison of the models and their ability to classify sentiments effectively. The best-performing model is then highlighted, and its results are analyzed to provide insights into its strengths. Additionally, the comparison between the feature extraction methods will be elaborated in the next section to offer a broader perspective on their influence on model performance.

\begin{figure}
    \centering
    \includegraphics[width=0.75 \linewidth]{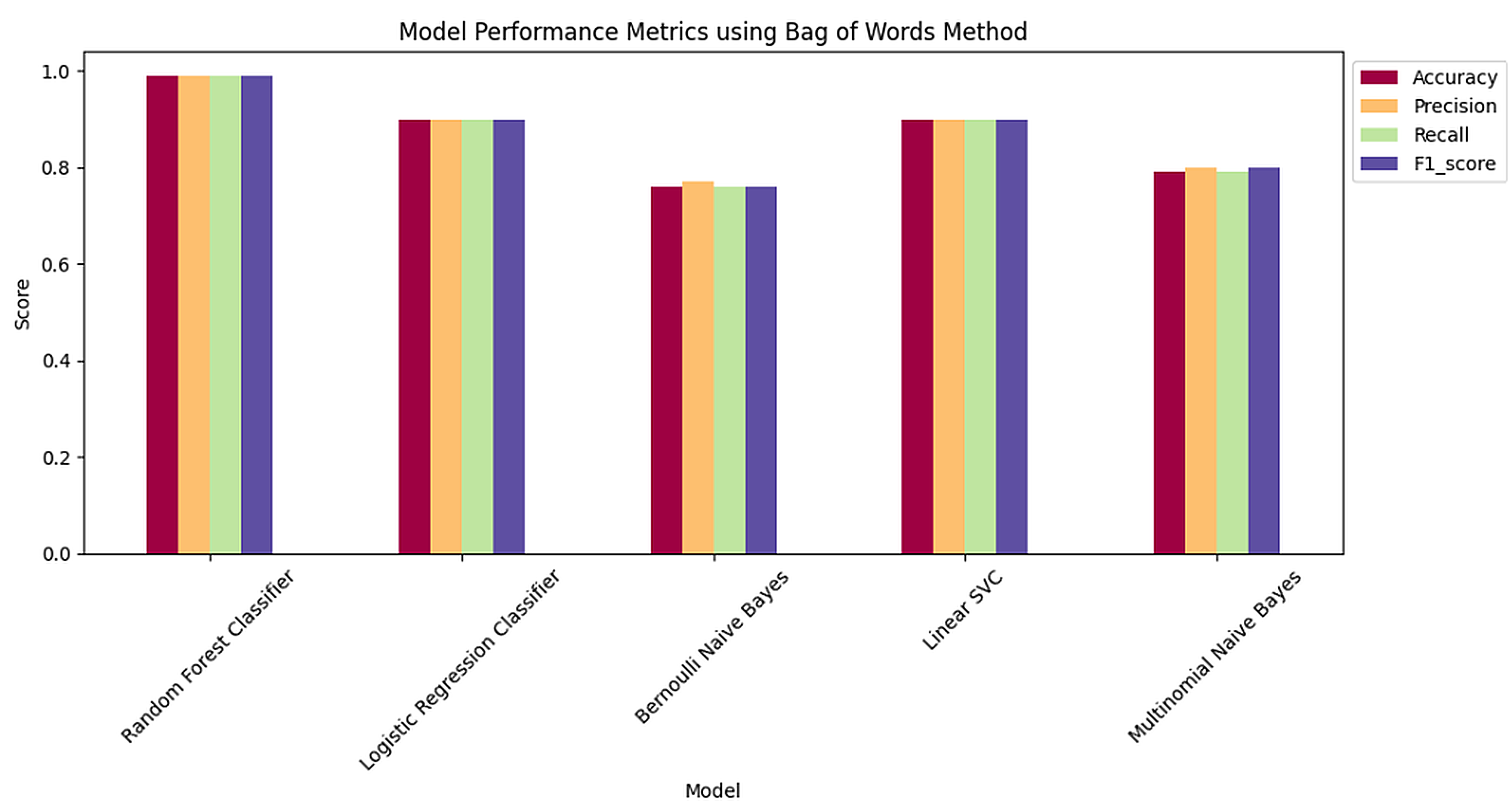}
    \caption{Model Metrics with BoW}
    \label{fig:enter-label}
\end{figure}

\subsubsection{Random Forest Classifier (RFC)}

The RFC is an ensemble machine learning algorithm designed to enhance classification performance by constructing multiple decision trees during training and aggregating their predictions. This approach reduces overfitting and improves model robustness by combining predictions from various trees, each trained on random subsets of features and data. Its ability to handle high-dimensional data, manage noisy features, and accommodate missing values makes it a widely adopted method for complex classification tasks, such as sentiment analysis.

\textbf{Bootstrap Sampling (Random Data Selection):}
Each of the 5 trees is trained on a random sample of the training data with replacement. This introduces variability and reduces overfitting.

\textbf{Random Feature Selection at Each Split:}
\begin{equation}
    m = \sqrt{M}
\end{equation}
where $m$ is the number of randomly selected features out of $M$ total features.

\textbf{Gini Impurity (Node Splitting):}
\begin{equation}
    G(t) = 1 - \sum_{k=1}^{K} p_k^2
\end{equation}
where $p_k$ is the proportion of samples belonging to class $k$ at node $t$.

\textbf{Majority Voting (Final Prediction):}
\begin{equation}
    \hat{y} = \text{mode}\left( \sum_{i=1}^{N} h_i(x) \right)
\end{equation}
where $h_i(x)$ is the prediction from the $i$-th decision tree and $N$ is the total number of trees. \\
$\hat{y}$ is the \textbf{predicted class label}.

\vspace{0.5 cm}

In this study, the RFC emerged as the most successful and accurate model, utilizing the BoW feature extraction method. The results from applying this model to the Amazon review dataset demonstrate its effectiveness, achieving an overall accuracy of 99\%. The model performed exceptionally across all sentiment classes, with precision, recall, and F1-scores exceeding 0.96 for each class. Analysis of the confusion matrix reveals perfect classification of neutral sentiments (71,300 instances) and minimal misclassifications for negative and positive sentiments. The parameter “support'' in parameter tables, as Figure 20, represents the total number of instances in the dataset that belong to each sentiment class and all dataset. Out of 71,628 negative samples, only 22 were misclassified, while 2,718 positive samples were incorrectly identified as negative, and 206 as neutral (Figure 21). These results highlight the model’s strength in accurately classifying neutral and negative sentiments, while minor misclassifications in positive sentiments suggest slight feature overlaps. Overall, the RFC proved highly reliable, demonstrating its capacity for large-scale sentiment classification with complex datasets.

\begin{figure}[h]
    \centering
    \begin{minipage}{0.45\linewidth}
        \centering
        \includegraphics[width=\linewidth]{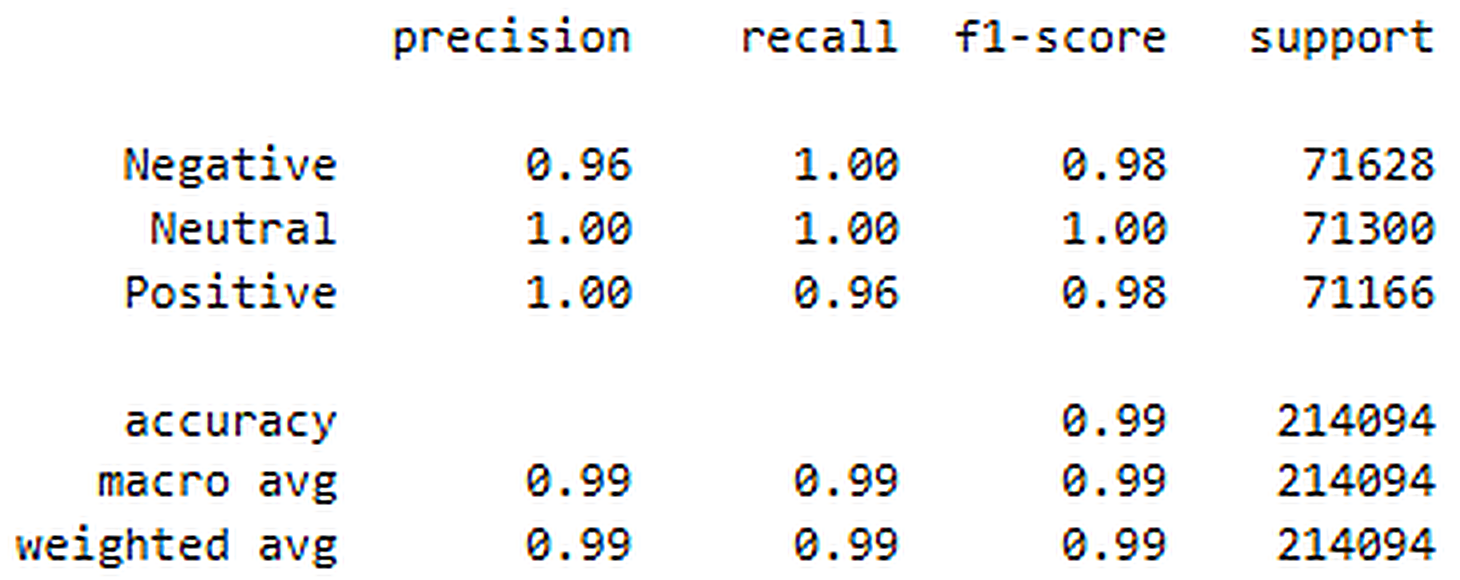}
        \caption{Parameters for Each Class in RFC}
        \label{fig:enter-label}
    \end{minipage}
    \hfill
    \begin{minipage}{0.45\linewidth}
        \centering
        \includegraphics[width=\linewidth]{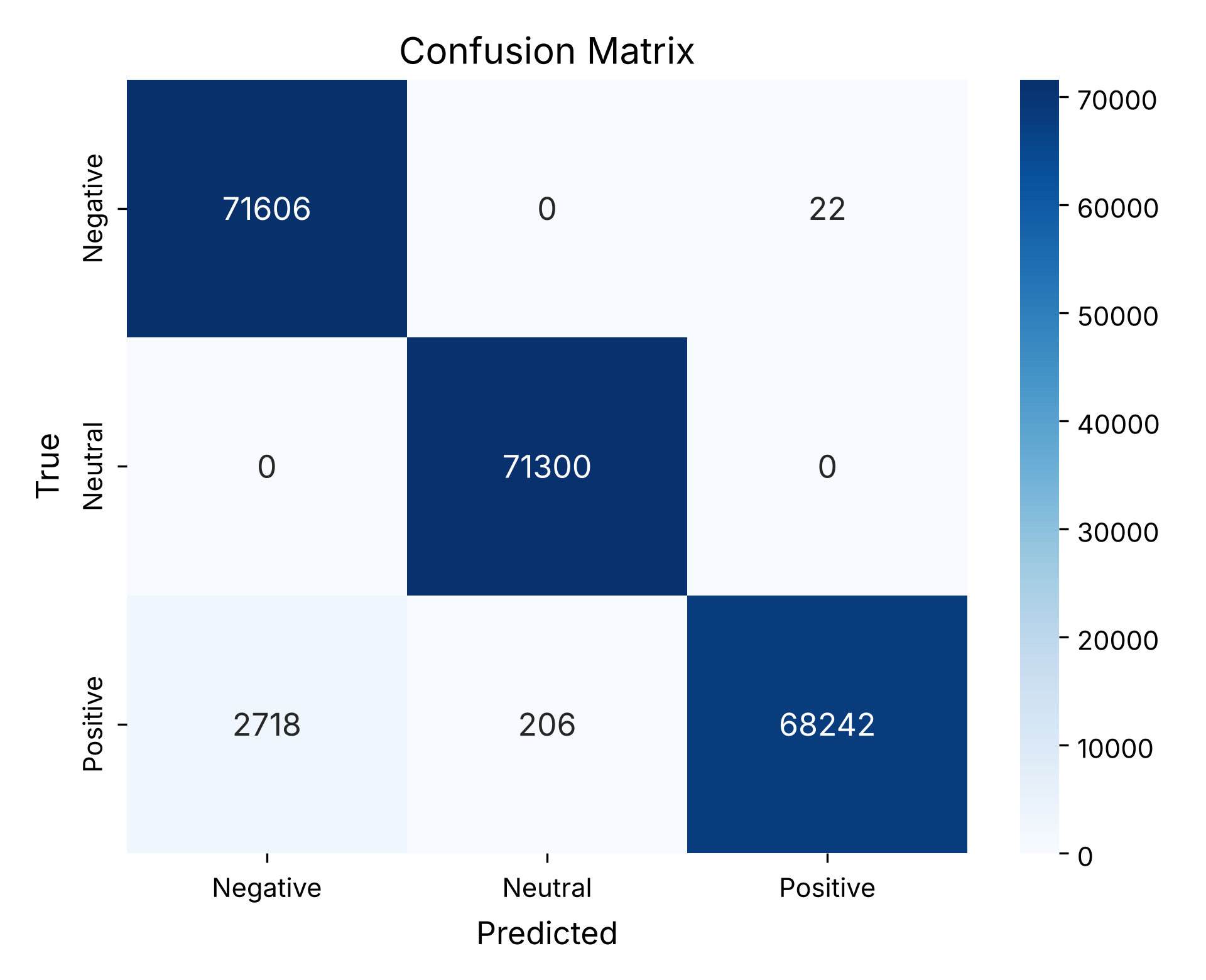}
        \caption{RFC - Confusion Matrix}
        \label{fig:enter-label}
    \end{minipage}
\end{figure}

\subsubsection{Logistic Regression Classifier (LRC)}
The LRC is a widely used algorithm in machine learning that is particularly effective for binary and multiclass classification tasks. It models the probability of a class by applying the logistic function to a linear combination of input features, ensuring the output probabilities remain between 0 and 1. This method is computationally efficient and interpretable, making it well-suited for large-scale datasets and sentiment analysis applications. Furthermore, LRC can handle high-dimensional data and produce probabilistic outputs, which are advantageous for understanding the confidence levels of predictions.

\textbf{Model Initialization:} \\
The Logistic Regression model starts by initializing the weight vector $\mathbf{w}$ and bias $b$.

\textbf{Model Training:} \\
During training, the model learns the best weights $\mathbf{w}$ and bias $b$ that minimize the \textbf{Cross-Entropy Loss} function:
\begin{equation}
    \mathbf{w} \leftarrow \mathbf{w} - \eta \nabla L(\mathbf{w})
\end{equation}
where $\eta$ is the learning rate (step size), and $\nabla L(\mathbf{w})$ is the gradient of the loss function, which indicates the direction to adjust the weights to reduce the loss.

\textbf{Prediction Using Sigmoid Function:} \\
After training, the model uses the \textbf{Sigmoid Function} to calculate the probability of a class for a given input $\mathbf{x}$:
\begin{equation}
    P(y=1 \mid \mathbf{x}) = \sigma(\mathbf{w}^\top \mathbf{x} + b) = \frac{1}{1 + e^{-(\mathbf{w}^\top \mathbf{x} + b)}}
\end{equation}
where $\mathbf{w}^\top \mathbf{x} + b$ is the linear combination of input features and their corresponding weights. The output is a probability between 0 and 1.

\textbf{Decision Rule:} \\
The class is predicted based on the probability:
\begin{equation}
    \hat{y} = 
    \begin{cases}
      1, & \text{if } P(y=1 \mid \mathbf{x}) \geq 0.5 \\
      0, & \text{if } P(y=1 \mid \mathbf{x}) < 0.5
    \end{cases}
\end{equation}
where $\hat{y}$ is the \textbf{predicted class label}.

\vspace{0.5 cm}

When applied to the Amazon review dataset, the LRC model achieved an overall accuracy of 90\%. The precision, recall, and F1-scores for negative, neutral, and positive sentiments were 0.88, 0.91, and 0.91, respectively, with the highest performance observed in the neutral class (F1-score of 0.94). The confusion matrix indicates that out of 71,628 negative reviews, 5,721 were misclassified as positive, and 4,577 as neutral. Similarly, among 71,166 positive reviews, 6,487 were misclassified as negative, and 2,052 as neutral, as Figure 22-23. Neutral reviews displayed minimal misclassification, with 2,092 and 145 instances misclassified as negative and positive, respectively. While the model demonstrates robust performance, the misclassifications highlight areas for potential improvement, such as addressing overlapping features or imbalanced class distributions.

\begin{figure}[h]
    \centering
    \begin{minipage}{0.45\linewidth}
        \centering
        \includegraphics[width=\linewidth]{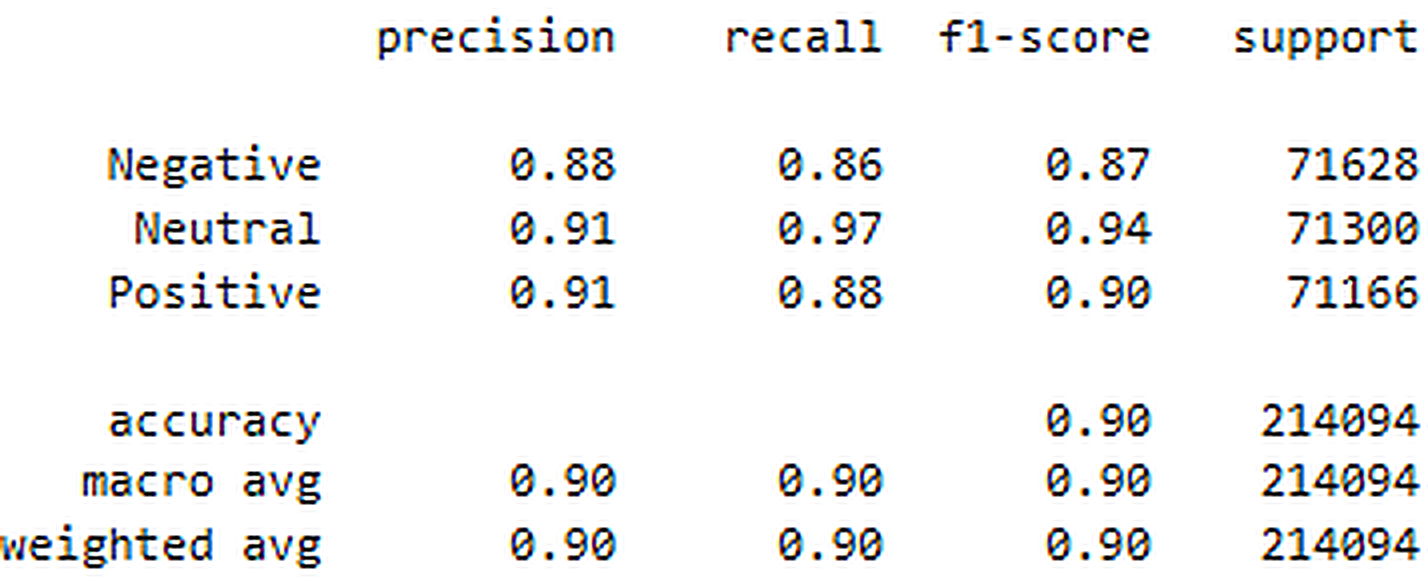}
        \caption{Parameters for Each Class in LRC}
        \label{fig:enter-label}
    \end{minipage}
    \hfill
    \begin{minipage}{0.45\linewidth}
        \centering
        \includegraphics[width=\linewidth]{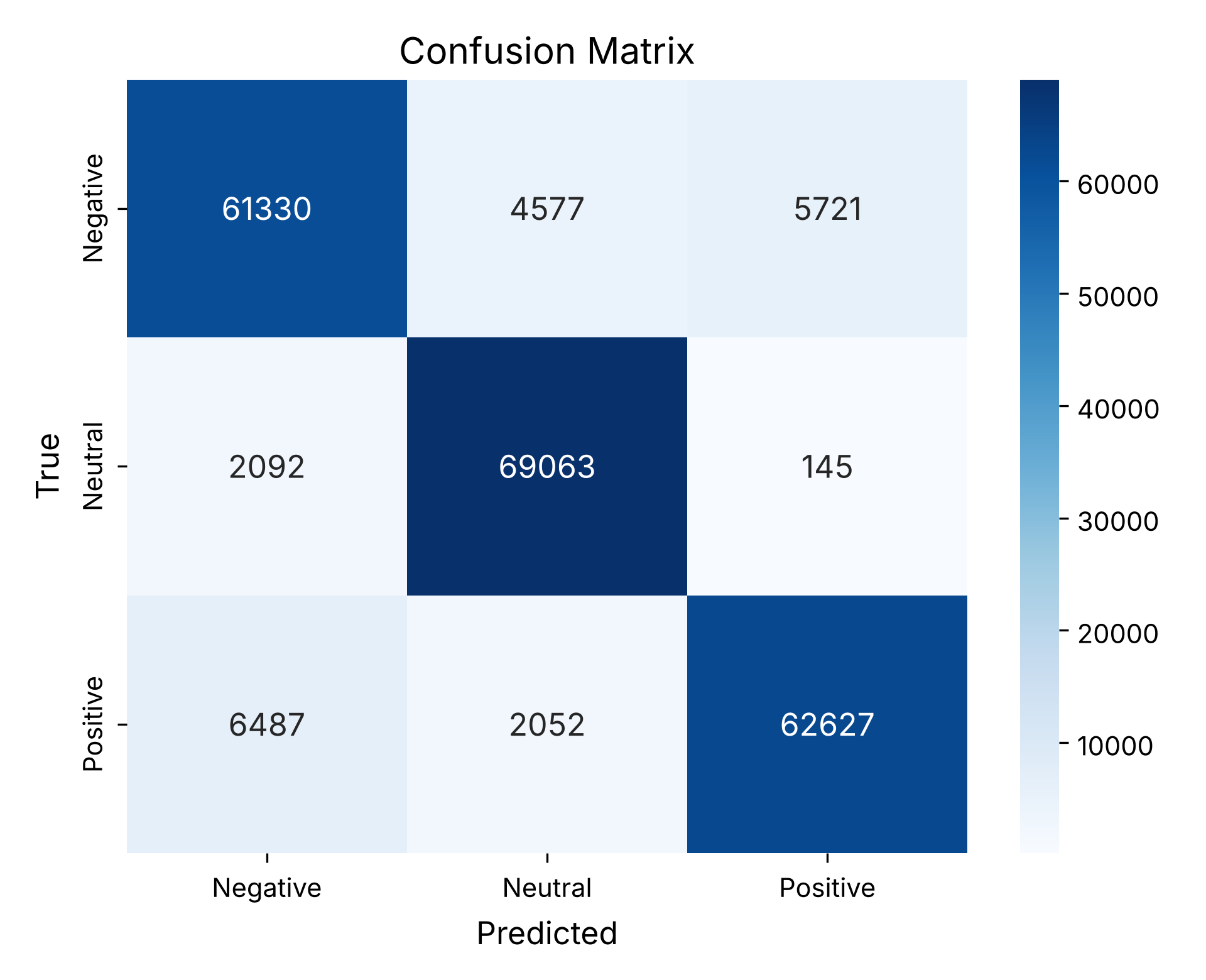}
        \caption{LRC - Confusion Matrix}
        \label{fig:enter-label}
    \end{minipage}
\end{figure}

\subsubsection{Linear Support Vector Classification (L-SVC)}
L-SVC is a machine learning algorithm well-suited for text classification tasks due to its efficiency with high-dimensional datasets. It uses a linear decision boundary to separate classes and maximizes the margin between data points, making it robust and scalable for large datasets. L-SVC is particularly advantageous for sentiment analysis, as it efficiently handles sparse data representations such as BoW and TF-IDF feature extraction methods.

\textbf{Decision Boundary:}  
The decision boundary is the line (or hyperplane) that separates the data into two classes. It is determined by the equation:
\begin{equation}
    f(x) = w^\top x + b
\end{equation}
where \( w \) is the weight vector, \( x \) is the feature vector, and \( b \) is the bias term.  

The class prediction is:
\begin{equation}
    \hat{y} = 
    \begin{cases} 
      1 & \text{if } f(x) > 0,\\
      -1 & \text{if } f(x) < 0
    \end{cases}
\end{equation}

\textbf{Margin Maximization:}  
To improve the classification performance, we maximize the margin between the two classes:
\begin{equation}
    y_i (w^\top x_i + b) \geq 1, \quad \forall i
\end{equation}
The margin is given by:
\begin{equation}
    \text{Margin} = \frac{2}{\|w\|}
\end{equation}

\textbf{Hinge Loss Function:}  
The hinge loss function penalizes misclassified points or points close to the margin. It is defined as:
\begin{equation}
    L(w) = \sum_{i=1}^{N} \max(0, 1 - y_i (w^\top x_i + b)) + \frac{\lambda}{2} \|w\|^2
\end{equation}
where \( \lambda \) is the regularization parameter that prevents overfitting.

\vspace{0.5 cm}

L-SVC was applied to the Amazon review dataset with BoW as the feature extraction method. The model demonstrated strong classification performance, achieving an overall accuracy of 90\%. The evaluation metrics highlight the model's balanced effectiveness across all sentiment classes. The precision, recall, and F1-scores for negative, neutral, and positive sentiments were consistently high, exceeding 0.85 in all cases. Specifically, neutral sentiments showed the highest precision (0.92) and recall (0.97), leading to an F1-score of 0.94. Negative and positive sentiments achieved F1-scores of 0.87 and 0.90, respectively.

The confusion matrix further illustrates the model's reliability. Neutral sentiments were classified with the highest accuracy, with only 1,971 misclassifications out of 71,300 instances. Negative sentiments saw 6,491 misclassified as positive and 4,277 as neutral. For positive sentiments, 6,123 samples were misclassified as negative, while 1,969 were labeled neutral, as Figure 24-25. These results emphasize L-SVC's robustness in sentiment classification, particularly for neutral and positive classes, while minor overlaps in feature representation influenced the misclassification of negative sentiments.

\begin{figure}[h]
    \centering
    \begin{minipage}{0.45\linewidth}
        \centering
        \includegraphics[width=\linewidth]{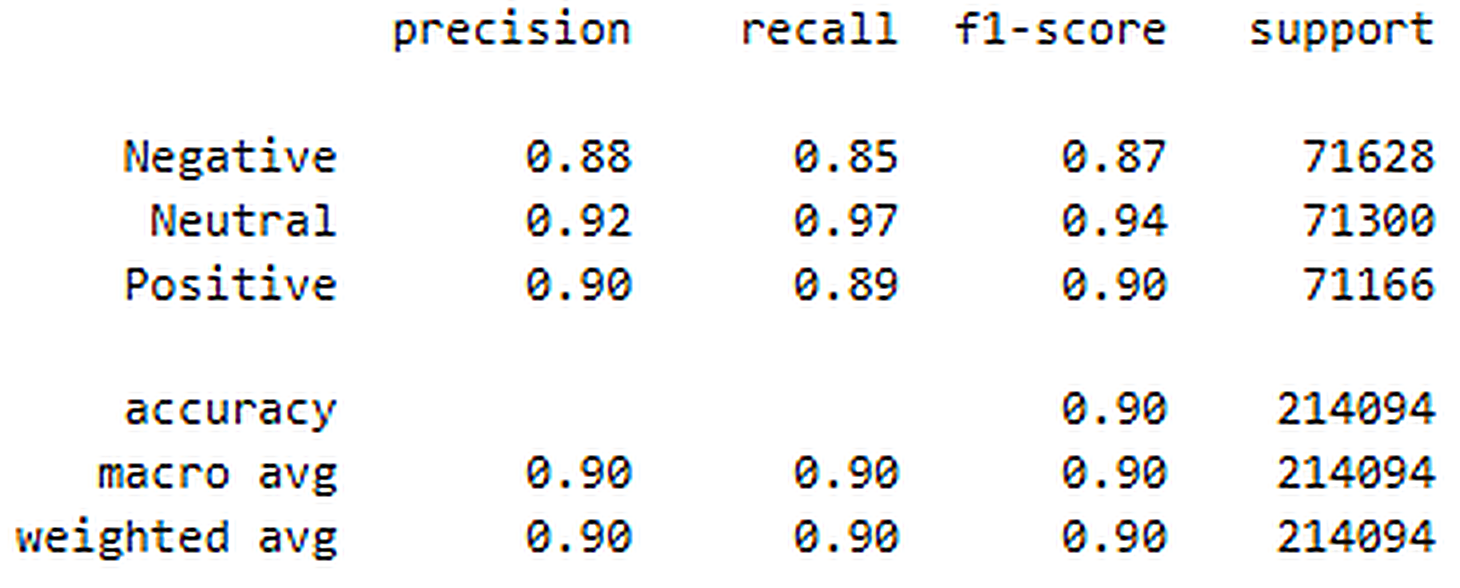}
        \caption{Parameters for Each Class in L-SVC}
        \label{fig:enter-label}
    \end{minipage}
    \hfill
    \begin{minipage}{0.45\linewidth}
        \centering
        \includegraphics[width=\linewidth]{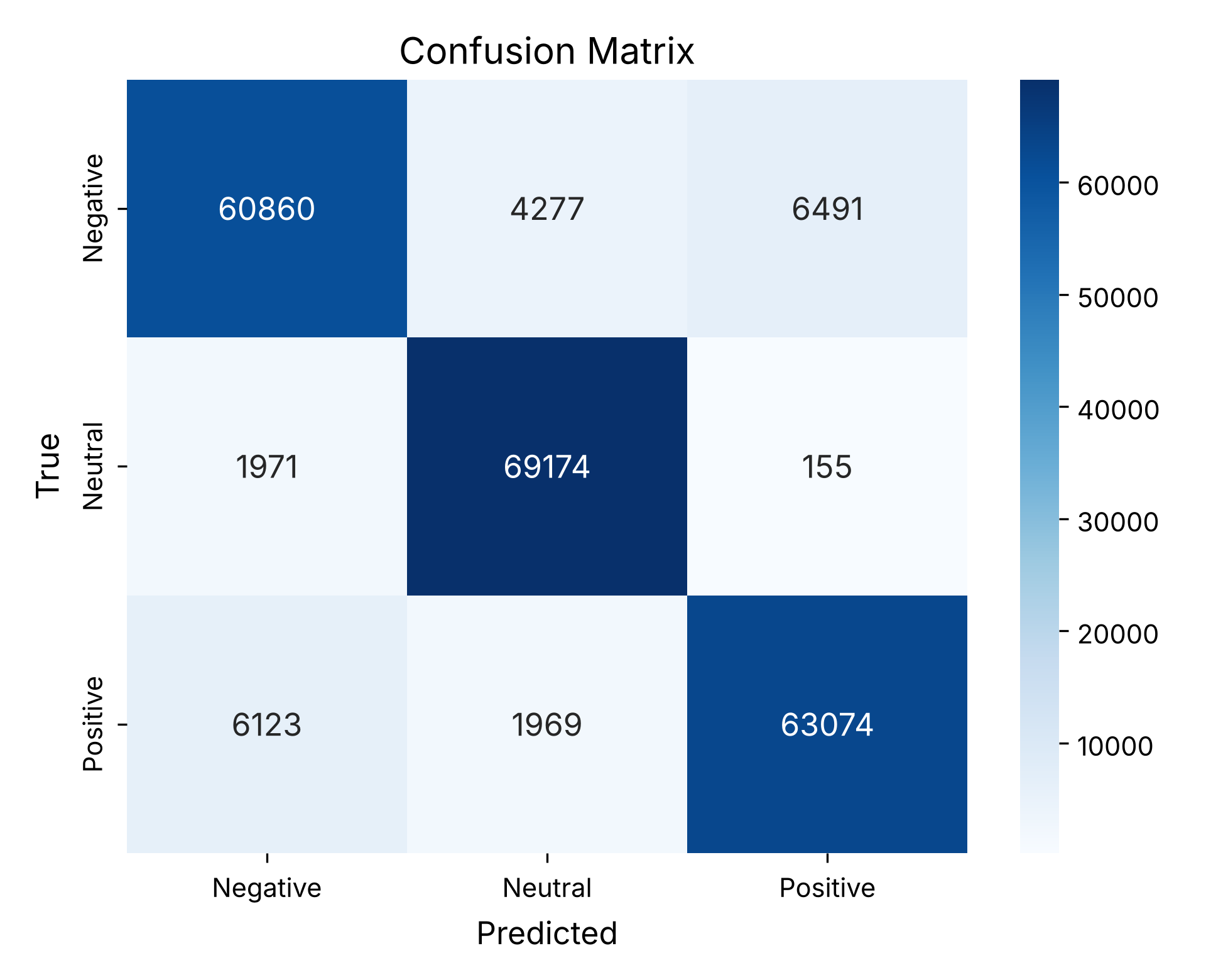}
        \caption{L-SVC Confusion Matrix}
        \label{fig:enter-label}
    \end{minipage}
\end{figure}

\subsubsection{Bernoulli Naive Bayes (BNB)}

BNB is a probabilistic classification algorithm commonly used for text classification tasks, especially when the input features are binary. In this model, each feature represents whether a word appears in a document (1) or not (0). This makes BNB particularly well-suited for Bag-of-Words representations where the frequency of a word is less important than its presence. The Bernoulli assumption simplifies computation and results in efficient classification performance, especially for large and sparse datasets such as sentiment analysis on customer reviews.
\textbf{Bernoulli Probability Model:}  
The likelihood of a document belonging to class \( c \) is computed as:

\begin{equation}
P(x_i = 1 \mid c) = \theta_{ic}, \quad
P(x_i = 0 \mid c) = 1 - \theta_{ic}
\end{equation}

where \( \theta_{ic} \) is the probability that feature \( x_i \) occurs in class \( c \).

\textbf{Posterior Class Prediction:}

\begin{equation}
\hat{y} = \arg\max_{c \in C} \; P(c) \prod_{i=1}^{n} P(x_i \mid c)
\end{equation}

where \( P(c) \) is the prior probability of class \( c \).

\vspace{0.5cm}

When applied to the Amazon review dataset, the BNB classifier achieved an overall accuracy of 76\%. The model demonstrated reasonable performance across all sentiment categories, with F1-scores of 0.70 (negative), 0.81 (neutral), and 0.76 (positive). Neutral sentiments were identified with the highest recall (0.92), indicating the model's strong ability to detect neutral reviews, likely because neutral text often contains fewer emotionally charged words and aligns well with the Bernoulli representation.

Negative reviews showed the lowest recall (0.64), revealing that a substantial portion of negative instances were predicted as either neutral or positive due to overlapping vocabulary. Positive sentiment classification was moderately strong with a precision of 0.81 and recall of 0.72. The macro and weighted averages (all 0.76) further indicate that the classifier performed consistently across the three sentiment classes.

Overall, the BNB model provides a fast and interpretable baseline for sentiment analysis. However, its binary feature formulation limits its ability to capture nuanced textual patterns found in longer customer reviews, which explains the moderate overall accuracy of 76\%, along with class-level F1-scores of 0.70 (negative), 0.81 (neutral), and 0.76 (positive). Compared to other classical and ensemble models used in this study, BNB underperforms noticeably. The RFC achieved an exceptionally high accuracy of 99\% with F1-scores exceeding 0.96 across all sentiment classes, demonstrating far greater robustness. Similarly, both L-SVC and LRC achieved 90\% accuracy, with F1-scores of 0.87–0.94 across classes-substantially outperforming the BNB model, as Figure 26-27. These differences highlight the limitations of binary feature modeling in capturing sentiment-rich language compared to linear and ensemble approaches that more effectively leverage the full representation of textual features.

\begin{figure}[h]
    \centering
    \begin{minipage}{0.45\linewidth}
        \centering
        \includegraphics[width=\linewidth]{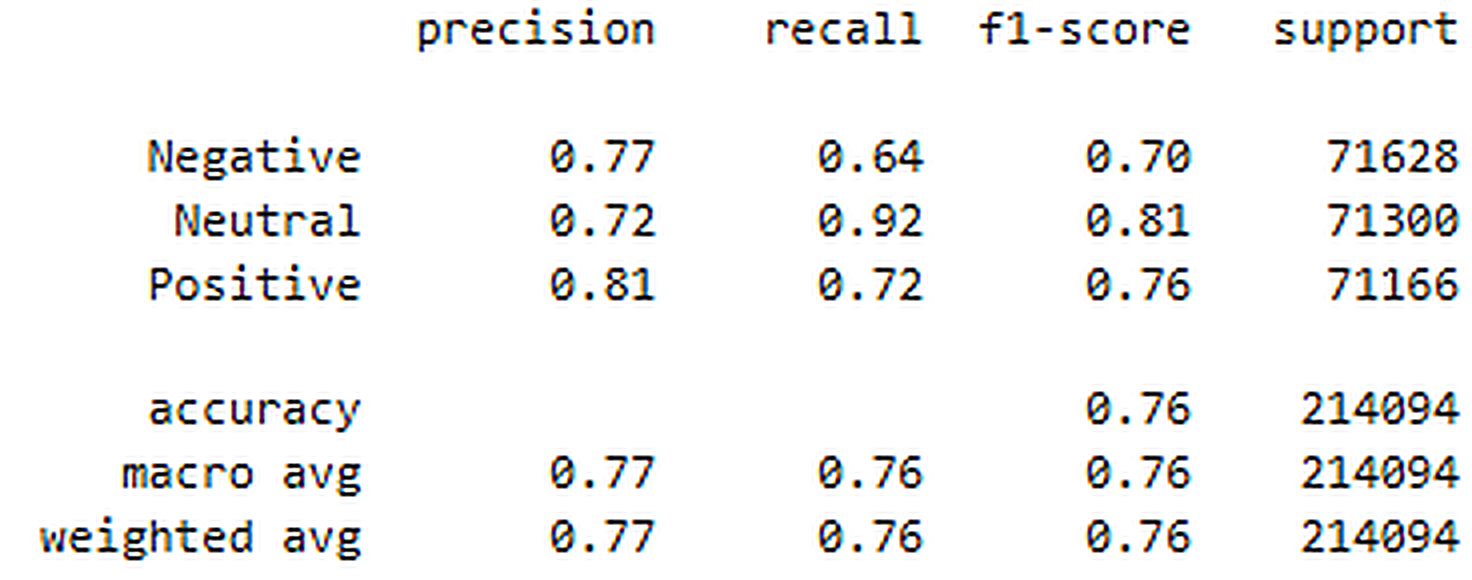}
        \caption{Parameters for Each Class in BNB}
        \label{fig:placeholder}
    \end{minipage}
    \hfill
    \begin{minipage}{0.45\linewidth}
        \centering
        \includegraphics[width=\linewidth]{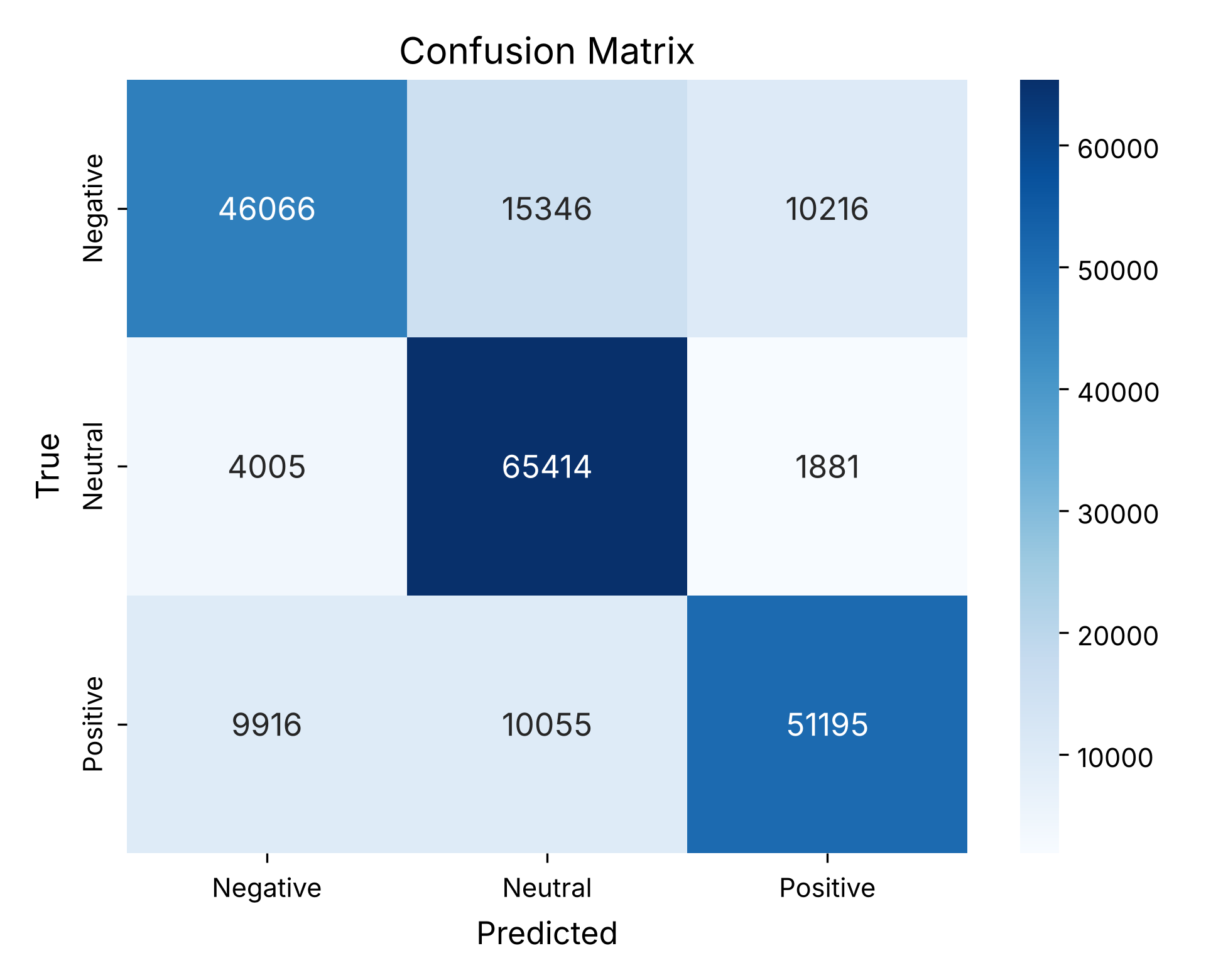}
        \caption{BNB Confusion Matrix}
        \label{fig:placeholder}
    \end{minipage}
\end{figure}

\subsubsection{Multinomial Naive Bayes (MNB)}

MNB is a widely used probabilistic classifier for text analysis and is particularly effective when features represent word counts or term frequencies. Unlike BNB, which considers only the presence or absence of words, the MNB model leverages the frequency of each word in a document. This makes it especially powerful for sentiment analysis tasks using BoW or TF-IDF features, where word occurrence rates carry important semantic information.

\textbf{Multinomial Likelihood Function:}

\begin{equation}
P(\mathbf{x} \mid c) = \frac{\left( \sum_{i} x_i \right)!}{\prod_{i} x_i!} \prod_{i=1}^{n} \theta_{ic}^{x_i}
\end{equation}

where \( x_i \) is the count of word \( i \), and \( \theta_{ic} \) is the probability of word \( i \) appearing in class \( c \).

\textbf{Posterior Prediction Rule:}
\begin{equation}
\hat{y} = \arg\max_{c \in C} ; P(c) \prod_{i=1}^{n} \theta_{ic}^{x_i}
\end{equation}

\vspace{0.5cm}

The MNB classifier outperformed the Bernoulli variant, achieving an overall accuracy of 79\% on the Amazon review dataset. The model demonstrated strong performance across all classes, with F1-scores of 0.75 (negative), 0.85 (neutral), and 0.79 (positive). Neutral sentiments achieved the highest precision (0.88) and strong recall (0.82), indicating the model’s ability to capture the typical structure of neutral reviews more effectively.

Positive sentiments were also classified well, with a recall of 0.81 and an F1-score of 0.79, reflecting the model's capability to recognize emotionally charged vocabulary when evaluated using word frequency information. Negative sentiment classification showed moderate performance, with balanced precision and recall values (0.74 and 0.75), suggesting some overlap between negative and neutral textual patterns but a stable overall predictive ability.

Both macro and weighted averages (precision: 0.80, recall: 0.79, F1-score: 0.80) highlight the model’s balanced performance across the dataset. The improvement of MNB over BNB can be attributed to its ability to leverage term frequencies, which carry richer contextual and semantic meaning.

Overall, the MNB model demonstrates strong suitability for large-scale sentiment analysis tasks, offering a good balance of accuracy, computational efficiency, and interpretability. Its overall accuracy of 79\%, along with F1-scores of 0.75 (negative), 0.85 (neutral), and 0.79 (positive), indicates stable and reliable performance across sentiment classes. Nevertheless, when compared to more advanced models in this study, such as RFC, L-SVC, and LRC, the MNB model is clearly outperformed. The RFC achieved 99\% accuracy and near-perfect F1-scores above 0.96, representing a significant performance margin over MNB. Likewise, L-SVC and LRC both reached 90\% accuracy, with class-level F1-scores ranging from 0.87 to 0.94, surpassing MNB’s performance in all sentiment categories, as Figure 28-29. These results show that although MNB is effective for text classification, modern linear and ensemble models capture sentiment-related patterns more comprehensively, especially in high-dimensional feature spaces generated by large review datasets.

\begin{figure}[h]
    \centering
    \begin{minipage}{0.45\linewidth}
        \centering
        \includegraphics[width=\linewidth]{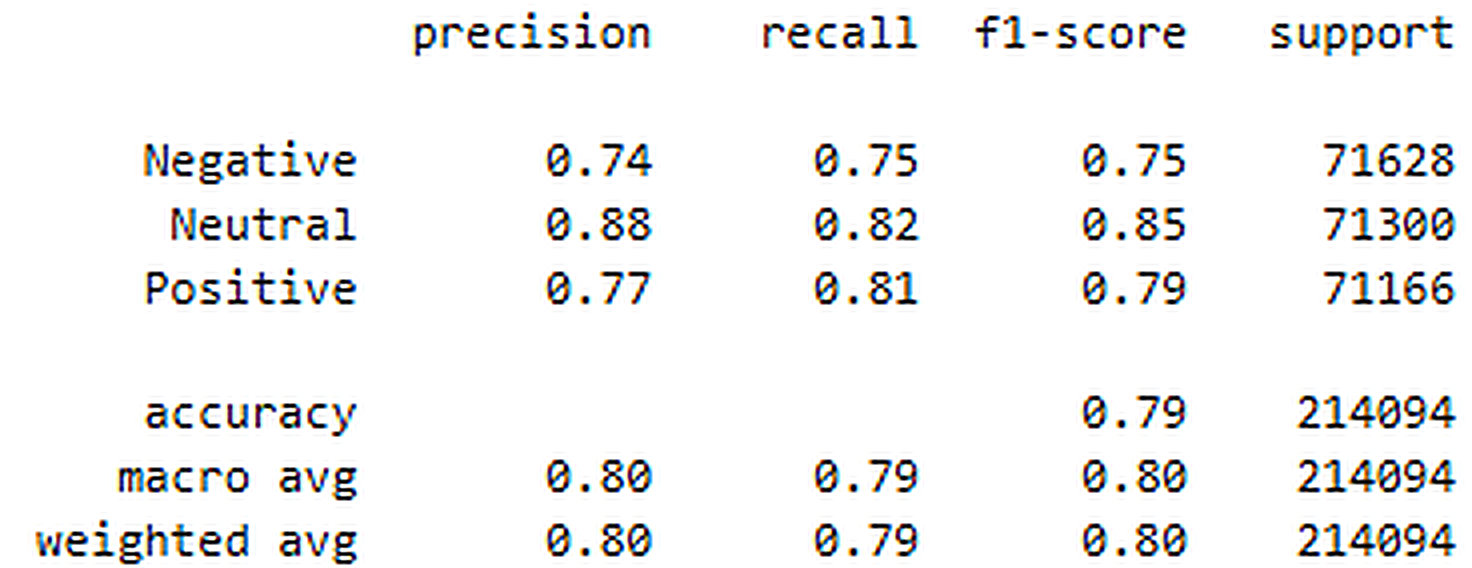}
        \caption{Parameters for Each Class in MNB}
        \label{fig:placeholder}
    \end{minipage}
    \hfill
    \begin{minipage}{0.45\linewidth}
        \centering
        \includegraphics[width=\linewidth]{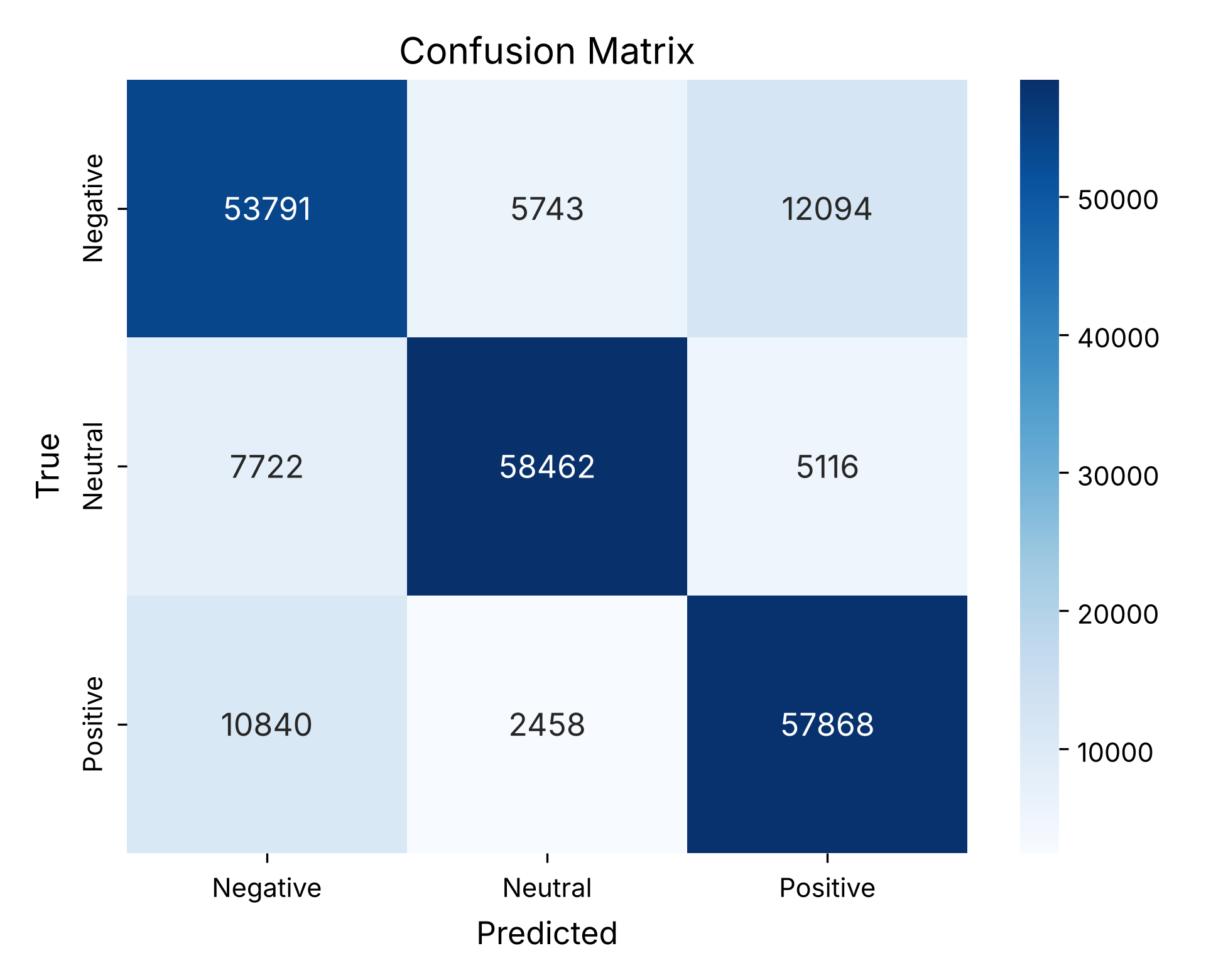}
        \caption{MNB Confusion Matrix}
        \label{fig:placeholder}
    \end{minipage}
\end{figure}

\subsection{Comparison of Top 3 Models}

RFC uses decision trees with random feature selection and majority voting to reduce variance and model non-linear relationships. LRC applies the logistic or softmax function to estimate class probabilities, optimized by cross-entropy loss. L-SVC maximizes the margin between classes, penalizing misclassifications using the hinge loss. Understanding these mathematical foundations clarifies how each model handles multi-class sentiment analysis.

L-SVC, a margin-maximizing classifier, aims to separate classes by identifying a hyperplane with the largest margin. While this approach works well for binary classification, its adaptation to multi-class problems introduces challenges, particularly in distinguishing between neutral and the more polarized positive and negative sentiments. The model performed admirably for positive and negative classes but showed difficulty in handling the neutral class, likely due to the lack of nuanced feature modeling. Compared, RFC outperformed both LRC and L-SVC in providing a more balanced classification across all three sentiment classes. LRC's straightforward probabilistic approach and L-SVC's margin-based separation were effective but exhibited limitations in capturing the subtleties of multi-class sentiment analysis, emphasizing the importance of model selection based on the dataset's complexity and classification goals.

\begin{figure}[h]
    \centering
    \includegraphics[width=0.75 \linewidth]{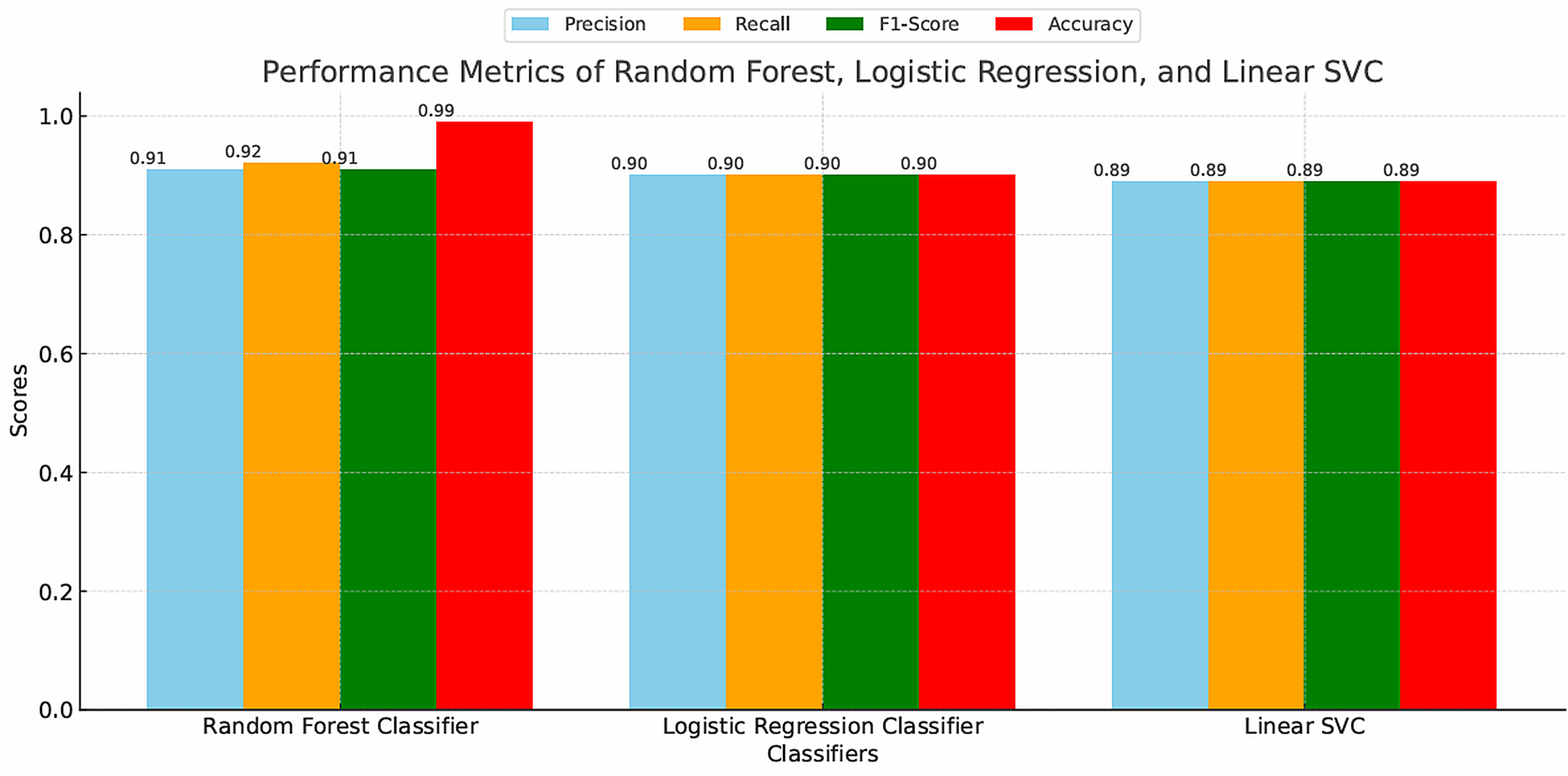}
    \caption{Metrics of Top 3 Model}
    \label{fig:enter-label}
\end{figure}

\subsubsection{Classification Reports}

Figure 30 visually compares the performance of the RFC, LRC, and L-SVC in a multi-class sentiment analysis task (negative, neutral, positive). 
\vspace{0.5cm}

In a confusion matrix, True Positives (TP) are cases where the model correctly predicts the positive class, while True Negatives (TN) are correct predictions of the negative class. False Positives (FP) occur when the model incorrectly labels a negative instance as positive (a false alarm), and False Negatives (FN) occur when the model fails to detect a positive instance and predicts it as negative (a missed detection). These four outcomes form the basis for evaluation metrics like accuracy, precision, recall, and F1-score. Evaluation metrics-\textbf{Precision}, \textbf{Recall}, \textbf{F1-Score}, and \textbf{Accuracy}-provide critical insights into model effectiveness.

\textbf{Precision} is defined as:
\begin{equation}
    P = \frac{TP}{TP + FP}
\end{equation}
It measures the model's ability to correctly identify positive instances, reflecting how reliable positive predictions are.
\vspace{0.5cm}

\begin{figure}[h]
    \centering
    \includegraphics[width=0.85\linewidth]{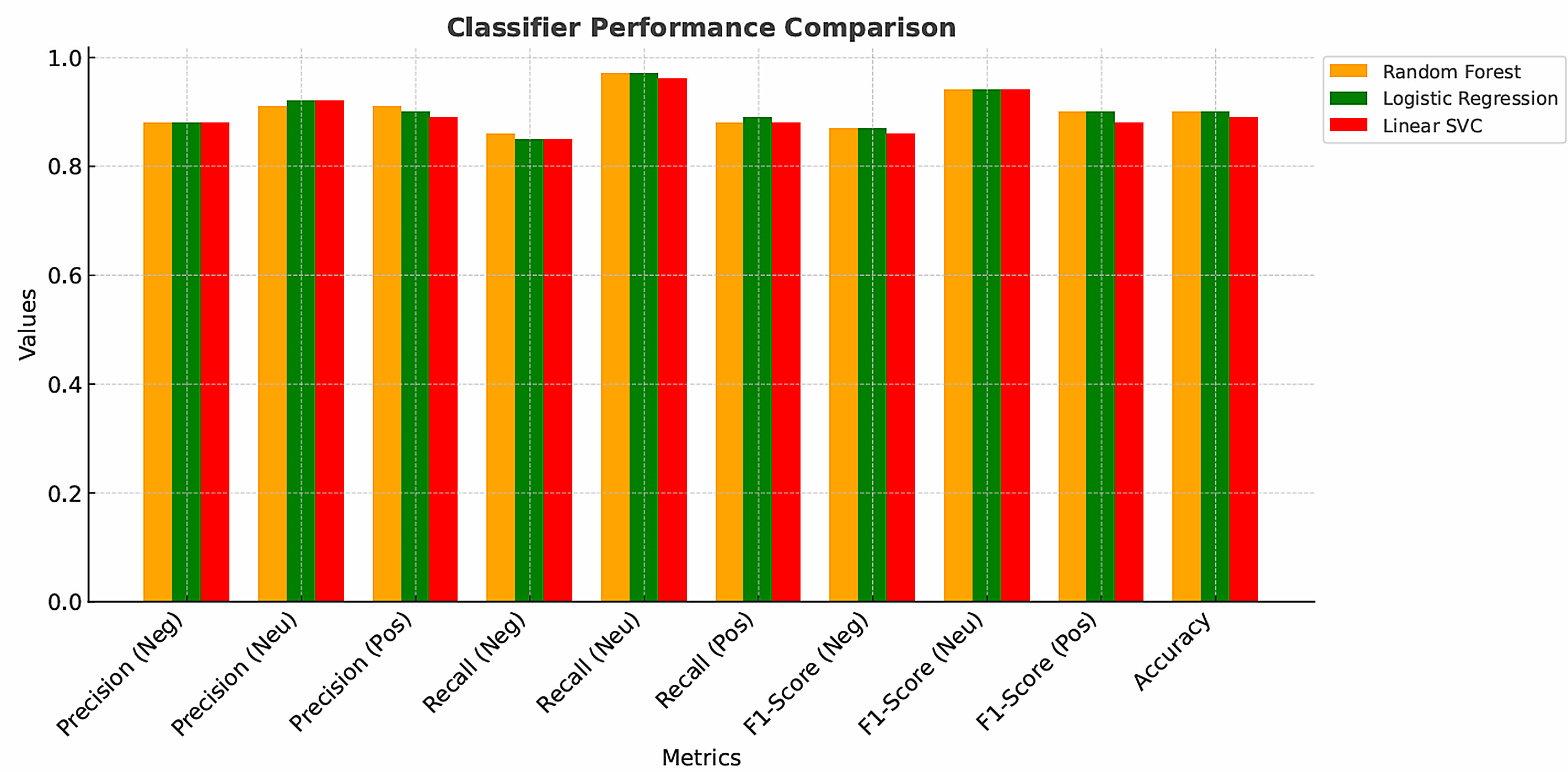}
    \caption{Classification Report Metrics}
    \label{fig:enter-label}
\end{figure}

\textbf{Recall} is calculated as:
\begin{equation}
    R = \frac{TP}{TP + FN}
\end{equation}
It evaluates how well the model captures all relevant positive cases, essential for minimizing false negatives.

\textbf{F1-Score}, the harmonic mean of precision and recall, is given by:
\begin{equation}
    F1 = \frac{2 \times P \times R}{P + R}
\end{equation}

\textbf{Accuracy} measures the overall correctness of the model:
\begin{equation}
    A = \frac{TP + TN}{TP + TN + FP + FN}
\end{equation}

In Figure 31, RFC demonstrates robust performance across metrics due to its ensemble structure, excelling in capturing complex patterns. LRC shows solid but slightly lower performance, while L-SVC struggles with neutral sentiment classification due to its linear separation approach. This comparison highlights RFC's superior adaptability for nuanced sentiment classification, making it the most balanced model for this task.[11]

\vspace{0.5cm}

\subsubsection{Running Time Comparison} L-SVC demonstrated the fastest execution time among the three models, benefiting from its efficient linear computations; however, it achieved the lowest accuracy at 89\%. LRC offered a balance between computational efficiency and predictive performance, with a moderate running time and an accuracy of 90\%. In contrast, RFC attained the highest accuracy at 99\% but incurred the longest execution time due to its ensemble of decision trees and complex feature evaluations. (Figure 32) Although RFC appears to be the most accurate model, the considerable training time raises concerns about its practicality in time-sensitive applications. Therefore, the trade-off between computational efficiency and predictive performance will be examined in greater detail in the subsequent sections.

\begin{figure}[h]
    \centering
    \includegraphics[width=0.5\linewidth]{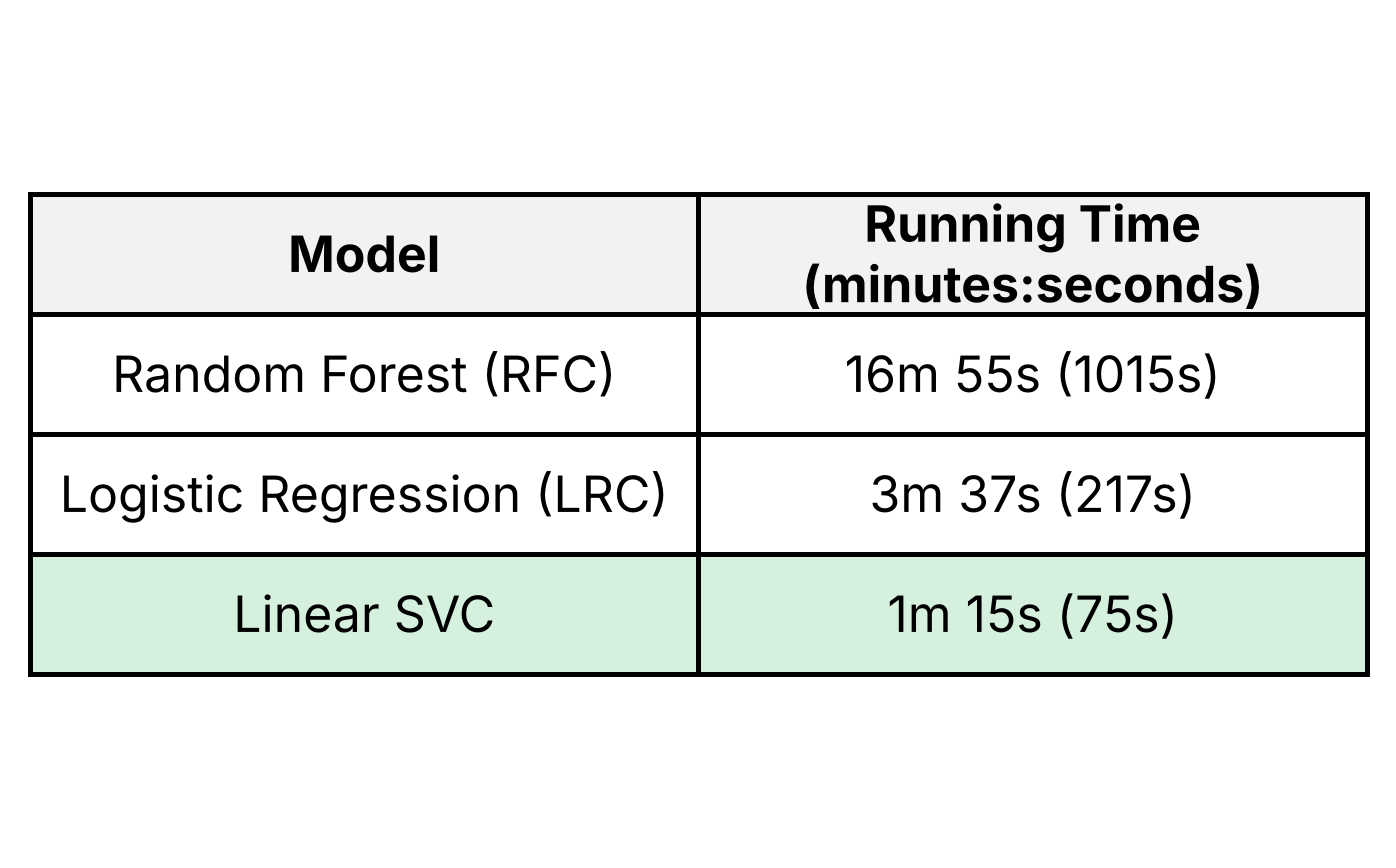}
    \caption{Running Times of Top 3 Model}
    \label{fig:enter-label}
\end{figure}

\subsection{Comparison of Feature Extraction Methods}

\subsubsection{Bag of Words (BoW)}
BoW model converts text data into a matrix of word counts, ignoring grammar and word order while retaining word frequency. Each document is transformed into a vector where each element reflects the occurrence of a specific word. This method is straightforward and effective for recognizing general word usage patterns but lacks the ability to capture contextual nuances.

\begin{table}[h] 
\renewcommand{\arraystretch}{1.3} 
\begin{center}
\resizebox{0.65\textwidth}{!}{%
\begin{tabular}{|c|c|c|}
\hline
\textbf{Model} & \textbf{BoW} & \textbf{TF-IDF} \\ 
\hline
\textbf{Random Forest Classifier (RFC)} & 0.99 & 0.98 \\ 
\hline
\textbf{Logistic Regression Classifier (LRC)} & 0.90 & 0.85 \\ 
\hline
\textbf{Linear Support Vector Classifier (L-SVC)} & 0.89 & 0.84 \\ 
\hline
\textbf{Bernoulli Naive Bayes (BNB)} & 0.76 & 0.74 \\ 
\hline
\textbf{Multinomial Naive Bayes (MNB)} & 0.79 & 0.76 \\ 
\hline
\end{tabular}%
}
\end{center}
\caption{Feature Extraction Method Accuracy Comparison}
\label{tab:model_comparison}
\end{table}

\subsubsection{Term Frequency-Inverse Document Frequency (TF-IDF)}
TF-IDF enhances the BoW approach by weighing word frequency against the uniqueness of words across documents. This model down-weights frequently occurring words and highlights rarer, more informative terms. Consequently, TF-IDF offers a more balanced and contextually rich feature representation, which can be more effective for sentiment classification. Model accuracies and training times can be found in Table 4-5 for better comparison across the feature extraction techniques. 

\begin{table} 
\renewcommand{\arraystretch}{1.3}
\begin{center}
\resizebox{0.75\textwidth}{!}{%
\begin{tabular}{|c|c|c|}
\hline
\textbf{Model} & \textbf{BoW (Time)} & \textbf{TF-IDF (Time)} \\ 
\hline
\textbf{Random Forest Classifier (RFC)} & 16m 55s & 17m 16s \\ 
\hline
\textbf{Logistic Regression Classifier (LRC)} & 3m 37s & 3m 36s \\ 
\hline
\rowcolor{green!20}\textbf{Linear Support Vector Classifier (L-SVC)} & 1m 15s & 1m 14s \\ 
\hline
\textbf{Bernoulli Naive Bayes (BNB)} & 0m 29s & 0m 29s \\ 
\hline
\textbf{Multinomial Naive Bayes (MNB)} & 0m 6s & 0m 6s \\ 
\hline
\end{tabular}%
}
\end{center}
\caption{Feature Extraction Method Time Comparison}
\label{tab:training_time_comparison}
\end{table}

\subsection*{Comparative Analysis of Feature Extraction Impact}
\begin{itemize}
    \item \textbf{Bag of Words} effectively captures frequent sentiment words, benefiting models like RFC that handle high-dimensional data.
    \item \textbf{TF-IDF} highlights contextually important words but introduces sparsity, which may hinder simpler models.
    \item RFC consistently outperformed other models with both methods, showing a slight preference for BoW.
    \item LRC and L-SVC showed slight declines with TF-IDF due to challenges in handling weighted features.
    \item Naive Bayes models underperformed with both methods, highlighting their sensitivity to feature representation.
\end{itemize}

The BoW method produced slightly higher accuracy across most models, suggesting that frequent word occurrence was sufficient for differentiating between positive, negative, and neutral sentiments. Although TF-IDF theoretically provides richer feature representations, its sparsity and weighting complexities slightly hindered model performance, especially for simpler classifiers. This comparison underscores the significance of selecting appropriate feature extraction techniques for optimal sentiment analysis performance.

\subsection{Best Machine Learning Model Discussion and Summary}

\begin{table}[h]
\renewcommand{\arraystretch}{1.7}
\begin{center}
\resizebox{1\textwidth}{!}{%
\begin{tabular}{|c|c|c|c|}
\hline
\textbf{Category} & \textbf{Model} & \textbf{Feature Extraction} & \textbf{Performance Metric} \\
\hline
\textbf{Fastest Model} & Multinomial Naive Bayes & BoW & 0m 6s \\
\hline
\textbf{Most Accurate Model} & Random Forest Classifier & BoW & 0.99 Accuracy \\
\hline
\textbf{Most Efficient Model} & Linear Support Vector Classifier & BoW & E = T $\times$ A \\
\hline
\end{tabular}%
}
\end{center}
\caption{Summary of Best Performing Models by Category}
\label{tab:best_models_summary}
\end{table}

A comprehensive evaluation of the machine learning models was conducted using BoW and TF-IDF feature extraction methods. The models are assessed based on their accuracy, training time, and overall efficiency.

The efficiency parameter, defined as 
\[
\text{Efficiency} = \text{Time} \times \text{Accuracy},
\] 
evaluates machine learning models by balancing their predictive performance and computational cost. Accuracy reflects how well the model performs in predictions, while training time measures the resources required for model training. Combining these factors, the efficiency metric provides a practical and holistic measure, highlighting models that achieve high accuracy with minimal computational expense. This balance ensures the selection of models that are not only effective but also resource-efficient, making the metric especially valuable for real-world applications where both performance and speed are critical.

\begin{itemize}
    \item \textbf{RFC} achieved the highest accuracy for both feature extraction methods. However, it is also the most time-consuming, requiring 16m 55s for BoW and 17m 16s for TF-IDF. This is nearly five times longer than the LRC.
    
    \item \textbf{LRC} has the second-highest accuracy and is also the second most time-consuming model, taking 3m 37s (BoW) and 3m 36s (TF-IDF).
    
    \item \textbf{L-SVC} stands out as the most efficient model. Despite being marginally less accurate than RFC (0.89 vs. 0.99), it requires significantly less training time-1m 15s for BoW and 1m 14s for TF-IDF-making it 15 times faster than RFC and three times faster than LRC.
    
    \item \textbf{BNB} and \textbf{MNB} models are the fastest, with MNB training in just 6 seconds. However, their maximum accuracy of 0.79 is notably lower, making them less suitable for accuracy-critical applications.

    \item \textbf{Efficiency Analysis}: L-SVC emerges as the most efficient model when considering both accuracy and training time. Although RFC provides the highest accuracy, its extensive training time reduces its efficiency. (Table 6) 
\end{itemize}

\begin{table}[h]
    \centering
    \caption{Classification Reports of All Models Across Dataset Sizes for ML in BoW}
    \label{tab:all_models_classification}
    \renewcommand{\arraystretch}{1.1}
    \resizebox{0.75\textwidth}{!}{
    \rotatebox{0}{
    	\begin{tabular}{|c|c|c|c|c|c|c|c|c|c|c|c|}
        	\hline
        	\multicolumn{12}{|c|}{\textbf{Classification Reports of All Models Across Dataset Sizes}} \\ 
        	\hline
        	\multirow{2}{*}{\textbf{Dataset Size}} & \multirow{2}{*}{\textbf{Model}} & \multicolumn{3}{c|}{\textbf{Precision}} & \multicolumn{3}{c|}{\textbf{Recall}} & \multicolumn{3}{c|}{\textbf{F1 Score}} & \multirow{2}{*}{\textbf{Accuracy}} \\ 
        	& & Negative & Neutral & Positive & Negative & Neutral & Positive & Negative & Neutral & Positive & \\ 
        	\hline
        	\multirow{5}{*}{Imbalanced}  
        	& RFC & 0.57 & 0.50 & 0.93 & 0.27 & 0.41 & 0.98 & 0.36 & 0.45 & 0.96 & 0.91 \\ 
        	& LRC & 0.68 & 0.59 & 0.94 & 0.40 & 0.31 & 0.98 & 0.50 & 0.40 & 0.96 & 0.93 \\ 
        	& BNB & 0.41 & 0.13 & 0.96 & 0.44 & 0.81 & 0.87 & 0.43 & 0.22 & 0.91 & 0.84 \\ 
        	& L-SVC & 0.71 & 0.57 & 0.94 & 0.35 & 0.20 & 0.98 & 0.47 & 0.29 & 0.96 & 0.92 \\ 
        	& MNB & 0.82 & 0.00 & 0.90 & 0.01 & 0.00 & 0.99 & 0.01 & 0.00 & 0.95 & 0.90 \\ 
        	\hline
        	\multirow{5}{*}{25\% Dataset}  
        	& RFC & 0.84 & 0.97 & 0.87 & 0.86 & 1.00 & 0.82 & 0.85 & 0.98 & 0.84 & 0.89 \\ 
        	& LRC & 0.84 & 0.87 & 0.90 & 0.82 & 0.95 & 0.85 & 0.83 & 0.91 & 0.87 & 0.88 \\ 
        	& BNB & 0.76 & 0.70 & 0.81 & 0.62 & 0.92 & 0.72 & 0.69 & 0.80 & 0.75 & 0.80 \\ 
        	& L-SVC & 0.86 & 0.89 & 0.90 & 0.83 & 0.96 & 0.86 & 0.84 & 0.92 & 0.88 & 0.89 \\ 
        	& MNB & 0.73 & 0.86 & 0.77 & 0.73 & 0.81 & 0.80 & 0.73 & 0.83 & 0.79 & 0.78 \\ 
        	\hline
        	\multirow{5}{*}{50\% Dataset}  
        	& RFC & 0.88 & 0.99 & 0.92 & 0.92 & 1.00 & 0.87 & 0.90 & 0.99 & 0.90 & 0.93 \\ 
        	& LRC & 0.86 & 0.89 & 0.91 & 0.84 & 0.96 & 0.87 & 0.85 & 0.92 & 0.89 & 0.90 \\ 
        	& BNB & 0.77 & 0.71 & 0.81 & 0.64 & 0.92 & 0.72 & 0.70 & 0.76 & 0.74 & 0.76 \\ 
        	& L-SVC & 0.88 & 0.90 & 0.90 & 0.84 & 0.96 & 0.89 & 0.86 & 0.93 & 0.89 & 0.90 \\ 
        	& MNB & 0.75 & 0.86 & 0.77 & 0.74 & 0.82 & 0.79 & 0.74 & 0.83 & 0.79 & 0.79 \\ 
        	\hline
        	\multirow{5}{*}{75\% Dataset}  
        	& RFC & 0.90 & 0.99 & 0.96 & 0.96 & 1.00 & 0.89 & 0.93 & 1.00 & 0.93 & 0.95 \\ 
        	& LRC & 0.86 & 0.90 & 0.91 & 0.84 & 0.96 & 0.88 & 0.85 & 0.92 & 0.89 & 0.90 \\ 
        	& BNB & 0.77 & 0.71 & 0.82 & 0.64 & 0.92 & 0.72 & 0.70 & 0.76 & 0.75 & 0.76 \\ 
        	& L-SVC & 0.88 & 0.90 & 0.90 & 0.84 & 0.96 & 0.89 & 0.86 & 0.93 & 0.89 & 0.90 \\ 
        	& MNB & 0.76 & 0.86 & 0.77 & 0.74 & 0.82 & 0.79 & 0.74 & 0.83 & 0.79 & 0.79 \\ 
        	\hline
        	\multirow{5}{*}{100\% Dataset}  
        	& RFC & 0.97 & 0.98 & 0.98 & 0.98 & 0.98 & 0.97 & 0.97 & 0.98 & 0.97 & 0.97 \\ 
        	& LRC & 0.87 & 0.91 & 0.91 & 0.85 & 0.96 & 0.88 & 0.86 & 0.94 & 0.90 & 0.90 \\ 
        	& BNB & 0.77 & 0.72 & 0.81 & 0.64 & 0.92 & 0.72 & 0.70 & 0.81 & 0.76 & 0.76 \\ 
        	& L-SVC & 0.88 & 0.92 & 0.90 & 0.85 & 0.97 & 0.89 & 0.87 & 0.94 & 0.90 & 0.90 \\ 
        	& MNB & 0.74 & 0.88 & 0.77 & 0.75 & 0.82 & 0.81 & 0.75 & 0.85 & 0.79 & 0.79 \\ 
        	\hline
        	\multirow{5}{*}{Oversampled}  
        	& RFC & 0.96 & 1.00 & 1.00 & 1.00 & 1.00 & 0.96 & 0.98 & 1.00 & 0.98 & 0.99 \\ 
        	& LRC & 0.87 & 0.91 & 0.91 & 0.86 & 0.97 & 0.88 & 0.87 & 0.94 & 0.90 & 0.90 \\ 
        	& BNB & 0.77 & 0.72 & 0.81 & 0.64 & 0.92 & 0.72 & 0.70 & 0.81 & 0.76 & 0.76 \\ 
        	& L-SVC & 0.88 & 0.92 & 0.90 & 0.85 & 0.97 & 0.89 & 0.87 & 0.94 & 0.90 & 0.90 \\ 
        	& MNB & 0.74 & 0.88 & 0.77 & 0.75 & 0.82 & 0.81 & 0.75 & 0.85 & 0.79 & 0.79 \\ 
        	\hline
    	\end{tabular}
    }}\\
    \tiny{Note: Numerical values are rounded to the nearest decimal point for consistent precision. All distributions are balanced, except for the Imbalanced one.}

\end{table}

\subsection{Comparison of Balanced and Imbalanced Machine Learning Models}

When comparing the performance of models trained on imbalanced versus balanced datasets, the difference in prediction quality and fairness becomes evident. In the imbalanced scenario, although models such as RFC, LR, and L-SVC achieved high overall accuracy (91–92\%), this performance was misleading. These models were primarily biased toward the dominant positive class, neglecting other classes. For instance, in the BoW approach with the imbalanced dataset as can be seen in Table 7, RFC achieved just 0.5005 precision, 0.4082 recall, and 0.4497 F1-score for the neutral class, revealing the model’s inability to properly classify less-represented classes. This imbalance led to unfair and unreliable predictions, especially in applications where all sentiments must be equally understood.

On the other hand, with a balanced dataset, all three models -RFC, LR, and L-SVC- not only achieved strong accuracy but also demonstrated consistent performance across all sentiment categories. Particularly in the balanced BoW and TF-IDF approaches, RFC delivered near-perfect results (98–99\%). Even LRC and L-SVC maintained well-rounded metrics, offering reliable predictions for every sentiment class,as can be seen in Table 8. The key advantage in the balanced dataset is that models could learn equally from all sentiment categories, resulting in a fairer and more generalizable prediction behavior.

Another critical factor is training efficiency. Although RFC consistently had the highest accuracy, it also had the longest training times (approx. 17 min), making it less efficient for practical applications. In contrast, L-SVC offered a strong balance between speed and accuracy, especially in the balanced dataset, where it completed training in just 1m 14s (TF-IDF) while still achieving 90\% accuracy - matching LR's performance, but in significantly less time.

In summary, using a balanced dataset significantly improves model performance across all sentiment classes. It avoids the misleading results caused by high accuracy in imbalanced datasets and ensures the model learns meaningful patterns for all classes - not just the majority. While RFC remains the top performer in terms of raw accuracy, L-SVC stands out as the most efficient and practical model, especially when considering training time and balanced prediction quality. Therefore, balancing the dataset before training is crucial for building robust sentiment analysis models.

\begin{table}
    \centering
    \caption{Classification Reports of All Models Across Dataset Sizes for ML in TF-IDF}
    \label{tab:all_models_classification_tfidf}
    \renewcommand{\arraystretch}{1.1}
    \resizebox{0.75\textwidth}{!}{
    \rotatebox{0}{
    	\begin{tabular}{|c|c|c|c|c|c|c|c|c|c|c|c|}
        	\hline
        	\multicolumn{12}{|c|}{\textbf{Classification Reports of All Models Across Dataset Sizes}} \\ 
        	\hline
        	\multirow{2}{*}{\textbf{Dataset Size}} & \multirow{2}{*}{\textbf{Model}} & \multicolumn{3}{c|}{\textbf{Precision}} & \multicolumn{3}{c|}{\textbf{Recall}} & \multicolumn{3}{c|}{\textbf{F1 Score}} & \multirow{2}{*}{\textbf{Accuracy}} \\ 
        	& & Negative & Neutral & Positive & Negative & Neutral & Positive & Negative & Neutral & Positive & \\ 
        	\hline
        	\multirow{5}{*}{Imbalanced}  
        	& RFC & 0.53 & 0.47 & 0.89 & 0.22 & 0.37 & 0.93 & 0.31 & 0.41 & 0.91 & 0.86 \\ 
        	& LRC & 0.63 & 0.54 & 0.90 & 0.35 & 0.26 & 0.94 & 0.45 & 0.35 & 0.91 & 0.88 \\ 
        	& BNB & 0.38 & 0.10 & 0.91 & 0.39 & 0.76 & 0.83 & 0.38 & 0.18 & 0.86 & 0.79 \\ 
        	& L-SVC & 0.45 & 0.33 & 0.92 & 0.54 & 0.78 & 0.90 & 0.49 & 0.54 & 0.88 & 0.86 \\ 
        	& MNB & 0.77 & 0.00 & 0.86 & 0.00 & 0.00 & 0.94 & 0.00 & 0.00 & 0.90 & 0.85 \\ 
        	\hline
        	\multirow{5}{*}{25\% Dataset}  
        	& RFC & 0.79 & 0.92 & 0.82 & 0.81 & 0.95 & 0.78 & 0.80 & 0.93 & 0.79 & 0.84 \\ 
        	& LRC & 0.79 & 0.83 & 0.86 & 0.78 & 0.90 & 0.80 & 0.78 & 0.86 & 0.82 & 0.83 \\ 
        	& BNB & 0.71 & 0.65 & 0.76 & 0.58 & 0.87 & 0.68 & 0.64 & 0.75 & 0.70 & 0.75 \\ 
        	& L-SVC & 0.81 & 0.85 & 0.86 & 0.79 & 0.91 & 0.81 & 0.79 & 0.87 & 0.83 & 0.84 \\ 
        	& MNB & 0.68 & 0.81 & 0.72 & 0.69 & 0.76 & 0.75 & 0.69 & 0.78 & 0.74 & 0.73 \\ 
        	\hline
        	\multirow{5}{*}{50\% Dataset}  
        	& RFC & 0.84 & 0.94 & 0.87 & 0.87 & 0.95 & 0.82 & 0.85 & 0.94 & 0.85 & 0.88 \\ 
        	& LRC & 0.82 & 0.85 & 0.87 & 0.80 & 0.91 & 0.82 & 0.81 & 0.88 & 0.84 & 0.85 \\ 
        	& BNB & 0.72 & 0.66 & 0.76 & 0.59 & 0.87 & 0.68 & 0.65 & 0.71 & 0.69 & 0.71 \\ 
        	& L-SVC & 0.83 & 0.85 & 0.86 & 0.79 & 0.91 & 0.84 & 0.81 & 0.88 & 0.84 & 0.85 \\ 
        	& MNB & 0.70 & 0.81 & 0.72 & 0.69 & 0.77 & 0.74 & 0.69 & 0.78 & 0.74 & 0.74 \\ 
        	\hline
        	\multirow{5}{*}{75\% Dataset}  
        	& RFC & 0.85 & 0.94 & 0.91 & 0.91 & 0.95 & 0.84 & 0.88 & 0.95 & 0.88 & 0.90 \\ 
        	& LRC & 0.82 & 0.85 & 0.87 & 0.80 & 0.91 & 0.83 & 0.81 & 0.87 & 0.84 & 0.85 \\ 
        	& BNB & 0.72 & 0.66 & 0.77 & 0.59 & 0.87 & 0.68 & 0.65 & 0.71 & 0.70 & 0.71 \\ 
        	& L-SVC & 0.83 & 0.85 & 0.86 & 0.79 & 0.91 & 0.84 & 0.81 & 0.88 & 0.84 & 0.85 \\ 
        	& MNB & 0.71 & 0.81 & 0.72 & 0.69 & 0.77 & 0.74 & 0.69 & 0.78 & 0.74 & 0.74 \\ 
        	\hline
        	\multirow{5}{*}{100\% Dataset}  
        	& RFC & 0.91 & 0.91 & 0.91 & 0.91 & 0.91 & 0.91 & 0.91 & 0.91 & 0.91 & 0.91 \\ 
        	& LRC & 0.84 & 0.88 & 0.87 & 0.87 & 0.91 & 0.83 & 0.85 & 0.89 & 0.84 & 0.84 \\ 
        	& BNB & 0.71 & 0.67 & 0.76 & 0.59 & 0.87 & 0.68 & 0.65 & 0.76 & 0.71 & 0.71 \\ 
        	& L-SVC & 0.83 & 0.87 & 0.86 & 0.80 & 0.92 & 0.84 & 0.82 & 0.89 & 0.85 & 0.85 \\ 
        	& MNB & 0.70 & 0.82 & 0.72 & 0.70 & 0.77 & 0.76 & 0.70 & 0.80 & 0.74 & 0.75 \\ 
        	\hline
        	\multirow{5}{*}{Oversampled}  
        	& RFC & 0.91 & 0.95 & 0.95 & 0.95 & 0.95 & 0.91 & 0.93 & 0.95 & 0.93 & 0.94 \\ 
        	& LRC & 0.82 & 0.86 & 0.87 & 0.81 & 0.91 & 0.83 & 0.82 & 0.89 & 0.85 & 0.85 \\ 
        	& BNB & 0.72 & 0.67 & 0.76 & 0.59 & 0.87 & 0.68 & 0.65 & 0.76 & 0.71 & 0.71 \\ 
        	& L-SVC & 0.83 & 0.87 & 0.86 & 0.80 & 0.92 & 0.84 & 0.82 & 0.89 & 0.85 & 0.85 \\ 
        	& MNB & 0.69 & 0.83 & 0.72 & 0.70 & 0.77 & 0.76 & 0.70 & 0.80 & 0.74 & 0.74 \\ 
        	\hline
    	\end{tabular}
        }}\\
    \tiny{Note: Numerical values are rounded to the nearest decimal point for consistent precision. All distributions are balanced, except for the imbalanced one.}
\end{table}


\section{Deep Learning Models}

\begin{figure}[h]
    \centering
    \resizebox{0.75\textwidth}{!}{%
        \begin{tikzpicture}[node distance=2cm]

            \tikzstyle{startstop} = [rectangle, rounded corners, minimum width=3cm, minimum height=0.75cm, text centered, draw=black]
            \tikzstyle{process} = [rectangle, minimum width=3cm, minimum height=0.75cm, text centered, draw=black]
            \tikzstyle{arrow} = [thick,->,>=stealth]

            \node (collection) [startstop] {Data Collection};
            \node (preprocessing) [process, right of=collection, xshift=2cm] {Data Preprocessing};
            \node (embedding) [process, right of=preprocessing, xshift=2cm] {Embedding Layer};
            \node (training) [process, below of=embedding, yshift=0.75cm] {Model Training};
            \node (evaluation) [process, left of=training, xshift=-2cm] {Model Evaluation};
            \node (metrics) [process, left of=evaluation, xshift=-2cm] {Metrics Calculation};
            \node (classification) [process, below of=metrics, yshift=0.75cm] {Sentiment Classification};
            \node (deployment) [process, right of=classification, xshift=2cm] {App Deployment};
            \node (decision) [startstop, right of=deployment, xshift=2cm] {Decision Making};

            \draw [arrow] (collection.east) -- (preprocessing.west);
            \draw [arrow] (preprocessing.east) -- (embedding.west);
            \draw [arrow] (embedding) -- (training);
            \draw [arrow] (training) -- (evaluation);
            \draw [arrow] (evaluation) -- (metrics);
            \draw [arrow] (metrics) -- (classification);
            \draw [arrow] (classification.east) -- (deployment.west);
            \draw [arrow] (deployment.east) -- (decision.west);

        \end{tikzpicture}%
    }
    \caption{Operational Flow Overview for Deep Learning}
    \label{fig:operational_flow}
\end{figure}
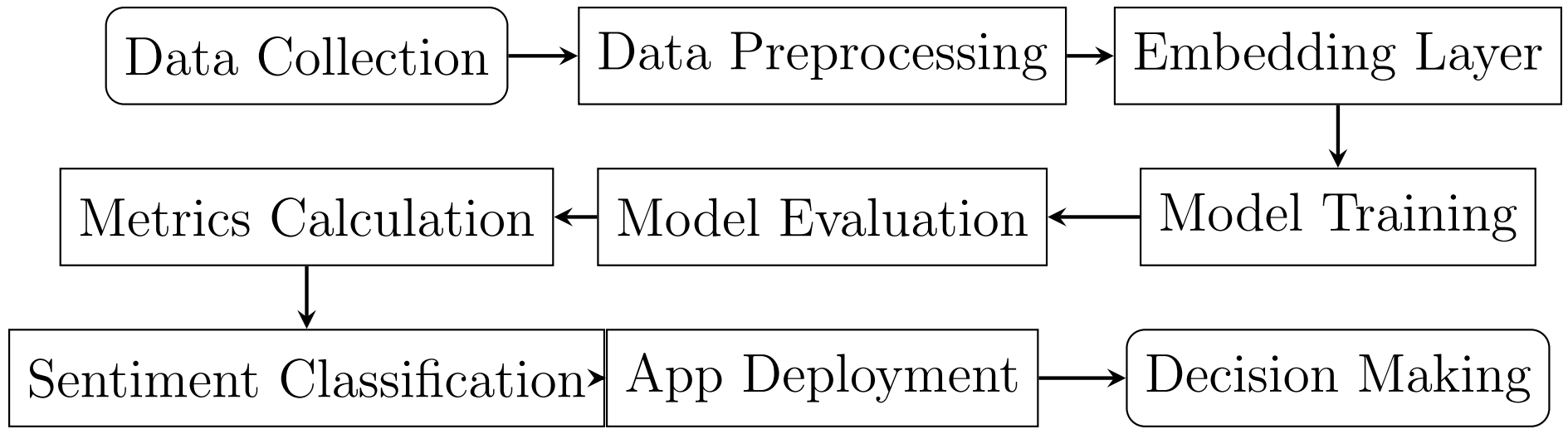

In this study, six advanced deep learning models - LSTM, Bi-LSTM, RNN, GRU, Attention and FastText - were employed for sentiment analysis on the Amazon customer reviews dataset.  LSTM was selected for its ability to capture long-range dependencies, while Bi-LSTM enhanced this by processing data in both directions. RNN served as a baseline for comparison, despite its limitations in retaining long-term context. GRU was chosen for its computational efficiency with fewer parameters than LSTM. The Attention mechanism allowed the model to focus on important parts of the sequence, improving interpretability and performance. FastText, a shallow neural network, was included for its efficient text representation and robustness to out-of-vocabulary words, making it valuable for large-scale sentiment tasks. These structural differences offered a comprehensive evaluation of model performance for sentiment classification. This comprehensive investigation offers valuable insights into the effectiveness of deep learning architectures for sentiment analysis, emphasizing the impact of model selection and architecture design on classification accuracy and practical deployment scenarios. The operational pipeline depicted in Figure 33 outlines the steps from data collection to decision making.

Sentiment analysis is a crucial Natural Language Processing (NLP) task that aims to determine the sentiment expressed in textual data, categorizing it as positive, negative, or neutral. This task has applications in various domains, including customer feedback analysis, social media monitoring, and market research. Traditional machine learning approaches have been widely used for sentiment analysis, relying on manually crafted features and basic text representations. However, these methods often struggle with the complexities of natural language, such as contextual understanding and handling long-term dependencies.

Deep learning models, by contrast, have demonstrated superior performance in sentiment analysis tasks due to their ability to automatically learn hierarchical features from raw text data. Architectures such as LSTM, RNN, GRU, and Attention-based models capture complex patterns in text, enabling more accurate sentiment predictions. Additionally, embeddings like FastText efficiently represent words with semantic relationships, enhancing model understanding.

This study presents a comprehensive sentiment analysis system for Amazon reviews, leveraging advanced deep learning models to classify sentiments. The dataset contains 568,454 reviews spanning October 1999 to October 2012, with detailed information on products and users. The software system includes an interactive user interface that allows customers to select their preferred deep learning model and input custom reviews for sentiment classification. Performance evaluation is conducted using metrics such as precision, recall, F1-score, accuracy, validation accuracy, loss, and validation loss. Experimentation shows that Attention-enhanced models capture contextual relationships effectively, while FastText handles large-scale data efficiently. This research underscores the transformative potential of deep learning in automating sentiment analysis, delivering accurate insights for informed decision-making.

\subsection{General Aspects of Deep Learning}\label{AA}

\subsubsection{Preprocessing and Word Embedding}\label{AA}
To prepare the Amazon customer review dataset for deep learning-based sentiment analysis, the textual reviews were first cleaned and standardized. Cleaning involved converting all text to lowercase, removing HTML tags, eliminating non-alphabetic characters, and trimming leading or trailing whitespace. Each review score was mapped to three sentiment classes: negative (0), neutral (1), and positive (2). The dataset was then split into training and testing sets using an 80/20 ratio while preserving the original class distribution through stratification.

For the deep learning models, tokenization was performed using Keras’ Tokenizer, limited to the 20,000 most frequent words. The tokenized sequences were standardized to a fixed length of 100 using padding and truncation, ensuring uniform input dimensions for neural network layers such as LSTM, Bi-LSTM, or GRU. Sentiment labels were converted into categorical format suitable for multi-class classification, enabling direct use in embedding and recurrent layers.

Although the same initial cleaning steps were applied for both machine learning and deep learning pipelines, deep learning preprocessing differs primarily in tokenization and input representation. While ML models use vectorization techniques such as BoW and TF-IDF to convert text into fixed-size feature vectors, DL models rely on sequential tokenization followed by embedding layers that preserve word order and semantic relationships. Additionally, DL sequences are padded or truncated to a uniform length to accommodate batch processing, allowing models to capture contextual dependencies and sequential patterns in text, which are not leveraged by traditional ML vector representations.

\begin{figure}
    \centering
    \includegraphics[width=0.5\linewidth]{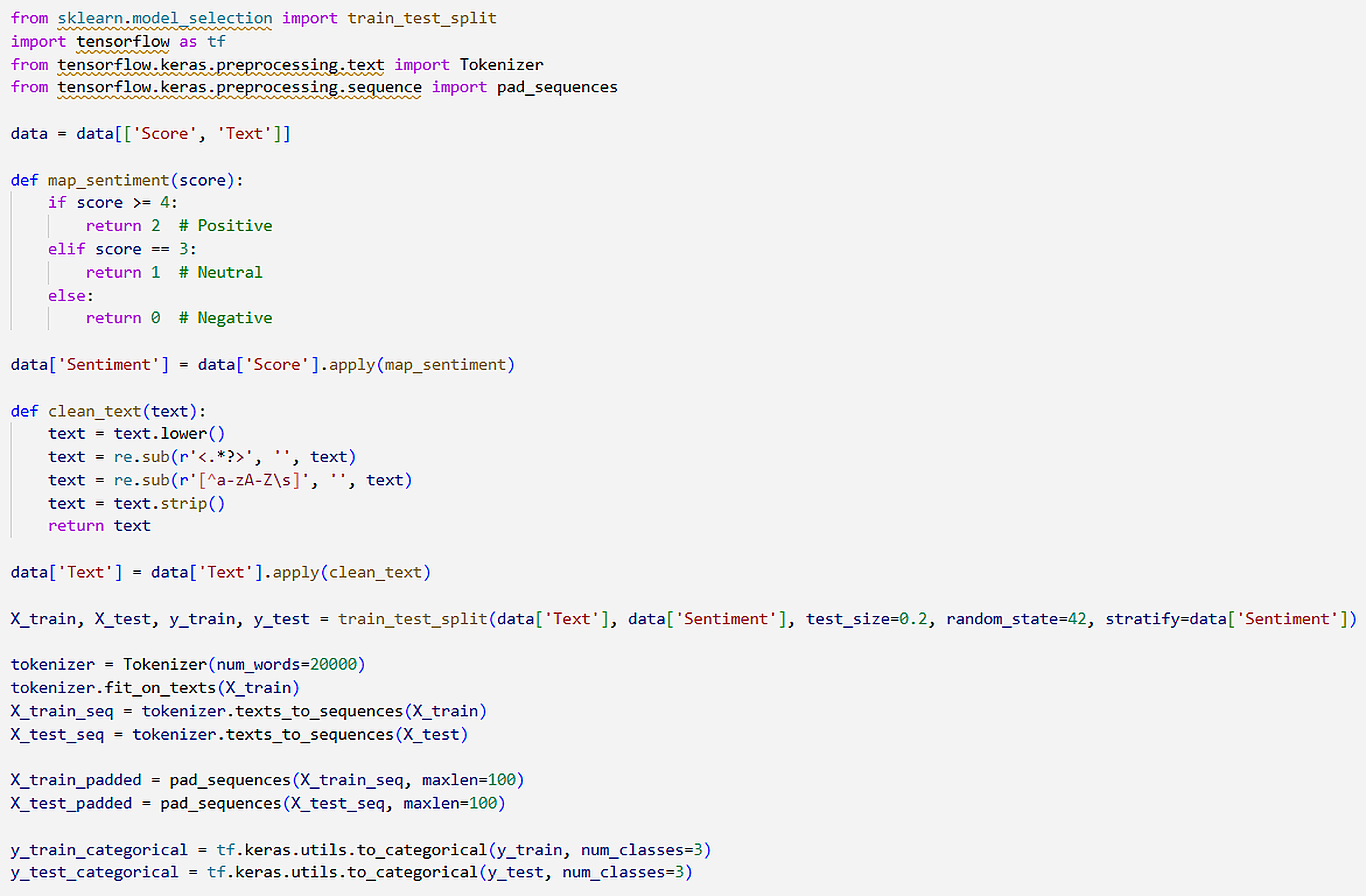}
    \caption{Embedding Coding}
    \label{fig:placeholder}
\end{figure}

Word embeddings are a fundamental technique in NLP that represent words as dense vectors in a continuous vector space. An embedding layer with a vocabulary size of 20,000, an embedding dimension of 128, and a sequence length of 100 was used as the first layer in all deep learning models. (Figure 34) This configuration ensures that semantically similar words are positioned closer in the vector space, improving sentiment classification. Unlike traditional methods that rely on sparse and high-dimensional representations such as BoW or TF-IDF, word embeddings capture both semantic and syntactic relationships between words, enabling models to better understand the context and meaning within a sentence.

A primary advantage of word embeddings is that they overcome the limitations of manual feature extraction in NLP tasks. Traditional methods often require handcrafted features and cannot fully capture complex linguistic patterns or contextual meanings. In contrast, embeddings are learned directly from large text corpora, capturing rich semantic information automatically. Moreover, embeddings make future feature extraction unnecessary, as they inherently encode both syntactic and semantic properties of text, serving as complete and compact representations. This simplification of the modeling process improves efficiency, model generalization, and overall accuracy in NLP tasks, including sentiment analysis, text classification, and machine translation. By leveraging word embeddings, this study enhances the capability of deep learning models to accurately and efficiently analyze customer reviews and extract actionable insights.

\subsubsection{Training Process}

Deep learning models are trained iteratively through multiple epochs, where the dataset is passed through the network repeatedly to refine its parameters and improve classification accuracy. Each epoch adjusts the model based on previous errors, allowing it to learn more effectively over time.

In this sentiment analysis project, the training process was designed to stop if no further increase in accuracy was observed over the last five consecutive iterations. The best-performing model among all trained epochs was automatically saved to ensure optimal performance. Additionally, the changes between epochs were recorded and will be presented in later sections to compare validation accuracy and loss, providing insights into the training process and model improvements. It is crucial to maintain consistent and balanced changes across epochs, as exemplified in Figure 35, avoiding irregular or erratic fluctuations in performance, whether positive or negative, as such instability may indicate suboptimal learning behavior.

\begin{figure}[h]
    \centering
    \includegraphics[width=0.85\linewidth]{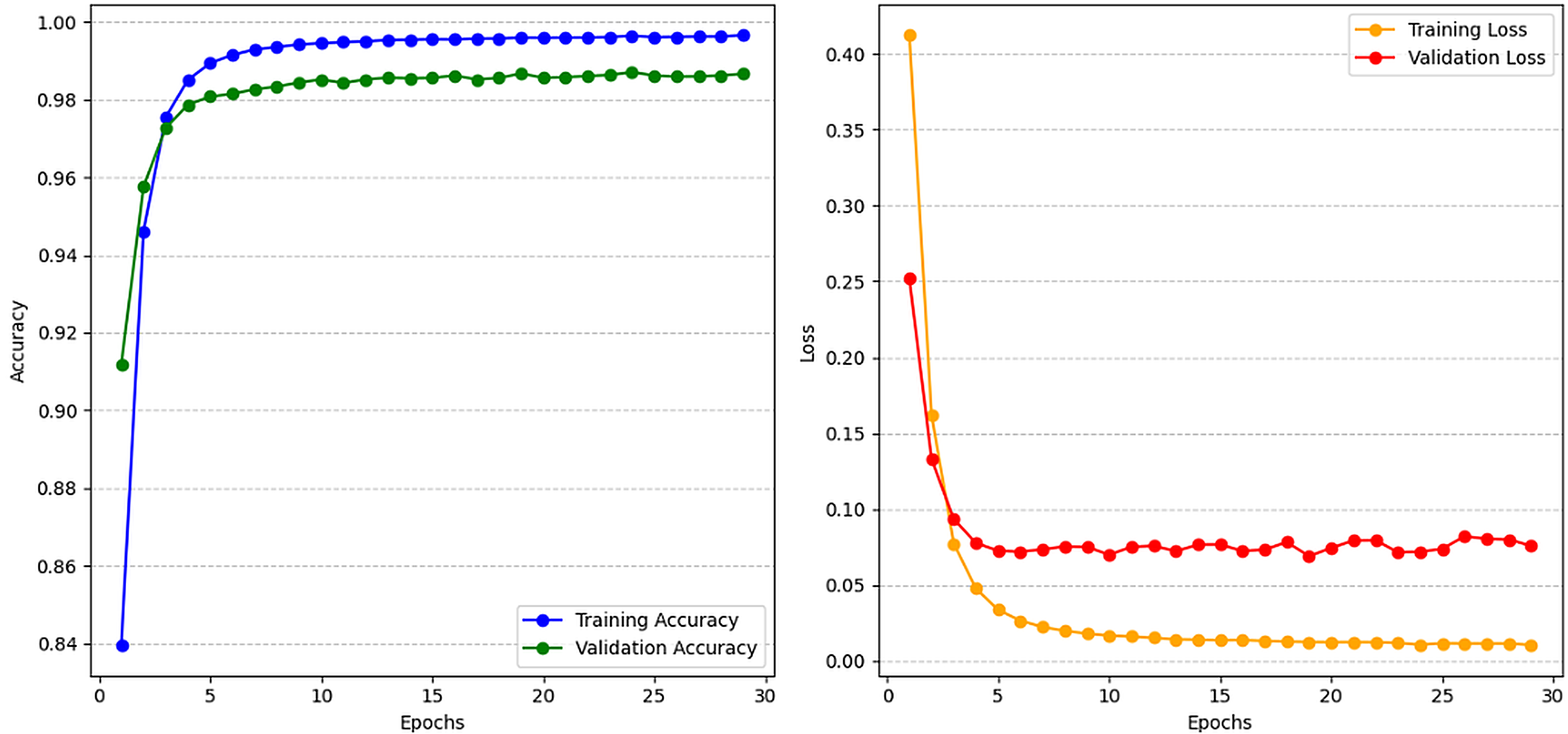}
    \caption{Accuracy and Loss Changes Between Epochs}
    \label{fig:enter-label}
\end{figure}

\begin{table}[h]
\caption{Architectures and training settings of the deep learning models. All models share the embedding layer, output layer, and training settings listed at the bottom.}
\label{tab:dl_params}
\centering
\footnotesize
\begin{tabular}{p{2.6cm}p{3.6cm}p{3.6cm}p{2.8cm}}
\toprule
\textbf{Model} & \textbf{Sequence layer} & \textbf{Pooling / attention} & \textbf{Dense head} \\
\midrule
LSTM & LSTM, 128 units & -- & Dense 64, ReLU \\
Bi-LSTM & Bidirectional LSTM, 128 units & -- & Dense 64, ReLU \\
GRU & GRU, 128 units & -- & Dense 64, ReLU \\
RNN & SimpleRNN, 128 units & -- & Dense 64, ReLU \\
Attention & LSTM, 128 units, \texttt{return\_sequences}=True & Self-attention (\texttt{Attention} layer on $[x,x]$), GlobalAveragePooling1D & Dense 64, ReLU \\
FastText-style & -- & GlobalAveragePooling1D & Dense 64, ReLU \\
\bottomrule
\end{tabular}

\vspace{0.75em}
\begin{tabular}{p{4.2cm}p{9.5cm}}
\toprule
\textbf{Shared setting} & \textbf{Value} \\
\midrule
Embedding & Trainable, randomly initialized (no pretrained vectors); \texttt{input\_dim}=20{,}000, \texttt{output\_dim}=128 \\
Sequence length & 100 (padding and truncation) \\
Tokenizer & Keras \texttt{Tokenizer}, \texttt{num\_words}=20{,}000 \\
Output layer & Dense 3, softmax \\
Loss & Categorical cross-entropy \\
Optimizer & Adam, learning rate 0.001 (Keras default) \\
Batch size / max epochs & 32 / 50 \\
Early stopping & \texttt{monitor}=\texttt{val\_accuracy}, \texttt{patience}=5, \texttt{restore\_best\_weights}=True \\
Dropout & None \\
Data split & 80/20, \texttt{random\_state}=42, stratified \\
\bottomrule
\end{tabular}
\end{table}

\subsection{Presentation of Deep Learning Models}
Six advanced deep learning models-LSTM, RNN, Bidirectional LSTM, GRU, Attention, and FastText-were implemented to classify sentiment within the dataset. To transform the textual data into meaningful numerical representations suitable for deep learning, word embeddings were applied. These embeddings capture semantic relationships between words, enabling the models to better understand and process the contextual information present in the reviews.

Performance evaluation was conducted using a comprehensive set of metrics, including precision, recall, F1 score, accuracy, validation accuracy, loss, and validation loss. These metrics provided a detailed assessment of the models’ ability to classify sentiments effectively across different sentiment categories. The experimental results revealed that Attention-based models outperformed other architectures by capturing long-range dependencies and contextual relationships within the text, while FastText demonstrated superior efficiency when handling large datasets. The best-performing model is then highlighted, and its results are analyzed to provide insights into its strengths and limitations. Additionally, a detailed comparison between the models is presented to offer a broader perspective on their influence on sentiment classification performance, highlighting the critical role of architecture design and word embeddings in achieving optimal results.

\subsubsection{Long Short-Term Memory (LSTM)}
The LSTM is a recurrent neural network architecture designed to overcome the limitations of traditional RNNs by capturing long-term dependencies in sequential data. Its unique structure, featuring memory cells and gating mechanisms, allows it to selectively retain or forget information, making it highly effective for processing complex textual data with varying contexts. This capability makes LSTM well-suited for sentiment analysis tasks where understanding the relationship between words over extended sequences is crucial.

In this study, the LSTM model emerged as a highly accurate deep learning solution for sentiment classification using word embeddings on the Amazon customer review dataset. The model achieved an impressive overall accuracy of 98\%. Detailed performance metrics indicate precision, recall, and F1-scores of 0.98 to 0.99 for negative, neutral, and positive sentiment classes. Analysis of the confusion matrix reveals that the model correctly classified 70,688 negative, 70,755 neutral, and 69,274 positive instances. Misclassifications were minimal, with 1,124 positive samples incorrectly identified as negative and 967 as neutral, as shown in Figure 36. The support parameter highlights the equal distribution of sentiment classes, with each category containing 71,365 samples. These results underscore the model's capability to effectively capture and classify sentiment nuances in textual data, demonstrating its robustness and reliability for large-scale sentiment analysis tasks.

\begin{figure} [h]
    \centering
    \includegraphics[width=0.45\linewidth]{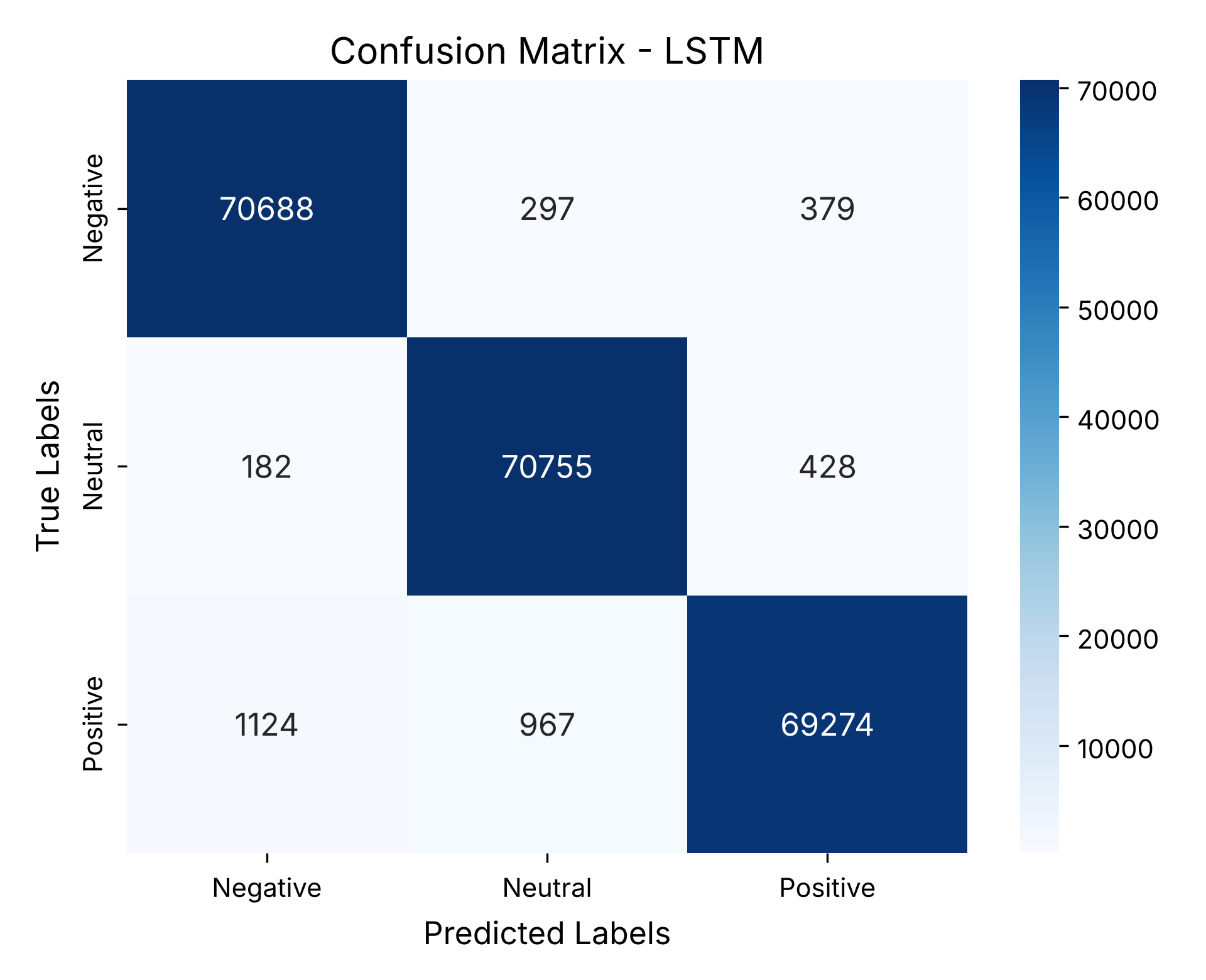}
    \caption{LSTM - Confusion Matrix}
    \label{fig:enter-label}
\end{figure}

\subsubsection{Attention}
The Attention mechanism enhances deep learning models by dynamically focusing on the most relevant parts of a text sequence, improving their ability to capture contextual dependencies. Unlike traditional recurrent architectures, Attention allows the model to weigh important words differently, leading to more precise sentiment classification. This mechanism is particularly advantageous for sentiment analysis, where the meaning of a phrase often depends on specific words or contextual relationships within a sentence.

In this study, the Attention-based model demonstrated exceptional performance in sentiment classification using word embeddings on the Amazon customer review dataset. The model achieved an impressive 99\% overall accuracy, significantly improving classification robustness. Performance metrics indicate precision, recall, and F1-scores of 0.98 to 0.99 across negative, neutral, and positive sentiment classes. The confusion matrix reveals that the model accurately classified 70,721 negative, 70,921 neutral, and 69,680 positive samples. Misclassifications were minimal, with 895 positive samples misclassified as negative and 790 as neutral, as shown in Figure 37. The balanced class distribution, with each sentiment category containing approximately 71,365 samples, further highlights the model’s consistency. These results underscore the Attention mechanism’s effectiveness in capturing nuanced sentiment variations, making it a powerful tool for large-scale sentiment analysis.

\begin{figure} [h]
    \centering
    \includegraphics[width=0.45\linewidth]{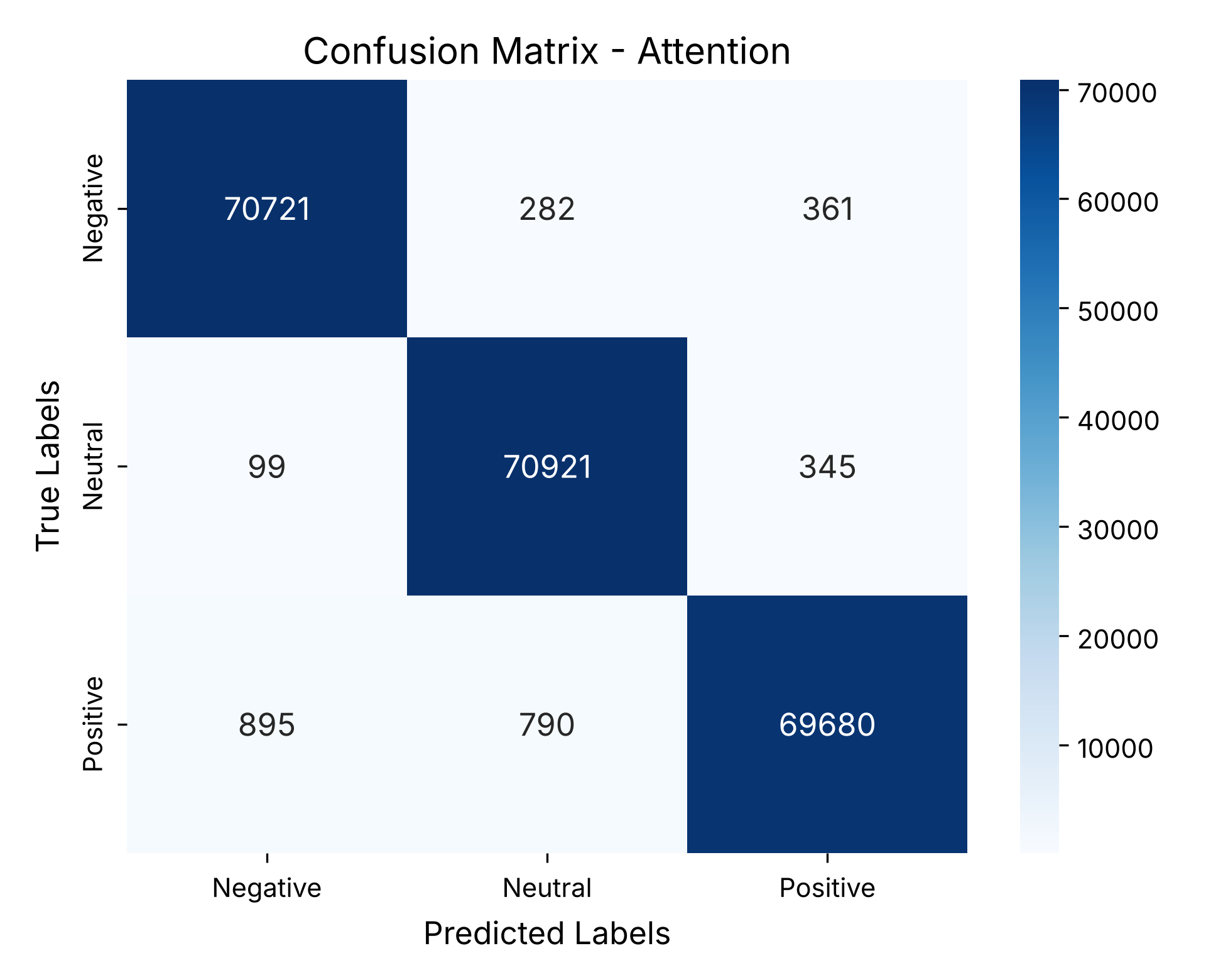}
    \caption{Attention - Confusion Matrix}
    \label{fig:enter-label}
\end{figure}

\subsubsection{Bidirectional Long Short-Term Memory (Bi-LSTM)}
The Bi-LSTM model extends the capabilities of traditional LSTMs by processing input sequences in both forward and backward directions. This bidirectional structure enables the model to capture dependencies from both past and future contexts, making it particularly effective for sentiment analysis, where the sentiment of a word can depend on preceding and succeeding words. By leveraging this dual perspective, the Bi-LSTM model enhances its ability to understand complex linguistic patterns and improves classification accuracy.

In this study, the Bi-LSTM model achieved an overall accuracy of 99\% in sentiment classification using word embeddings on the Amazon customer review dataset. The model demonstrated strong performance, with precision, recall, and F1-scores ranging from 0.97 to 0.99 across negative, neutral, and positive sentiment classes. The confusion matrix shows that the model correctly classified 70,662 negative, 70,689 neutral, and 69,486 positive instances. Misclassifications were minimal, with 924 positive samples incorrectly labeled as negative and 955 as neutral, as shown in Figure 38. The dataset maintained a balanced sentiment distribution, with each class containing approximately 71,365 samples. These results highlight the Bi-LSTM model's ability to effectively capture sentiment nuances by incorporating both past and future context, making it a robust solution for large-scale sentiment analysis.

\begin{figure} [h]
    \centering
    \includegraphics[width=0.45\linewidth]{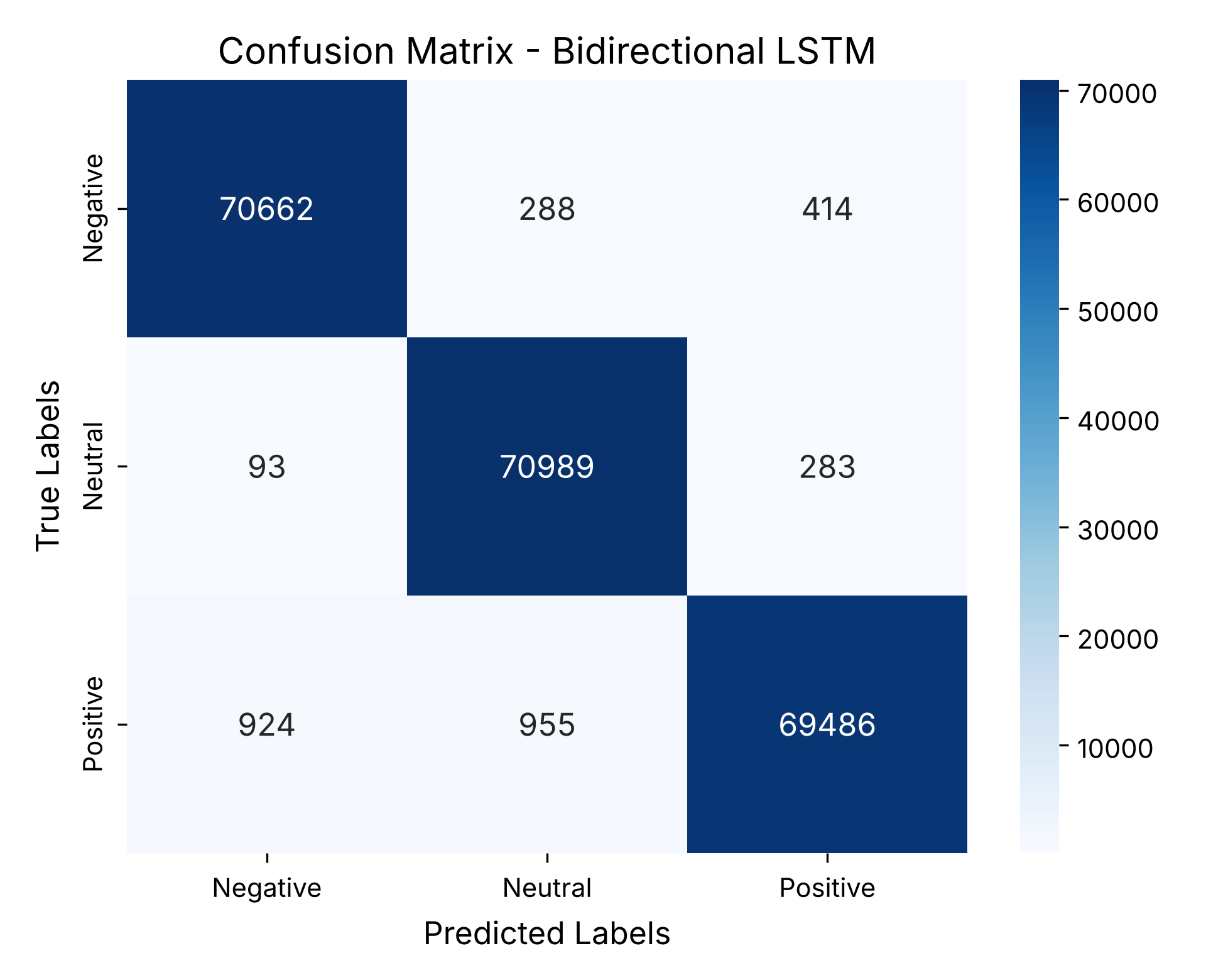}
        \caption{Bi-LSTM - Confusion Matrix}
        \label{fig:enter-label}
\end{figure}

\subsubsection{FastText}
FastText is an efficient and lightweight text classification model developed by Facebook AI, designed to capture subword information through its use of character n-grams. Unlike traditional embedding approaches that learn a single vector for each word, FastText represents words as the sum of their subword embeddings, enabling it to handle rare words and spelling variations more effectively. This structure makes FastText computationally efficient while still maintaining strong performance on large-scale NLP tasks.

In this study, the FastText model achieved an overall accuracy of 95\% in sentiment classification using word embeddings on the Amazon customer review dataset. Performance metrics indicate precision, recall, and F1-scores of 0.77, 0.80, and 0.97 for the negative, neutral, and positive classes, respectively. The model demonstrated particularly strong performance in identifying positive sentiments, achieving an F1-score of 0.98. However, performance on the negative class was comparatively weaker, with a recall of 0.66 and an F1-score of 0.71, suggesting difficulties in distinguishing subtle negative sentiment cues. Neutral sentiment detection remained moderately strong, with a recall of 0.85 and an F1-score of 0.82, as shown in Figure 39. 

Despite its limitations in capturing complex linguistic dependencies compared to recurrent architectures, FastText’s rapid training time of only 3 minutes and 26 seconds highlights its efficiency. These results demonstrate that FastText provides a strong balance between speed and accuracy, making it an effective baseline model for large-scale sentiment classification tasks where computational efficiency is prioritized. However, when compared to more advanced architectures such as LSTM, Bi-LSTM, and Attention, FastText falls short in predictive accuracy due to its inability to model long-range contextual dependencies. Although it is the fastest model in the study, its overall accuracy ranks only above the RNN, indicating that its efficiency comes at the cost of reduced classification performance.

\begin{figure}[h]
    \centering
    \includegraphics[width=0.45\linewidth]{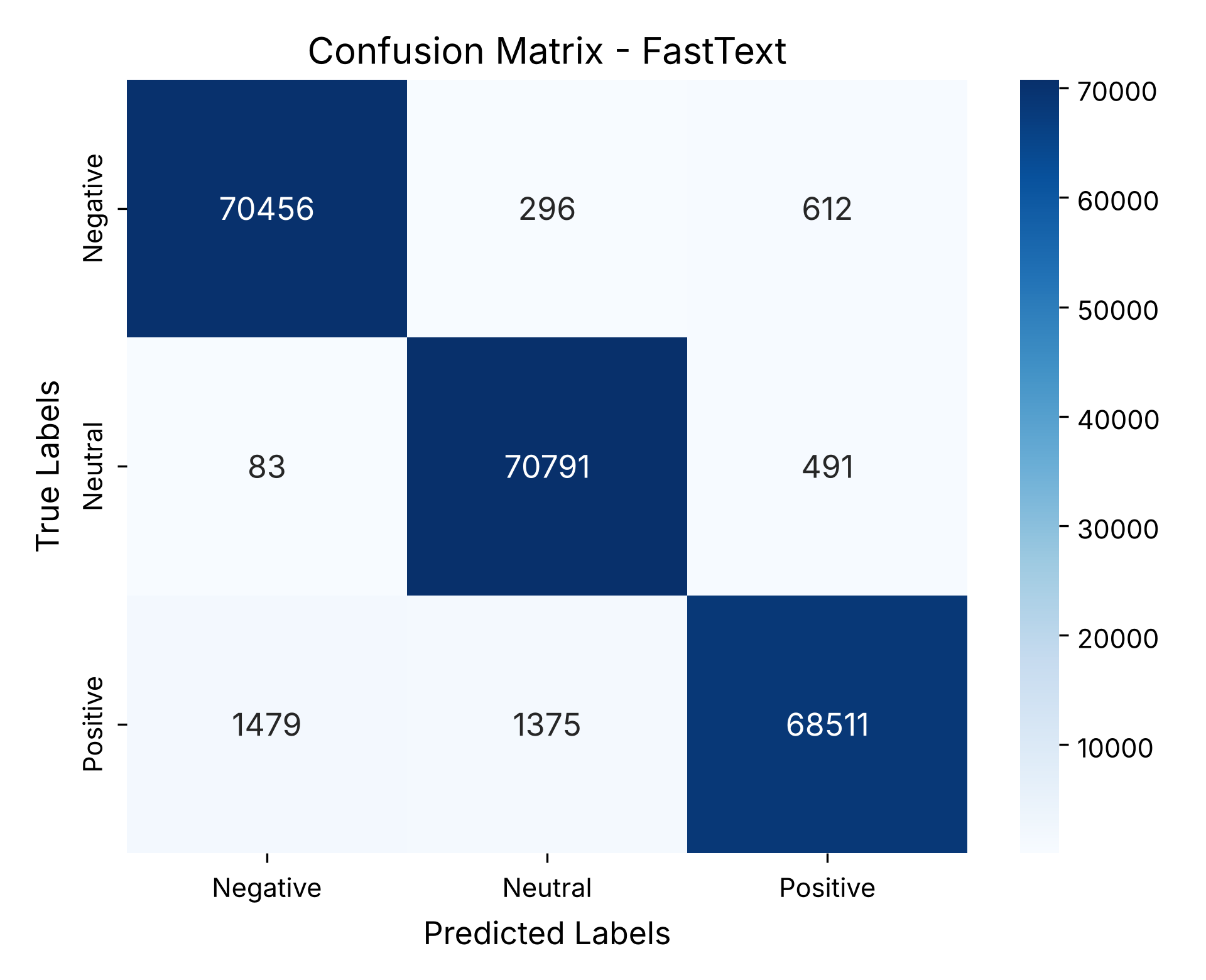}
    \caption{FastText - Confusion Matrix}
    \label{fig:placeholder}
\end{figure}

\subsubsection{Gated Recurrent Unit (GRU)}
The GRU is a streamlined variant of the LSTM architecture, designed to capture long-term dependencies while using fewer parameters and gating mechanisms. GRUs integrate the forget and input gates into a single update gate, allowing for faster training and reduced computational complexity without sacrificing the ability to model sequential patterns. This makes GRUs particularly suitable for sentiment analysis tasks involving long customer reviews.

In this study, the GRU model achieved an overall accuracy of 98\% in sentiment classification using word embeddings on the Amazon customer review dataset. The model delivered strong performance across all sentiment categories, with F1-scores of 0.86 for negative, 0.90 for neutral, and 0.99 for positive sentiments. Precision scores reached 0.89, 0.96, and 0.98 for negative, neutral, and positive classes, respectively, while recall scores remained consistently strong, particularly for positive sentiments, where the model achieved a recall of 0.99, as shown in Figure 40.

With a training time of 9 minutes and 10 seconds, the GRU proved to be both efficient and highly accurate. The model’s performance closely aligns with that of LSTM and Bi-LSTM architectures, demonstrating its effectiveness in capturing contextual dependencies in text despite having a simpler structure. These results highlight the GRU as a powerful and computationally efficient alternative for large-scale sentiment analysis. Nevertheless, the GRU did not reach the same level of accuracy achieved by Bi-LSTM or Attention-based models, which benefit from bidirectional processing and enhanced context modeling. Furthermore, although GRU trains faster than LSTM variants, it is still notably slower than FastText, placing it in a middle ground where it is neither the most accurate nor the most computationally efficient among the evaluated architectures.

\begin{figure}[h]
    \centering
    \includegraphics[width=0.45\linewidth]{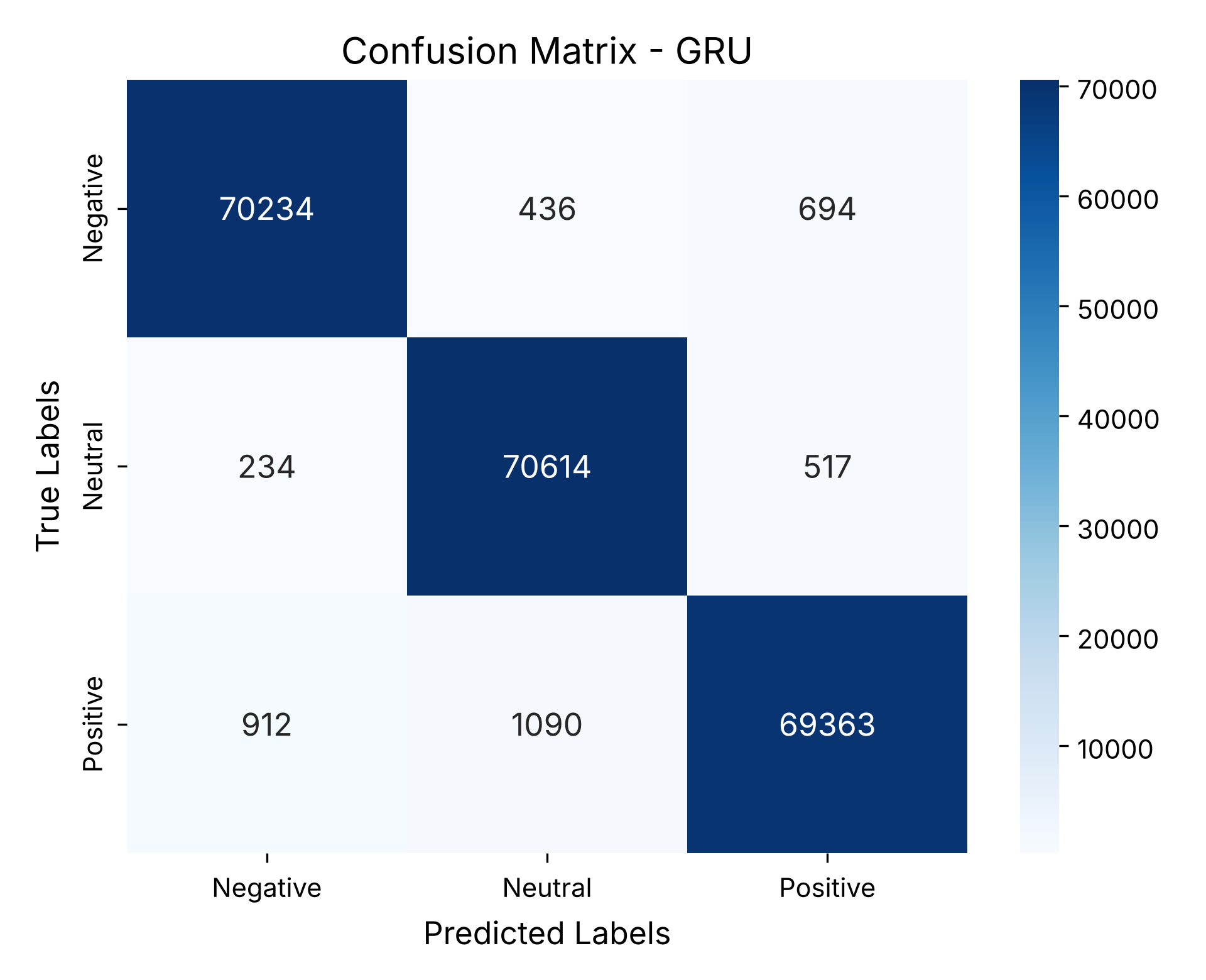}
    \caption{GRU - Confusion Matrix}
    \label{fig:placeholder}
\end{figure}

\subsubsection{Recurrent Neural Networks (RNN)}
The RNN represents the foundational architecture for sequential text modeling, designed to process input one timestep at a time while maintaining a hidden state. Although RNNs provide a basic mechanism to capture temporal patterns in text, they are limited by vanishing and exploding gradient problems, making it difficult for them to learn long-term dependencies. These limitations often lead to reduced performance compared to more advanced architectures such as LSTM, GRU, and Attention-based models.

In this study, the RNN model achieved an overall accuracy of 92\% on the Amazon customer review dataset using word embeddings. The model’s performance varied significantly across the three sentiment classes. While it achieved a high precision of 0.94 and an F1-score of 0.96 for positive sentiments, performance on the negative and neutral classes was substantially weaker. Negative sentiment detection resulted in a recall of 0.29 and an F1-score of 0.38, while the neutral class experienced the lowest performance with a recall of only 0.05 and an F1-score of 0.08, as shown in Figure 41. These results indicate that the RNN struggled to capture complex and nuanced sentiment cues, especially for neutral reviews where contextual understanding is more subtle.

With the longest training time among all models-16 minutes and 53 seconds-the RNN demonstrated both computational inefficiency and reduced predictive accuracy. These findings reinforce the known limitations of traditional RNNs in natural language processing and highlight the importance of more advanced architectures for effective sentiment classification. Given its comparatively low accuracy, especially on negative and neutral classes, and its significantly longer training time, the RNN represents the weakest-performing model in this study. Its inability to capture long-term dependencies places it at a clear disadvantage when compared to LSTM, Bi-LSTM, and Attention models, all of which surpassed 98\% accuracy. Even FastText, despite its simpler design, delivered faster training and stronger overall performance, underscoring the limited practical utility of traditional RNNs for large-scale sentiment analysis.

\begin{figure}[h]
    \centering
    \includegraphics[width=0.45\linewidth]{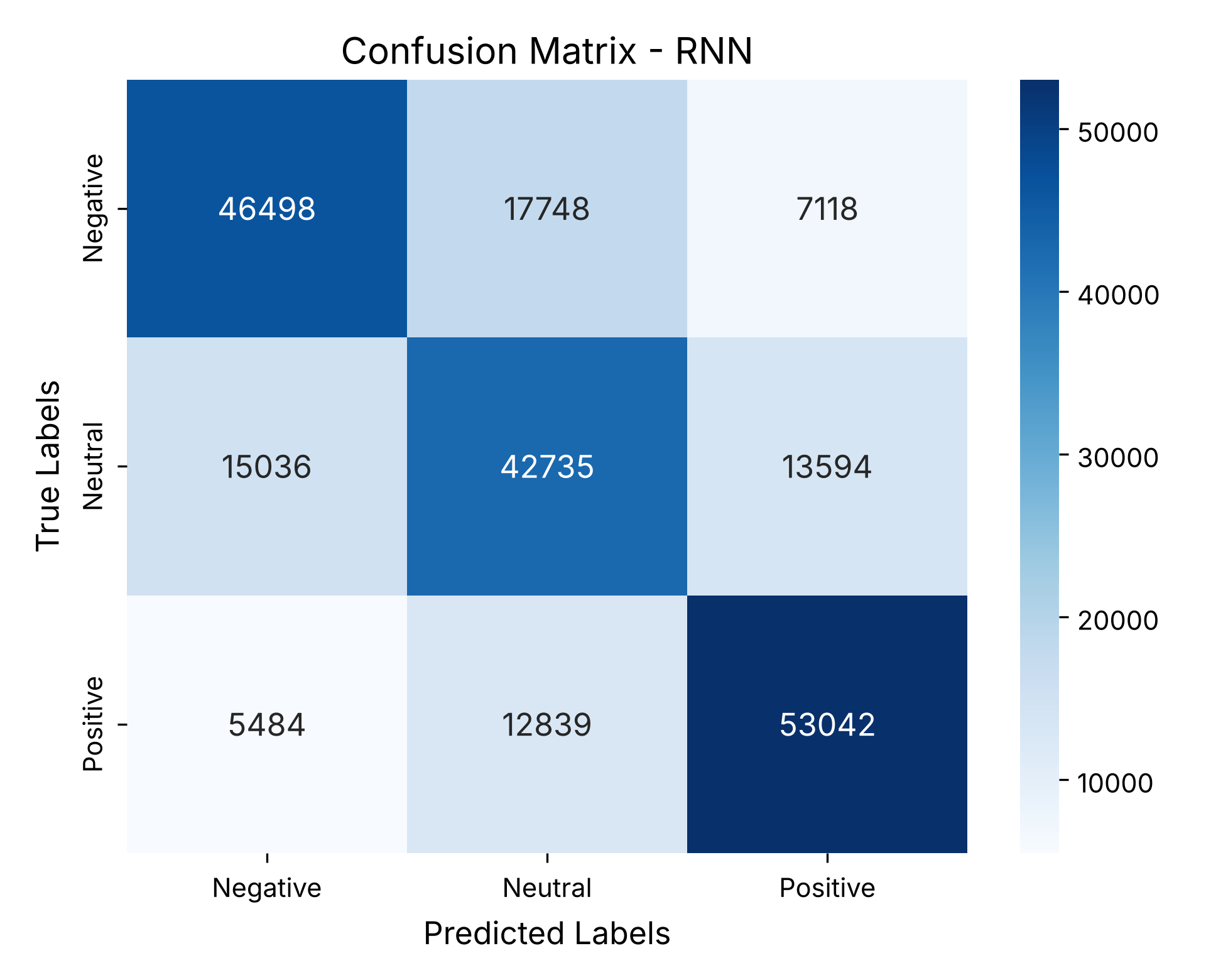}
    \caption{RNN - Confusion Matrix}
    \label{fig:placeholder}
\end{figure}

\subsection{Comparison of Top 3 Models}
The Attention-based model, Bi-LSTM, and LSTM were the top-performing deep learning architectures in this sentiment analysis study. Each model leveraged word embeddings to process textual data and exhibited high accuracy in classifying sentiments within the Amazon customer review dataset. Their individual strengths and differences highlight the impact of model architecture on classification performance.

The Attention-based model emerged as the most effective, achieving 99\% overall accuracy. Its primary advantage lies in its ability to dynamically focus on the most relevant parts of a text sequence, allowing it to capture long-range dependencies and contextual relationships. This mechanism significantly improved classification robustness, leading to minimal misclassifications. The confusion matrix revealed that the Attention model effectively distinguished between sentiment classes, demonstrating superior precision and recall across negative, neutral, and positive categories.

The Bi-LSTM model also achieved 99\% accuracy, processing input sequences in both forward and backward directions. Unlike standard LSTMs, Bi-LSTM considers both past and future contexts, enhancing its ability to model dependencies within a sentence. While its overall performance closely matched the Attention-based model, slight misclassifications in positive and neutral sentiments indicate that contextual understanding was slightly less precise compared to Attention mechanisms. However, Bi-LSTM’s ability to capture bidirectional dependencies made it a strong competitor.

The LSTM model, with an accuracy of 98\%, remained a highly reliable approach. Its memory cell and gating mechanisms allowed it to retain long-term dependencies effectively, making it well-suited for sequential data processing. However, compared to Attention and Bi-LSTM, standard LSTM lacked additional mechanisms for focusing on important words or leveraging bidirectional context, resulting in slightly more misclassified samples.

Overall, while all three models exhibited high precision, recall, and F1 scores, the Attention-based model outperformed the others by better distinguishing between sentiment classes. Bi-LSTM provided improved context modeling, while LSTM served as a strong baseline for sequential text classification. The results emphasize that architecture design plays a crucial role in optimizing sentiment classification, with Attention-based mechanisms proving particularly advantageous for large-scale sentiment analysis.

\begin{figure}[h]
    \centering
    \includegraphics[width=0.75 \linewidth]{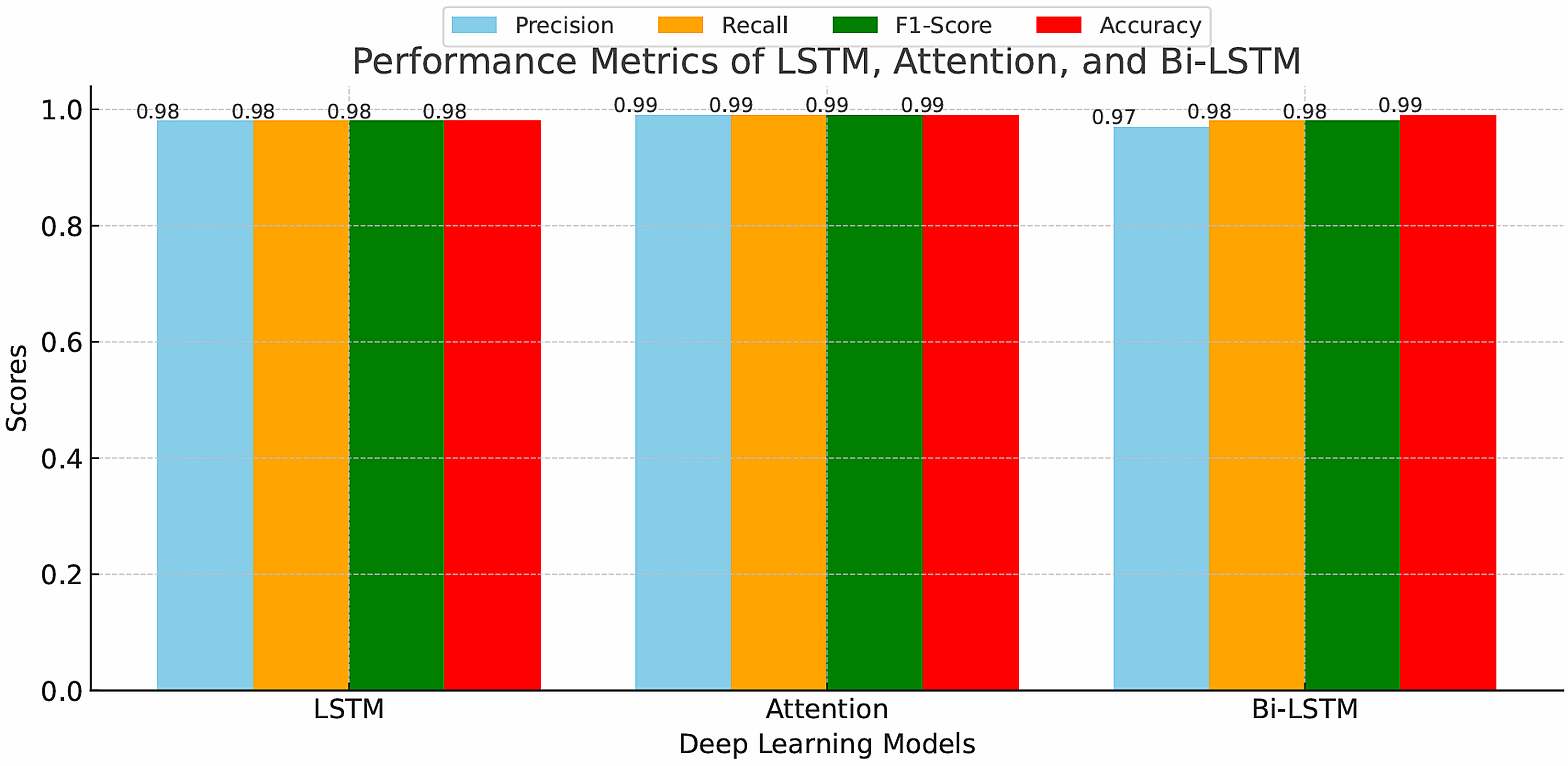}
    \caption{Metrics of Top 3 Model}
    \label{fig:enter-label}
\end{figure}

\subsubsection{Classification Reports}
Figure 42, visually compares the performance of the LSTM, Attention, and Bi-LSTM in a multi-class sentiment analysis task (negative, neutral, positive).

\begin{figure}[h]
    \centering
    \includegraphics[width=0.85 \linewidth]{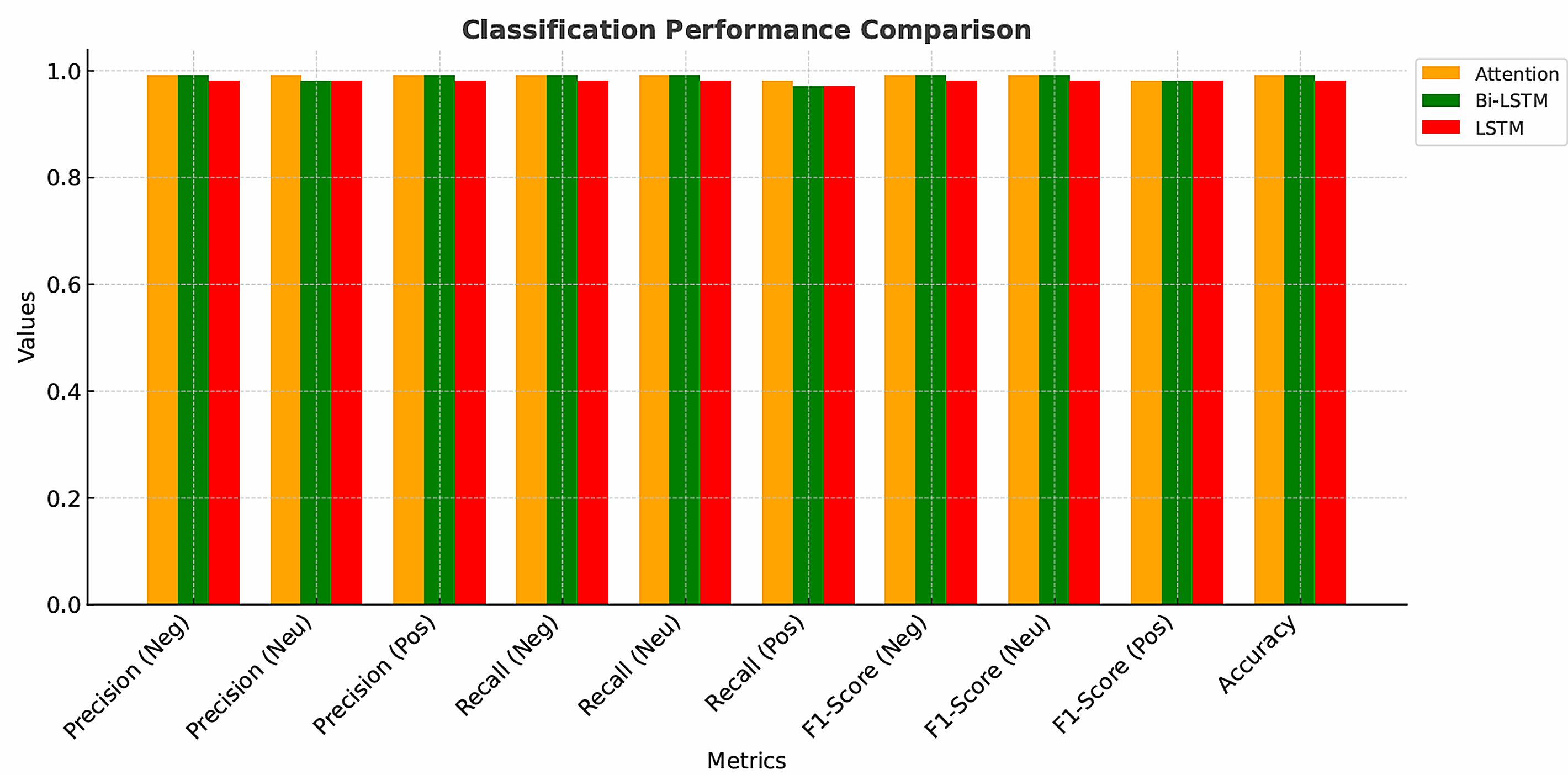}
    \caption{Classification Report Metrics}
    \label{fig:enter-label}
\end{figure}

In Figure 43, the Attention model exhibits superior performance across all metrics, attributed to its ensemble structure, which enables it to effectively capture complex patterns. Both the LSTM and Bi-LSTM models demonstrate solid performance, though slightly lower compared to Attention. This comparison underscores Attention's enhanced adaptability for nuanced sentiment classification, positioning it as the most balanced model for this task, while LSTM and Bi-LSTM also show high accuracy.

\vspace{0.5cm}

\subsubsection{Running Time Comparison} The Attention-based model demonstrated the longest execution time among the deep learning models, taking 112 minutes and 3 seconds for training. Despite this extended runtime, it achieved the highest overall accuracy of 99\%. The ability to dynamically focus on the most relevant parts of a text sequence allowed the Attention model to effectively capture long-range dependencies, significantly enhancing its classification robustness. However, the model's complex architecture, which includes dynamic weighting of sequence elements, contributed to the increased processing time.

In comparison, the Bi-LSTM model, which also achieved 99\% accuracy, had a slightly faster training time of 95 minutes and 20 seconds. By processing input sequences in both forward and backward directions, Bi-LSTM effectively captured bidirectional dependencies within sentences. While its performance was very close to the Attention-based model, the slightly faster runtime and its ability to model both past and future contexts provided a valuable alternative for large-scale sentiment classification tasks. The LSTM model, with an accuracy of 98\%, was the quickest of the three, requiring only 53 minutes and 51 seconds for training. Although it lacked the advanced mechanisms of Attention and bidirectional context, its efficient sequential data processing made it a reliable and faster alternative for sentiment classification tasks.

While all three models exhibited high accuracy, the trade-off between running time and performance is evident. The Attention-based model, though the most accurate, comes with a higher computational cost, whereas LSTM provides a faster, yet slightly less precise, solution. The Bi-LSTM offers a balanced approach, capturing both past and future context with a moderate runtime, making it a competitive model for sentiment analysis in practical applications where training time is a critical consideration.

\begin{table}[h]
\renewcommand{\arraystretch}{1.3}
\begin{center}
\resizebox{0.4\textwidth}{!}{%
\begin{tabular}{|c|c|c|}
\hline
\textbf{Model} & \textbf{Running Time} & \textbf{Accuracy} \\ 
\hline
\textbf{LSTM} & 53m 51s & 0.984 \\ 
\hline
\textbf{RNN} & 34m 47s & 0.664 \\ 
\hline
\textbf{Bi-LSTM} & 95m 20s & 0.986 \\ 
\hline
\textbf{GRU} & 61m 37s & 0.981 \\ 
\hline
\textbf{Attention} & 112m 3s & \cellcolor{green!20} 0.987 \\ 
\hline
\textbf{FastText} & \cellcolor{green!20} 29m 54s & 0.979 \\ 
\hline
\end{tabular}%
}
\end{center}
\caption{Running Time and Accuracy Table}
\label{tab:total_running_time_comparison}
\end{table}

In this section, we assess the performance of various deep learning models used for sentiment analysis, considering both their predictive accuracy and computational efficiency. The efficiency parameter, defined as
\[
\text{Efficiency} = \text{Time} \times \text{Accuracy},
\]
combines the model’s accuracy with its training time to offer a balanced evaluation. Accuracy represents how well the model predicts sentiment, while training time measures the computational resources required. By combining these two factors, the efficiency metric helps identify models that perform well while minimizing computational cost. This is particularly valuable in real-world applications where both speed and accuracy are crucial.

\begin{itemize}
    \item \textbf{LSTM} achieved an accuracy of 0.9842, with precision, recall, and F1-score all above 0.98. However, it requires more training time compared to simpler models, making it less efficient in terms of speed.
    
    \item \textbf{RNN} showed the lowest performance in terms of accuracy (0.6645) among the deep learning models. While it has low training time, its accuracy makes it unsuitable for applications where high precision is critical.
    
    \item \textbf{Bi-LSTM} outperformed LSTM with an accuracy of 0.9862, as well as better precision, recall, and F1-scores. This bidirectional approach provides a more comprehensive understanding of context, but it requires additional computational resources compared to standard LSTM.
    
    \item \textbf{GRU} exhibited an accuracy of 0.9819, with performance metrics close to those of LSTM. It offers better efficiency, requiring less training time while maintaining competitive accuracy levels.
    
    \item \textbf{Attention} emerged as the most effective model, achieving the highest accuracy (0.9871) among all models, along with precision, recall, and F1-scores in the 0.98 range. Despite its superior performance, the training time is higher, making it a trade-off between speed and accuracy.
    
    \item \textbf{FastText} is the fastest model, as highlighted in Table 9, with a significantly lower training time. However, its accuracy (0.9797) is lower than other models, which makes it less suitable for accuracy-critical applications but highly efficient for speed-dependent tasks. 
\end{itemize}

\textbf{Efficiency Analysis}: The \textit{Attention} model stands out as the best-performing model based on accuracy, while also achieving relatively high efficiency. Although it requires more computational resources, its performance justifies the investment. \textit{GRU} and \textit{Bidirectional LSTM} are good alternatives, offering a balance between accuracy and efficiency. \textit{RNN} and \textit{FastText}, while fast, sacrifice accuracy and thus may not be suitable for high-stakes sentiment classification tasks, as shown in Table 10.[12]

\begin{table}[h]
\renewcommand{\arraystretch}{1.7}
\begin{center}
\resizebox{0.5\textwidth}{!}{%
\begin{tabular}{|c|c|c|}
\hline
\textbf{Category} & \textbf{Model} & \textbf{Performance Metric} \\
\hline
\textbf{Fastest Model} & FastText & 29m 54s \\
\hline
\textbf{Most Accurate Model} & Attention & 0.987 Accuracy \\
\hline
\textbf{Most Efficient Model} & Bi-LSTM & E = T $\times$ A \\
\hline
\end{tabular}%
}
\end{center}
\caption{Summary of Best Performing Models by Category}
\label{tab:best_models_summary}
\end{table}

\subsection{Evaluation of Deep Learning Models Under Class Imbalance}

\begin{table}[h]
    \centering
    \caption{Classification Reports of Trained Deep Learning Models for Evaluation}
    \label{tab:classification_dl}
    \renewcommand{\arraystretch}{1.5}
    \resizebox{0.85\textwidth}{!}{
    \rotatebox{0}{
        \begin{tabular}{c c c c c c c c c c c} 
            \hline
            \multicolumn{11}{c}{\textbf{Panel A: Deep Learning Models Classification Report}} \\ 
            \hline 
            \multirow{2}{*}{\textbf{Model}} & \multicolumn{3}{c}{\textbf{Precision}} & \multicolumn{3}{c}{\textbf{Recall}} & \multicolumn{3}{c}{\textbf{F1 Score}} & \multirow{2}{*}{\textbf{Accuracy}} \\ 
            & Negative & Neutral & Positive & Negative & Neutral & Positive & Negative & Neutral & Positive & \\ 
            \hline 
            Attention & 0.9048 & 0.9286 & 0.9849 & 0.8416 & 0.8661 & 0.9923 & 0.8721 & 0.8963 & 0.9886 & 0.9780 \\
            11m 6s \\
            \hline 
            Bi-LSTM & 0.9022 & 0.9426 & 0.9837 & 0.8340 & 0.8478 & 0.9921 & 0.8668 & 0.8927 & 0.9879 & 0.9770 \\
            15m 27s \\
            \hline 
            LSTM & 0.8840 & 0.9429 & 0.9864 & 0.8572 & 0.8785 & 0.9902 & 0.8704 & 0.9096 & 0.9883 & 0.9777 \\
            10m 25s \\
            \hline 
            GRU & 0.8854 & 0.9593 & 0.9837 & 0.8335 & 0.8428 & 0.9909 & 0.8587 & 0.8973 & 0.9873 & 0.9758 \\
            9m 10s \\
            \hline 
            FastText & 0.7677 & 0.7977 & 0.9704 & 0.6629 & 0.8494 & 0.9813 & 0.7115 & 0.8227 & 0.9758 & 0.9533 \\
            3m 26s \\
            \hline 
            RNN & 0.5605 & 0.2372 & 0.9366 & 0.2896 & 0.0499 & 0.9900 & 0.3819 & 0.0825 & 0.9626 & 0.9185 \\
            16m 53s \\
            \hline 
        \end{tabular}
    }}
\end{table}

Six DL models were trained and evaluated separately on balanced and imbalanced datasets. In the balanced dataset, models leveraged structured class distributions to enhance learning, with performance metrics analyzed across sentiment categories. Conversely, in the imbalanced dataset, models encountered challenges related to class dominance, affecting their ability to generalize across all sentiment classes. However, despite the advantages of a balanced dataset, DL models demonstrated relatively consistent performance across both balanced and imbalanced datasets. This is because DL architectures can mitigate class imbalance effects through advanced training strategies, loss function tuning, and representational learning. Techniques such as class weighting, mini-batch gradient descent, Attention mechanisms, and contextual embeddings help models generalize well, reducing the need for explicit data balancing.

While it is widely understood that DL models do not necessarily require balanced datasets to perform effectively, we trained them under both balanced and imbalanced conditions to ensure consistency in comparison with traditional ML models. After evaluating the results, we confirmed that performance differences were minimal across the two settings, validating our prior understanding. Therefore, it was deemed unnecessary to present the imbalanced results for every DL model in this study.

The dataset used for model training was heavily imbalanced, consisting of 356,823 positive, 32,312 negative, and 6,012 neutral samples, totaling 395,147 reviews. Nevertheless, all six DL models demonstrated strong performance, as can be seen on Table 11. This outcome reflects DL models’ stronger ability to handle class imbalance, with a detailed analysis available in [12].

Among the models, the Attention mechanism achieved the highest overall results, reaching 97.8\% accuracy and consistently strong precision, recall, and F1-scores across all classes, especially improving the detection of minority classes (negative and neutral). Bi-LSTM and LSTM closely followed Attention in terms of performance, each maintaining over 97\% accuracy with similarly balanced metrics. GRU also performed well but slightly lagged behind in precision and recall for minority classes compared to Attention and Bi-LSTM. FastText, although faster in training time (3m 26s), showed a moderate drop in precision and recall, particularly for the negative and neutral classes, highlighting its relative simplicity compared to more complex architectures. RNN, on the other hand, significantly underperformed, especially on the neutral class (only 0.0825 F1-score) and negative class, due to its difficulty in handling longer dependencies and class imbalance, despite showing 91.85\% accuracy largely driven by its strong positive class performance.(Table 11)

Overall, the results confirm that modern DL architectures like Attention, Bi-LSTM, and LSTM are robust against class imbalance, successfully capturing patterns even from underrepresented classes. However, simpler or older models like RNN and FastText struggle more under such imbalanced conditions, emphasizing the importance of architectural complexity in achieving balanced and reliable sentiment classification.

\section {Comparison of Deep Learning and Machine Learning models}

\begin{table} [h]
    \centering
    \caption{Classification Reports of Best ML (RFC) and Best DL (Attention) Models Across Dataset Sizes}
    \label{tab:best_models_classification}
    \renewcommand{\arraystretch}{1.5}
    \resizebox{0.85\textwidth}{!}{
    \rotatebox{0}{
        \begin{tabular}{c c c c c c c c c c c c} 
            \toprule
            \multicolumn{12}{c}{\textbf{Panel A: Best ML (RFC) and Best DL (Attention) Models Classification Report}} \\ 
            \midrule
            \multirow{2}{*}{\textbf{Dataset Size}} & \multirow{2}{*}{\textbf{Model}} & \multicolumn{3}{c}{\textbf{Precision}} & \multicolumn{3}{c}{\textbf{Recall}} & \multicolumn{3}{c}{\textbf{F1 Score}} & \multirow{2}{*}{\textbf{Accuracy}} \\ 
            & & Negative & Neutral & Positive & Negative & Neutral & Positive & Negative & Neutral & Positive & \\ 
            \midrule
            \multirow{2}{*}{Imbalanced}  
            \vspace{0.15cm}  
            & RFC 
            & 0.5671 & 0.5005 & 0.9336 & 0.2689 & 0.4082 & 0.9809 & 0.3648 & 0.4497 & 0.9567 & 0.9143 \\ 
            P/356,823 Neu/32,312 N/6,012 T/395,147 & Attention 
            & 0.8805 & 0.9444 & 0.9873 & 0.8745 & 0.8336 & 0.9898 & 0.8775 & 0.8856 & 0.9885 & 0.9780 \\ 
            \midrule
            \multirow{2}{*}{25\% Dataset}  
            \vspace{0.15cm}  
            & RFC 
            & 0.8400 & 0.9700 & 0.8700 & 0.8600 & 1.0000 & 0.8200 & 0.8500 & 0.9800 & 0.8400 & 0.8900 \\ 
            P/Neu/N/32,928 T/98,784 & Attention 
            & 0.8209 & 0.7787 & 0.8568 & 0.8270 & 0.7933 & 0.8344 & 0.8239 & 0.7859 & 0.8455 & 0.8182 \\ 
            \midrule
            \multirow{2}{*}{50\% Dataset}  
            \vspace{0.15cm}  
            & RFC 
            & 0.8800 & 0.9900 & 0.9200 & 0.9200 & 1.0000 & 0.8700 & 0.9000 & 0.9900 & 0.9000 & 0.9300 \\ 
            P/Neu/N/65,857 T/197,571 & Attention 
            & 0.8889 & 0.8566 & 0.9042 & 0.8906 & 0.8768 & 0.8811 & 0.8898 & 0.8666 & 0.8925 & 0.8829 \\ 
            \midrule
            \multirow{2}{*}{75\% Dataset}  
            \vspace{0.15cm}  
            & RFC 
            & 0.9000 & 0.9900 & 0.9600 & 0.9600 & 1.0000 & 0.8900 & 0.9300 & 1.0000 & 0.9300 & 0.9500 \\ 
            P/Neu/N/98,786 T/296,358 & Attention 
            & 0.9275 & 0.8899 & 0.9384 & 0.9181 & 0.9313 & 0.9043 & 0.9228 & 0.9101 & 0.9210 & 0.9179 \\ 
            \midrule
            \multirow{2}{*}{100\% Dataset}  
            \vspace{0.15cm}  
            & RFC 
            & 0.9700 & 0.9800 & 0.9800 & 0.9800 & 0.9800 & 0.9700 & 0.9700 & 0.9800 & 0.9700 & 0.9722 \\ 
            P/Neu/N/131,715 T/395,145 & Attention 
            & 0.9526 & 0.9526 & 0.9526 & 0.9531 & 0.9531 & 0.9531 & 0.9529 & 0.9529 & 0.9529 & 0.9532 \\ 
            \midrule
            \multirow{2}{*}{Oversampled}  
            \vspace{0.15cm}  
            & RFC 
            & 0.9600 & 1.0000 & 1.0000 & 1.0000 & 1.0000 & 0.9600 & 0.9800 & 1.0000 & 0.9800 & 0.9900 \\ 
            P/Neu/N/356,823 T/1,070,469 & Attention 
            & 0.9861 & 0.9851 & 0.9899 & 0.9910 & 0.9938 & 0.9764 & 0.9886 & 0.9894 & 0.9831 & 0.9871 \\ 
            \bottomrule
        \end{tabular}
    }}
\end{table}

\subsection{Machine Learning vs Deep Learning for Balanced Data}

When both ML and DL models were trained on the oversampled, balanced dataset, a significant improvement was observed across all evaluation metrics compared to training on the imbalanced original dataset and Table 12 is showing all parameters in one table. ML models, particularly RFC and L-SVC, demonstrated substantial gains in precision, recall, F1 score, and overall accuracy, achieving results that were much closer to those obtained by the DL models. For instance, the RFC achieved an accuracy of 99\%, while L-SVC and LRC maintained accuracies around 90\%, reflecting the importance of balanced class distributions for traditional ML algorithms.[13]

Despite the improvement in ML model performance after balancing the data, DL models continued to outperform ML models slightly across most metrics. The Attention-based DL model, for example, achieved a precision of 0.9861 for negative class, 0.9851 for neutral class, and 0.9899 for positive class, along with an overall accuracy of 98.71\%. Furthermore, DL models demonstrated more consistent and stable classification across all classes, with particularly high recall and F1 scores, indicating better generalization to minority classes even in the balanced setting.

However, despite the advantages of a balanced dataset, DL models demonstrated relatively consistent performance across both balanced and imbalanced datasets. This is because DL architectures can mitigate class imbalance effects through advanced training strategies, loss function tuning, and representational learning. Techniques such as class weighting, mini-batch gradient descent, Attention mechanisms, and contextual embeddings help models generalize well, reducing the need for explicit data balancing.

Overall, while balancing the dataset significantly enhanced the performance of ML models and narrowed the gap between ML and DL outcomes, DL models still maintained a slight advantage, particularly in handling subtle variations in the data and achieving consistently high performance across all sentiment categories.

\subsection{Machine Learning vs Deep Learning for Imbalanced Data}

DL models consistently outperformed ML models across all metrics on the imbalanced dataset. ML models, being sensitive to class imbalance, showed strong bias toward the majority class, while DL models leveraged their complex architectures to maintain balanced performance.

ML models with BoW achieved 84–93\% accuracy but struggled with minority classes. For example, RFC reached 0.93 precision for the positive class but only 0.27 recall for the negative class. BNB and MNB also performed poorly on the neutral class, with MNB failing completely (Table 7-8).

In contrast, DL models trained with TF-IDF feature extraction achieved consistently higher and more balanced scores across all classes. Models such as Attention, Bi-LSTM, and LSTM achieved accuracy values around 97-98\%, with high precision, recall, and F1 scores across negative, neutral, and positive classes. The Attention-based model, for instance, reached an overall accuracy of 97.8\%, with F1 scores around 0.87, 0.90, and 0.99 for all classes, respectively. DL models also handled the imbalanced data distribution more effectively, showing strong recall values for all classes, unlike ML models that tended to favor the majority (positive) class as can be seen in Table 11-12.

Additionally, it is worth noting that even faster and simpler DL architectures like FastText outperformed most ML models, demonstrating the overall robustness of DL approaches in handling class imbalance challenges.

In conclusion, DL models, particularly Attention, Bi-LSTM, and LSTM, offered significantly better performance than Machine Learning models on the imbalanced dataset, highlighting their strength in capturing complex patterns and managing class imbalance more effectively.

\section{Results}

This study presents a comprehensive exploration of both machine learning and deep learning approaches for large-scale sentiment analysis on Amazon customer reviews, emphasizing the critical roles of feature extraction, model selection, and data balancing techniques. Through extensive experimentation, the research demonstrates that algorithmic performance varies significantly depending on accuracy requirements, computational constraints, and the complexity of textual representations.

Among the machine learning models evaluated, RFC achieved the highest accuracy (0.99 with BoW), though at the cost of substantial computational time. In contrast, Linear SVC provided a strong balance between speed and performance, completing classification in just over one minute with an accuracy of 0.89. These findings underscore the importance of aligning model choice with real-world resource and latency requirements. Furthermore, the impact of feature extraction methods such as Bag-of-Words and TF-IDF proved highly influential, and oversampling strategies effectively addressed class imbalance-validated additionally through undersampled datasets to prevent misleading performance gains.

Deep learning experiments expanded this analysis by evaluating LSTM, Bi-LSTM, GRU, RNN, Attention, and FastText models. The Attention model delivered the highest accuracy (0.978), demonstrating strong capability in capturing long-range dependencies and contextual features, albeit with increased computational cost. FastText, by contrast, offered an optimal efficiency–accuracy trade-off, completing tasks in only three minutes while achieving 0.95 accuracy. These outcomes indicate that deep learning models, especially those leveraging Attention mechanisms or efficient embedding structures, provide substantial advantages for context-rich sentiment tasks. Models such as GRU and LSTM additionally offer practical midpoints between computational burden and predictive performance.

\subsection{Comparison of Top 3 Models Among the Machine Learning and Deep Learning Models}

To assess the overall efficiency and performance of the sentiment analysis system, the three best-performing models-Attention, LSTM, and RFC-were selected for direct comparison based on their accuracy, training time, and model architecture. (Table 13)

\subsubsection{Attention Model}
The Attention model achieved the highest accuracy among all models, reaching \textbf{97.8\%} on the imbalanced dataset. The architecture combines sequence modeling with an Attention mechanism that allows the network to focus on the most relevant parts of the input sequence. This dynamic focus enables better understanding of context, especially in long and complex sentences, contributing to high performance even on underrepresented classes (neg. and neu.). Although training time for the Attention model was relatively high (approx. \textbf{12 minutes}), its balanced precision, recall, and F1-scores across all classes make it an ideal choice for tasks where accuracy and robustness matter more than speed.

\subsubsection{LSTM Model}
The LSTM model followed closely behind the Attention model, maintaining an accuracy of over \textbf{97\%}. LSTMs are designed to capture long-term dependencies in sequential data using memory cells and gating mechanisms, which makes them effective in understanding sentence structure and meaning over time. LSTM showed slightly lower performance on minority classes compared to the Attention model but remained consistent overall. Its training time was moderately high, around \textbf{10 minutes}, due to its sequential computation and complexity.

\subsubsection{Random Forest Classifier (RFC)}
Among the ML models, RFC achieved the highest accuracy of \textbf{99\%} when trained on the \textbf{balanced dataset} using the BoW representation. RFC operates by constructing a multitude of decision trees and aggregating their outputs for prediction, which makes it robust to overfitting and well-suited for structured data. However, it required significantly longer training time-\textbf{16 minutes and 55 seconds}-due to its ensemble nature and complexity in evaluating many decision paths. RFC also relied heavily on balanced class distribution to reach high performance, unlike DL models which performed well even on imbalanced data.

While all three models demonstrated strong performance, the \textbf{Attention model} stands out as the most efficient overall. Despite slightly lower accuracy compared to RFC (97.8\% vs. 99\%), it achieved \textbf{balanced and stable performance across all sentiment classes}, especially under imbalanced data conditions, which is critical in real-world scenarios. Furthermore, it required \textbf{less training time than RFC}, and its architecture is better suited for capturing nuanced contextual information in text. LSTM followed closely in accuracy and efficiency, while RFC, despite high accuracy, was slower and less robust to imbalance. Thus, considering both \textbf{accuracy and robustness to imbalance, along with training efficiency}, the \textbf{Attention model} is the most efficient and practically reliable model among the three.

\begin{table}[h]
\renewcommand{\arraystretch}{1.7}
\begin{center}
\resizebox{0.5\textwidth}{!}{%
\begin{tabular}{|c|c|c|}
\hline
\textbf{Category} & \textbf{Model} & \textbf{Performance Metric} \\
\hline
\textbf{Fastest Model} & MNB (BoW) & 0m 6s \\
\hline
\textbf{Most Accurate Model} & RFC (BoW) & 0.99 Accuracy \\
\hline
\textbf{Most Efficient Model} & FastText & $E = T \times A$ \\
\hline
\textbf{Best Model} & Attention & Accuracy $\sim$ Time \\
\hline
\end{tabular}%
}
\end{center}
\caption{Summary of Best Performing Models by Category}
\label{tab:best_models_summary}
\end{table}

\subsection{Application Phase}

In the final stage of this study, a user-friendly sentiment analysis application was developed to enable real-time classification of customer reviews using the trained machine learning (RFC) and deep learning (Attention) models. The interface is intentionally designed with simplicity and accessibility in mind. Users enter text into the “Enter Review” field and initiate analysis by clicking the “Detect” or “Submit” button. The software system then processes the input and displays the predicted sentiment-positive, negative, or neutral-along with an associated confidence score, allowing users to easily interpret the model's output.

\begin{figure} [h]
    \centering
    \includegraphics[width=0.5\linewidth]{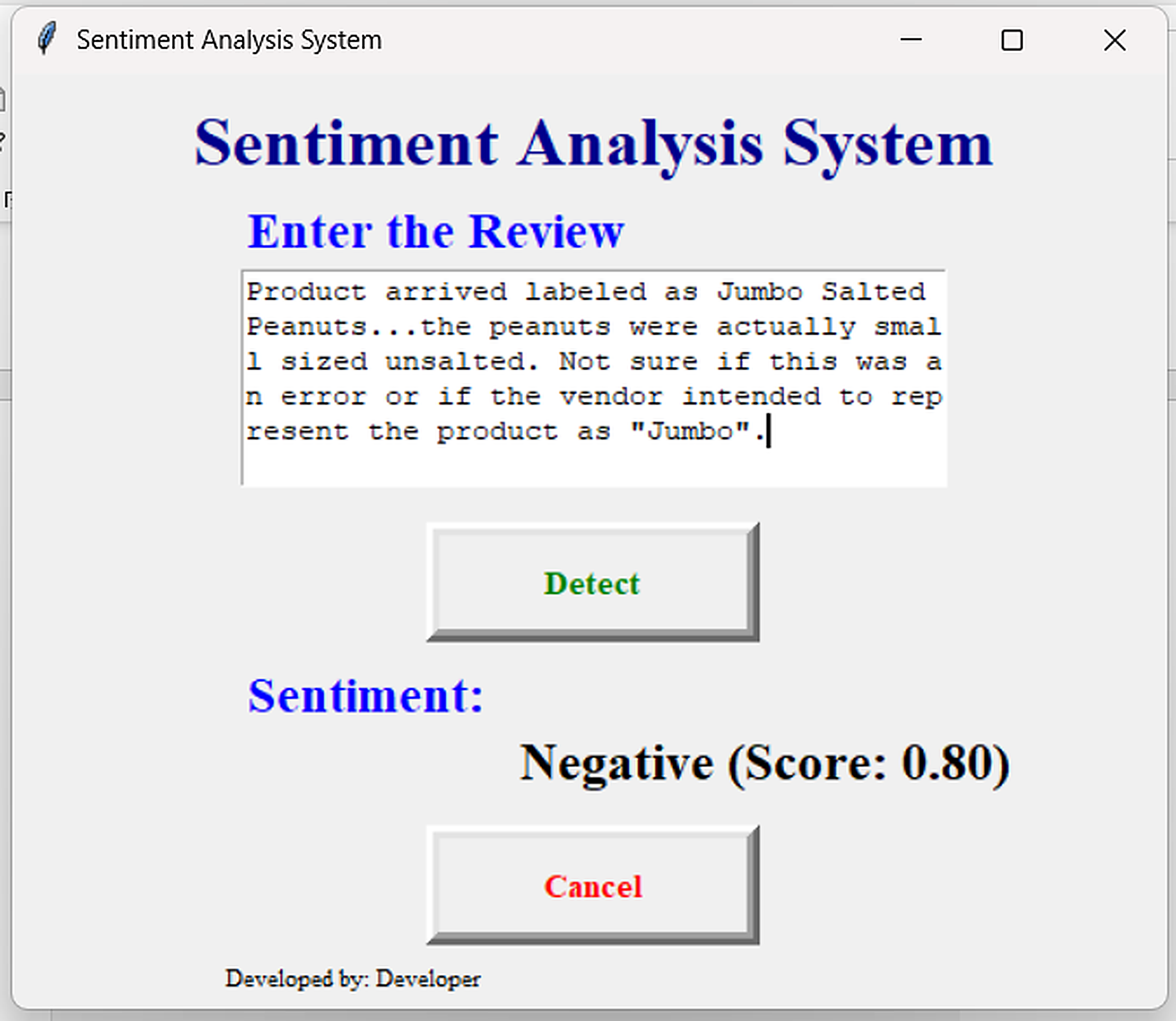}
    \caption{Application Interface}
    \label{fig:enter-label}
\end{figure}

Additional interface features, such as “Clear” for resetting the input and optional reporting functionality (e.g., a “Flag” button), enhance usability and support practical integration into customer feedback analysis workflows, as illustrated in Figure 44. This seamless interaction between the trained models and the interface ensures an efficient and intuitive user experience, making the application a functional tool for real-time sentiment detection across diverse use cases.

Collectively, the results highlight that there is no universal best model for sentiment analysis; rather, model selection must consider the intended operational environment-whether high precision, real-time processing, scalability, or computational efficiency is prioritized. The insights gained from machine learning and deep learning benchmarks contribute to the development of robust, interpretable, and scalable sentiment analysis systems capable of supporting business decision-making processes such as customer experience optimization, product improvement, and trend monitoring.

Future research can build upon this work by integrating transformer-based architectures like BERT, RoBERTa, and GPT, which promise superior contextual understanding. Additional opportunities include developing hybrid Attention–recurrent models, exploring multilingual and noisy-text scenarios, and adopting advanced embedding strategies such as subword tokenization, higher-dimensional vector spaces, and longer sequence modeling. Enhancing interpretability, particularly for Attention-driven and deep neural architectures, remains a key area for improving trust and transparency in business-critical sentiment analysis applications. Together, these advancements will contribute to more adaptable, accurate, and real-time sentiment-driven insights across diverse industries.

\section{Discussion}

The findings of this study provide a comprehensive perspective on the comparative effectiveness of traditional machine learning and deep learning architectures for large-scale sentiment analysis. By analyzing more than 500,000 Amazon customer food reviews, the results demonstrate how model performance can vary considerably depending on the representation method, data balance, and architectural complexity. Overall, the outcomes reinforce established observations from previous studies while also offering new insights into the practical trade-offs between accuracy, computational efficiency, and robustness to class imbalance.

Consistent with the literature, the deep learning models-particularly LSTM, Bi-LSTM, and Attention-exhibited strong capabilities in capturing long-range contextual dependencies, leading to high predictive accuracy. Prior studies have emphasized the superiority of LSTM-based models over classical machine learning approaches due to their gating mechanisms and ability to manage sequential information more effectively. This trend was replicated in our results, where both LSTM and Bi-LSTM achieved performance exceeding 97\%, aligning closely with research reporting accuracies of 90–95\% for similar tasks using Amazon datasets. The Attention model, which surpassed all other deep learning architectures in this study with 97.8\% accuracy, reflects advancements highlighted in recent literature emphasizing the transformative impact of Attention mechanisms and transformer-based systems. These architectures excel by dynamically focusing on the most informative components of textual input, enabling nuanced understanding of sentiment even in complex or lengthy reviews.

The machine learning results also align with prior work suggesting that classical models remain competitive under specific conditions. RFC achieved a remarkably high accuracy of 99\% when trained on a balanced dataset with Bag-of-Words representations, consistent with studies reporting robust performance from ensemble techniques. However, the computational overhead of RFC-nearly 17 minutes of training-mirrors broader critiques that traditional models become inefficient with large feature spaces. Likewise, the strong showing of L-SVC in terms of speed and stability corresponds with earlier research asserting Support Vector Machine as a reliable choice when computational efficiency is prioritized. The dependency of machine learning models on balanced datasets, however, also reinforces findings that these models are notably less resilient to class imbalance than their deep learning counterparts.

The comparative analysis between machine learning and deep learning approaches highlights important implications for real-world sentiment analysis. While deep learning models generally provide superior contextual understanding and robustness, their computational demands may limit deployment in resource-constrained environments. Conversely, machine learning models such as L-SVC, MNB or BNB can offer fast, interpretable, and lightweight alternatives when accuracy is not the sole priority. The results also illustrate that no model performs universally best across all evaluation dimensions-confirming observations in prior research that model selection must consider domain-specific constraints such as latency requirements, hardware availability, dataset size, and class imbalance.

A notable finding is the performance distribution among deep learning models. Although GRU delivered high accuracy while using fewer parameters than LSTM or Bi-LSTM, it did not exceed the performance of these architectures or the Attention model, aligning with the literature that positions GRU as a computationally efficient but slightly less expressive alternative. FastText demonstrated excellent training speed and reasonable accuracy, confirming claims that subword-based embeddings are particularly effective for large-scale applications; however, its lower performance compared to LSTM-based architectures highlights the limitations of shallow models in capturing deeper semantic cues. The RNN, as expected, delivered the weakest performance, supporting long-standing evidence regarding vanishing-gradient limitations and reinforcing why modern sentiment analysis workflows rarely rely on basic RNN architectures.

The implications of these findings extend beyond model benchmarking. From a practical standpoint, the demonstrated variance in performance and speed underscores the necessity of aligning methodological choices with operational objectives. For organizations prioritizing real-time sentiment monitoring, FastText or L-SVC may be optimal. For applications requiring high precision on nuanced or imbalanced datasets-such as customer satisfaction analytics-Attention-based or LSTM-based architectures represent superior solutions. The successful development and integration of a user-facing sentiment analysis application further demonstrate the feasibility of deploying both machine learning and deep learning models in interactive, real-world settings.

Future research directions are abundant. Incorporating transformer-based models such as BERT, RoBERTa, DistilBERT, or GPT could provide substantial gains in contextual understanding and classification accuracy. Extending the system to multilingual sentiment analysis, noisy user-generated content, or domain adaptation scenarios would improve its applicability across industries. Hybrid approaches combining classical machine learning with deep learning-such as transformer embeddings paired with lightweight classifiers-represent another promising avenue. Moreover, improved interpretability techniques, particularly for Attention-based and recurrent architectures, will be critical for increasing transparency and trust in decision-making environments. Finally, exploring more advanced tokenization and embedding strategies, such as byte-pair encoding or high-dimensional contextual embeddings, may further enhance classification performance.

In summary, this study contributes to the ongoing discourse on sentiment analysis by providing a large-scale, detailed comparison of machine learning and deep learning approaches. The results verify insights from previous works while offering practical recommendations for selecting appropriate models under various operational constraints. As sentiment analysis continues to play a central role in understanding user behavior and guiding business strategy, the integration of efficient, context-aware, and interpretable modeling frameworks will remain a priority for future research and application development.

\section{Conclusions}

This study provides a comprehensive evaluation of both machine learning and deep learning techniques for large-scale sentiment analysis on Amazon customer reviews. Through systematic experimentation, we demonstrated that model performance is influenced by feature extraction methods, dataset balancing, and architectural complexity. Deep learning models, particularly LSTM, Bi-LSTM, and Attention-based architectures, achieved the highest accuracy and robustness, effectively capturing contextual dependencies in textual data. Among machine learning models, RFC offered strong predictive performance, though with higher computational cost, while L-SVC presented an efficient alternative when processing speed is critical.

The comparative analysis revealed important trade-offs between speed, accuracy, and interpretability. FastText and GRU provided efficient solutions with competitive performance, yet they were outperformed by Attention-driven and recurrent architectures in nuanced sentiment detection. Basic RNNs, despite being foundational, showed limited capacity to handle long-term dependencies, resulting in lower accuracy. These findings underscore the necessity of aligning model selection with practical application requirements, considering both computational resources and the complexity of the input data.

The outcomes of this research have practical implications for real-world sentiment analysis software systems, including customer feedback monitoring, product evaluation, and business intelligence applications. Furthermore, the development of a user-oriented sentiment analysis interface demonstrates the feasibility of deploying both high-performing deep learning and traditional machine learning models in interactive settings.

Future work may explore transformer-based architectures, hybrid modeling approaches, multilingual applications, and advanced embedding techniques to further enhance sentiment classification performance. Additionally, improving model interpretability remains a critical area to ensure transparency and trust in sentiment-driven decision-making processes.

In conclusion, this study confirms that while deep learning architectures provide superior contextual understanding and accuracy, careful consideration of computational efficiency and operational constraints is essential. By providing a detailed benchmark across multiple models and feature extraction strategies, this research contributes to the development of robust, scalable, and practically applicable sentiment analysis systems.


\begin{thebibliography}{99}

\bibitem{ref1} J. J. McAuley and J. Leskovec, ``From amateurs to connoisseurs: modeling the evolution of user expertise through online reviews'' in \textit{Proceedings of the 22nd International Conference on World Wide Web (WWW ’13)}, Rio de Janeiro, Brazil, May 13-17, 2013, pp. 897--908. doi: 10.1145/2488388.2488466

\bibitem{ref2} E. Coyne, J. Smit, and L. Güner, ``Sentiment analysis for Amazon.com reviews" \textit{ResearchGate}, 2019. doi: 10.13140/RG.2.2.13939.37920. 

\bibitem{ref3} A. S. M. AlQahtani, ``Product sentiment analysis for Amazon reviews" \textit{International Journal of Computer Science \& Information Technology (IJCSIT)}, vol. 13, no. 3, pp. 1--10, Jun. 2021.

\bibitem{ref4} S. Wassan, X. Chen, T. Shen, M. Waqar, and N. Jhanjhi, ``Amazon product sentiment analysis using machine learning techniques" \textit{Int. J. Comput. Sci. Netw. Security}, vol. 30, pp. 695--703, 2021. doi: 10.24205/03276716.2020.2065.

\bibitem{ref5} S. J. Kumaran, ``Sentiment analysis of Amazon user reviews using a hybrid approach'' in \textit{Measurement: Sensors}, vol. 27, 2023, Art. no. 100790.

\bibitem{ref6} M. Qorich and R. El Ouazzani, ``Text sentiment classification of Amazon reviews using word embeddings and convolutional neural networks'' in \textit{J. of Supercomputing}, vol. 79, pp. 11029--11054, 2023.

\bibitem{ref7} T. Haque, N. Saber, and F. Shah, ``Sentiment analysis on large scale Amazon product reviews'' in \textit{ Proc. of 2018 IEEE Int. Conf. on Innovations in Research and Development (ICIRD'18)}, Bangkok, Thailand, May 11-12, 2018, pp. 1--6.

\bibitem{ref8} N. S. Suryawanshi, ``Sentiment analysis with machine learning and deep learning: A survey of techniques and applications,'' \textit{International Journal of Science and Research Archive}, 2024.

\bibitem{ref9} L. Ashbaugh and Y. Zhang, ``A Comparative Study of Sentiment Analysis on Customer Reviews Using Machine Learning and Deep Learning,'' \textit{Computers}, 2024.

\bibitem{ref10} Y. Mamani-Coaquira and E. Villanueva, ``A Review on Text Sentiment Analysis With Machine Learning and Deep Learning Techniques,'' \textit{IEEE Access}, vol. 12, pp. 193115–193130, 2024.

\bibitem{ref11} M. I. Corak, C. Wang, and X. Zhang, ``Sentiment Analysis of Amazon Customer Review by Using Machine Learning Models'' in \textit{Proc. of the 5th IEEE Int. Conf. on Computer Communication and Artificial Intelligence (CCAI '25)}, Haikou, China, May 23-25, 2025, pp. 1--6. 

\bibitem{ref12} M. I. Corak, C. Wang and X. Zhang, ``Sentiment Analysis of Amazon Customer Review by Using Deep Learning Models,'' 2025 IEEE 5th International Conference on Software Engineering and Artificial Intelligence (SEAI ’25), Fuzhou, China, 2025, pp. 161-166.

\bibitem{ref13} M. I. Corak, C. Wang and X. Zhang, ``A Comparative Study of Sentiment Analysis on Amazon Customer Reviews: Deep Learning vs. Machine Learning with Balanced and Imbalanced Datasets,'' 2025 IEEE 16th Annual Ubiquitous Computing, Electronics \& Mobile Communication Conference (UEMCON ’25), New York, USA, 2025
\bibitem{ref14} J. Devlin, M.-W. Chang, K. Lee, and K. Toutanova, ``BERT: Pre-training of deep bidirectional transformers for language understanding'' in \textit{Proceedings of the 2019 Conference of the North American Chapter of the Association for Computational Linguistics: Human Language Technologies (NAACL-HLT)}, vol. 1, Minneapolis, USA, 2019, pp. 4171--4186. doi: 10.18653/v1/N19-1423

\bibitem{ref15} J. Demšar, ``Statistical comparisons of classifiers over multiple data sets,'' \textit{Journal of Machine Learning Research}, vol. 7, pp. 1--30, 2006.

\bibitem{ref16} S. García and F. Herrera, ``An extension on `Statistical comparisons of classifiers over multiple data sets' for all pairwise comparisons,'' \textit{Journal of Machine Learning Research}, vol. 9, pp. 2677--2694, 2008.

\bibitem{ref17} T. G. Dietterich, ``Approximate statistical tests for comparing supervised classification learning algorithms,'' \textit{Neural Computation}, vol. 10, no. 7, pp. 1895--1924, 1998. doi: 10.1162/089976698300017197

\bibitem{ref18} B. Pang and L. Lee, ``Opinion mining and sentiment analysis,'' \textit{Foundations and Trends in Information Retrieval}, vol. 2, no. 1--2, pp. 1--135, 2008.

\bibitem{ref19} A. L. Maas, R. E. Daly, P. T. Pham, D. Huang, A. Y. Ng, and C. Potts, ``Learning word vectors for sentiment analysis'' in \textit{Proceedings of the 49th Annual Meeting of the Association for Computational Linguistics: Human Language Technologies}, Portland, USA, 2011, pp. 142--150.

\bibitem{ref20} A. Vaswani \textit{et al.}, ``Attention is all you need'' in \textit{Advances in Neural Information Processing Systems}, vol. 30, 2017, pp. 5998--6008.

\bibitem{ref21} S. Hochreiter and J. Schmidhuber, ``Long short-term memory,'' \textit{Neural Computation}, vol. 9, no. 8, pp. 1735--1780, 1997.

\bibitem{ref22} K. Cho \textit{et al.}, ``Learning phrase representations using RNN encoder--decoder for statistical machine translation'' in \textit{Proceedings of the 2014 Conference on Empirical Methods in Natural Language Processing (EMNLP)}, Doha, Qatar, 2014, pp. 1724--1734.

\bibitem{ref23} V. Sanh, L. Debut, J. Chaumond, and T. Wolf, ``DistilBERT, a distilled version of BERT: Smaller, faster, cheaper and lighter,'' \textit{arXiv preprint}, arXiv:1910.01108, 2019.

\bibitem{ref24} Y. Liu \textit{et al.}, ``RoBERTa: A robustly optimized BERT pretraining approach,'' \textit{arXiv preprint}, arXiv:1907.11692, 2019.

\bibitem{ref25} A. Joulin, E. Grave, P. Bojanowski, and T. Mikolov, ``Bag of tricks for efficient text classification'' in \textit{Proceedings of the 15th Conference of the European Chapter of the Association for Computational Linguistics (EACL)}, Valencia, Spain, 2017, pp. 427--431.

\bibitem{ref26} P. Bojanowski, E. Grave, A. Joulin, and T. Mikolov, ``Enriching word vectors with subword information,'' \textit{Transactions of the Association for Computational Linguistics}, vol. 5, pp. 135--146, 2017.

\bibitem{ref27} Y. Kim, ``Convolutional neural networks for sentence classification'' in \textit{Proceedings of the 2014 Conference on Empirical Methods in Natural Language Processing (EMNLP)}, Doha, Qatar, 2014, pp. 1746--1751.

\bibitem{ref28} X. Zhang, J. Zhao, and Y. LeCun, ``Character-level convolutional networks for text classification'' in \textit{Advances in Neural Information Processing Systems}, vol. 28, 2015.

\bibitem{ref29} C. J. Hutto and E. Gilbert, ``VADER: A parsimonious rule-based model for sentiment analysis of social media text'' in \textit{Proceedings of the 8th International AAAI Conference on Weblogs and Social Media (ICWSM)}, Ann Arbor, USA, 2014.

\bibitem{ref30} M. Taboada, J. Brooke, M. Tofiloski, K. Voll, and M. Stede, ``Lexicon-based methods for sentiment analysis,'' \textit{Computational Linguistics}, vol. 37, no. 2, pp. 267--307, 2011.

\bibitem{ref31} S. Kiritchenko, X. Zhu, and S. M. Mohammad, ``Sentiment analysis of short informal texts,'' \textit{Journal of Artificial Intelligence Research}, vol. 50, pp. 723--762, 2014.

\bibitem{ref32} H. He and E. A. Garcia, ``Learning from imbalanced data,'' \textit{IEEE Transactions on Knowledge and Data Engineering}, vol. 21, no. 9, pp. 1263--1284, 2009.

\bibitem{ref33} L. Breiman, ``Random forests,'' \textit{Machine Learning}, vol. 45, no. 1, pp. 5--32, 2001.

\bibitem{ref34} F. Pedregosa \textit{et al.}, ``Scikit-learn: Machine learning in Python,'' \textit{Journal of Machine Learning Research}, vol. 12, pp. 2825--2830, 2011.

\end{thebibliography}

\begin{thebibliography}{999}
\bibitem[Aranceta-Bartrina(1999a)]{ref-journal}
Aranceta-Bartrina, Javier. 1999a. Title of the cited article. \textit{Journal Title} 6: 100--10.
\bibitem[Aranceta-Bartrina(1999b)]{ref-book1}
Aranceta-Bartrina, Javier. 1999b. Title of the chapter. In \textit{Book Title}, 2nd ed. Edited by Editor 1 and Editor 2. Publication place: Publisher, vol. 3, pp. 54–96.
\bibitem[Baranwal and Munteanu {[1921]}(1955)]{ref-book2}
Baranwal, Ajay K., and Costea Munteanu. 1955. \textit{Book Title}. Publication place: Publisher, pp. 154--96. First published 1921 (op-tional).
\bibitem[Berry and Smith(1999)]{ref-thesis}
Berry, Evan, and Amy M. Smith. 1999. Title of Thesis. Level of Thesis, Degree-Granting University, City, Country. Identifi-cation information (if available).
\bibitem[Cojocaru et al.(1999)]{ref-unpublish}
Cojocaru, Ludmila, Dragos Constatin Sanda, and Eun Kyeong Yun. 1999. Title of Unpublished Work. \textit{Journal Title}, phrase indicating stage of publication.
\bibitem[Driver et al.(2000)]{ref-proceeding}
Driver, John P., Steffen Rohrs, and Sean Meighoo. 2000. Title of Presentation. In \textit{Title of the Collected Work} (if available). Paper presented at Name of the Conference, Location of Conference, Date of Conference.
\bibitem[Harwood(2008)]{ref-url}
Harwood, John. 2008. Title of the cited article. Available online: URL (accessed on Day Month Year).
\end{thebibliography}
\end{document}